\documentclass[sigconf]{acmart}
\AtBeginDocument{%
  \providecommand\BibTeX{{%
    \normalfont B\kern-0.5em{\scshape i\kern-0.25em b}\kern-0.8em\TeX}}}

\usepackage[utf8]{inputenc} % allow utf-8 input
\usepackage[T1]{fontenc}    % use 8-bit T1 fonts
\usepackage{hyperref}       % hyperlinks
\usepackage{url}            % simple URL typesetting
\usepackage{booktabs}       % professional-quality tables
\usepackage{amsfonts}       % blackboard math symbols
\usepackage{nicefrac}       % compact symbols for 1/2, etc.
\usepackage{microtype}      % microtypography
\usepackage{xcolor}         % colors

\usepackage{amsmath}
\usepackage{mathtools}
\usepackage{amsthm}
\usepackage{rotating}
\usepackage{booktabs}
\usepackage{listings}
\usepackage{mathtools}
\usepackage{enumitem}
\usepackage{balance}

\newtheorem{theorem}{Theorem}[section]
\newtheorem{corollary}[theorem]{Corollary}
\newtheorem{definition}{Definition}

\newtheorem{lemma}[theorem]{Lemma}
\newtheorem{assumption}[theorem]{Assumption}

\theoremstyle{definition} 
\newtheorem{remark}[theorem]{Remark}

\DeclarePairedDelimiter\abs{\lvert}{\rvert}%
\DeclarePairedDelimiter\norm{\lVert}{\rVert}%

\DeclarePairedDelimiterX{\inp}[2]{\langle}{\rangle}{#1, #2}
\newcommand{\matr}[1]{\boldsymbol{#1}}
\newcommand{\vect}[1]{\boldsymbol{#1}}

\makeatletter
\newcommand*{\rom}[1]{\expandafter\@slowromancap\romannumeral #1@}
\makeatother

\usepackage[ruled,linesnumbered]{algorithm2e} % For algorithms

\newcommand{\name}{GNB}

\newcommand{\blemma}{\textbf{Lemma}~}
\newcommand{\btheorem}{\textbf{Theorem}~}


\copyrightyear{2023} 
\acmYear{2023} 
\setcopyright{acmlicensed}\acmConference[KDD '23]{Proceedings of the 29th ACM SIGKDD Conference on Knowledge Discovery and Data Mining}{August 6--10, 2023}{Long Beach, CA, USA}
\acmBooktitle{Proceedings of the 29th ACM SIGKDD Conference on Knowledge Discovery and Data Mining (KDD '23), August 6--10, 2023, Long Beach, CA, USA}
\acmPrice{15.00}
\acmDOI{10.1145/3580305.3599371}
\acmISBN{979-8-4007-0103-0/23/08}

\begin{document}

\title{Graph Neural Bandits}

\author{Yunzhe Qi}
\authornote{Both authors contributed equally.}
\affiliation{%
  \institution{University of Illinois at Urbana-Champaign}
  \country{}
 }
\email{yunzheq2@illinois.edu}

\author{Yikun Ban}
\authornotemark[1]
\affiliation{%
  \institution{University of Illinois at Urbana-Champaign}
  \country{}
} 
\email{yikunb2@illinois.edu}

\author{Jingrui He}
\affiliation{%
  \institution{University of Illinois at Urbana-Champaign}
  \country{}
}
\email{jingrui@illinois.edu}

\renewcommand{\shortauthors}{Yunzhe Qi, Yikun Ban, \& Jingrui He}

\begin{abstract}
  Contextual bandits algorithms aim to choose the optimal arm with the highest reward out of a set of candidates based on the contextual information. Various bandit algorithms have been applied to real-world applications due to their ability of tackling the exploitation-exploration dilemma. Motivated by online recommendation scenarios, in this paper, we propose a framework named \textbf{Graph Neural Bandits (\name)} to leverage the collaborative nature among users empowered by graph neural networks (GNNs). Instead of estimating rigid user clusters as in existing works, we model the ``fine-grained'' collaborative effects through estimated user graphs in terms of exploitation and exploration respectively. Then, to refine the recommendation strategy, we utilize separate GNN-based models on estimated user graphs for exploitation and adaptive exploration. 
  Theoretical analysis and experimental results on multiple real data sets in comparison with state-of-the-art baselines are provided to demonstrate the effectiveness of our proposed framework.
\end{abstract}



\maketitle

\section{Introduction}   \label{sec_introduction}

Contextual bandits are a specific type of multi-armed bandit problem where the additional contextual information (contexts) related to arms are available in each round, 
and the learner intends to refine its selection strategy based on the received arm contexts and rewards.
Exemplary applications include online content recommendation, advertising \citep{linucb-Ind_li2010contextual,colla_environ_2016}, and clinical trials \citep{MAB_clinical_trial-durand2018contextual,MAB_clinical_trial_2_villar2015multi}.
Meanwhile, collaborative effects among users provide researchers the opportunity to design better recommendation strategies, 
since the target user's preference can be inferred based on other similar users. Such effects have been studied by many bandit works \citep{club_2014,SCLUB_li2019improved,CAB_2017,co_filter_bandits_2016,local_clustering-ban2021local}. 
Different from the conventional collaborative filtering methods \citep{NCF_he2017neural,neural_graph_colla_filtering_wang2019neural}, bandit-based approaches focus on more dynamic environments under the online learning settings without pre-training \cite{dynamic_ensemble_wu2019dynamic}, especially when the user interactions are insufficient during the early stage of recommendation (such as dealing with new items or users under the news, short-video recommendation settings), which is also referred to as the ``cold-start'' problem \cite{linucb-Ind_li2010contextual}. In such cases, the exploitation-exploration dilemma \cite{UCB_auer2002finite} inherently exists in the decisions of recommendation.

\iffalse

extensively studied by collaborative filtering methods \cite{NCF_he2017neural,neural_graph_colla_filtering_wang2019neural}. 

since they enable the sharing of user preference information among similar users.
For example, these effects could help the learner adapt to items with limited user feedback. In music recommendation, if two users share similar tastes over a specific genre, then this similarity might generalize to the new music pieces from this genre with a high probability. 

We could therefore leverage the feedback from existing users to infer the preferences of their highly correlated users.
However, compared with static settings where collaborative filtering algorithms have proved their effectiveness, e.g., , it is more difficult to exploit the collaborative effects under the bandit settings due to multiple challenges, such as insufficient user feedback and the exploitation-exploration dilemma. 

\fi

Existing works for clustering of bandits \citep{club_2014,SCLUB_li2019improved,CAB_2017,co_filter_bandits_2016,local_clustering-ban2021local,Meta-Ban} 
model the user correlations (collaborative effects) by clustering users into \textbf{rigid user groups} and then assigning each user group an estimator to learn the underlying reward functions, combined with an Upper Confidence Bound (UCB) strategy for exploration.
However, these works only consider the ``coarse-grained'' user correlations. To be specific, they assume that users from the same group would share identical preferences, i.e., the users from the same group are compelled to make equal contributions to the final decision (arm selection) with regard to the target user. 
Such formulation of user correlations (``coarse-grained'') fails to comply with many real-world application scenarios, because users within the same group can have similar but subtly different preferences.
For instance, under the settings of movie recommendation, although we can allocate two users that ``favor'' the same movie into a user group, their ``favoring levels'' can differ significantly: one user could be a die hard fan of this movie while the other user just considers this movie to be average-to-good. In this case, it would not be the best strategy to vaguely consider they share the same preferences and treat them identically.
Note that similar concerns also exist even when we switch the binary ranking system to a categorical one (e.g., the rating system out of 5 stars or 10 points), because as long as we model with user categories (i.e., rigid user groups), there will likely be divergence among the users within the same group.
Meanwhile, with more fine-sorted user groups, there will be less historical user interactions allocated to each single group because of the decreasing number of associated users. This can lead to the bottlenecks for the group-specific estimators due to the insufficient user interactions.
Therefore, given a target user, it is more practical to assume that the rest of the users would impose different levels of collaborative effects on this user.

\iffalse

\fi

Motivated by the limitations of existing works, in this paper, we propose a novel framework, named \textbf{Graph Neural Bandits (\name)}, to formulate ``fine-grained'' user collaborative effects, where the correlation of user pairs is preserved by user graphs. Given a target user, other users are allowed to make different contributions to the final decision based on the strength of their correlation with the target user.
However, the user correlations are usually unknown, and the learner is required to estimate them on the fly. Here, the learner aims to approximate the correlation between two users by exploiting their past interactions; on the other hand, the learner can benefit from exploring the potential correlations between users who do not have sufficient interactions, or the correlations that might have changed. In this case, we formulate this problem as the exploitation-exploration dilemma in terms of the user correlations.  
To solve this new challenge, \name\ separately constructs two kinds of user graphs, named ``user exploitation graphs'' and ``user exploration graphs''. Then, we apply two individual graph neural networks (GNNs) on the user graphs, to incorporate the collaborative effects in terms of both exploitation and exploration in the decision-making process.
Our main contributions are:

\iffalse

Motivated by aforementioned limitations of existing works, in this paper, we propose a novel framework called \name~based,
on graph neural networks (GNNs). Unlike existing works with rigid estimated user groups where the divergence within each group is neglected, \name~considers \textbf{fine-grained} collaborative effects, where the pair-wise user correlations are modeled through estimated user graphs.
In particular, multiple GNN models are adopted to leverage the collaborative effects among users for better arm recommendation strategies. To be specific, first, based on existing received arm contexts and user feedback, we estimate the collaborative effects on existing records into estimated user graphs named ``user exploitation graphs'' for each candidate arm, and apply the first GNN-based model for arm reward estimation (exploitation).
Meanwhile, to solve the exploitation-exploration dilemma, we formulate the second kind of user graphs for each arm, namely the ``user exploration graphs'', to encode the estimated user correlations for exploration. Then, we utilize a second GNN-based model to estimate the confidence score for reward estimation (exploration) inspired by recent advances of adaptive exploration in contextual bandits \cite{EE-Net_ban2021ee}. 
Our main contributions can be summarized as follows:
\fi

\begin{itemize}[leftmargin=*]
    \vspace{-0.1cm}
    \item \textbf{[Problem Settings]} Different from existing works formulating the ``coarse-grained'' user correlations by neglecting the divergence within user groups, we introduce a new problem setting to model the ``fine-grained'' user collaborative effects via user graphs. Here, pair-wise user correlations are preserved to contribute differently to the decision-making process. (\textbf{Section} \ref{sec_problem_def_notation})
    %
    % \vspace{-0.1cm}
    \item \textbf{[Proposed Framework]} We propose a framework named \name, which has the novel ways to build two kinds of user graphs in terms of exploitation and exploration respectively. Then, \name~utilizes GNN-based models for a refined arm selection strategy by leveraging the user correlations encoded in these two kinds of user graphs for the arm selection. (\textbf{Section} \ref{sec_proposed_framework})
    %
    % \vspace{-0.1cm}
    \item \textbf{[Theoretical Analysis]} With standard assumptions, we provide the theoretical analysis showing that \name\ can achieve the regret upper bound of complexity $ \mathcal{O}(\sqrt{T\log(Tn)})$, where $T$ is the number of rounds and $n$ is the number of users. This bound is sharper than the existing related works. (\textbf{Section} \ref{sec_theoretical_analysis})
    %
    % \vspace{-0.1cm}
    \item \textbf{[Experiments]} Extensive experiments comparing \name\ with nine state-of-the-art algorithms are conducted on various data sets with different specifications, which demonstrate the effectiveness of our proposed \name\ framework. (\textbf{Section} \ref{sec_experiments})
\end{itemize}
\vspace{-0.05cm}
Due to the page limit, interested readers can refer to the paper Appendix for supplementary contents.

\vspace{-0.2cm}
\section{Related Works}   \label{sec_related_works}
Assuming the reward mapping function to be linear, the linear upper confidence bound (UCB) algorithms \citep{Lin_UCB-2011,linucb-Ind_li2010contextual,UCB_auer2002finite,improved_linear_bandits_abbasi2011improved} were first proposed to tackle the exploitation-exploration dilemma. 
After kernel-based methods \citep{kernel_ucb-2013,kmtl-ucb_2017} were used to tackle the kernel-based reward mapping function under the non-linear settings, neural algorithms \citep{Neural-UCB,neural_thompson-zhang2020neural,neural_multifacet-ban2021multi} have been proposed to utilize neural networks to estimate the reward function and confidence bound. 
Meanwhile, AGG-UCB \citep{AGG-UCB_qi2022neural} adopts GNN to model the arm group correlations. GCN-UCB \citep{GCN-UCB_upadhyay2020graph} manages to apply the GNN model to embed arm contexts for the downstream linear regression, and GNN-PE \citep{GNN-PE_kassraie2022graph} utilizes the UCB based on information gains to achieve exploration for classification tasks on graphs.
Instead of using UCB, EE-Net \citep{EE-Net_ban2021ee} applies a neural network to estimate prediction uncertainty.
Nonetheless, all of these works fail to consider the collaboration effects among users under the real-world application scenarios.

To model user correlations, \citep{colla_environ_2016,GangOfBandits-cesa2013gang} assume the user social graph is known, and apply an ensemble of linear estimators. Without the prior knowledge of user correlations, CLUB \citep{club_2014} introduces the user clustering problem with the graph-connected components, and SCLUB \citep{SCLUB_li2019improved} adopts dynamic user sets and set operations, while DynUCB \citep{Dyn-UCB_nguyen2014dynamic} assigns users to their nearest estimated clusters. Then, CAB \citep{CAB_2017} studies the arm-specific user clustering, and LOCB \citep{local_clustering-ban2021local} estimates soft-margin user groups with local clustering. COFIBA \citep{co_filter_bandits_2016} utilizes user and arm co-clustering for collaborative filtering.
Meta-Ban \citep{Meta-Ban} applies a neural meta-model to adapt to estimated user groups.
However, all these algorithms consider rigid user groups, where users from the same group are treated equally with no internal differentiation. 
Alternatively, we leverage GNNs \citep{GCN_kipf2016semi,fastGCN-chen2018fastgcn,SGC_wu2019simplifying,APPNP-klicpera2018predict,light_GCN-he2020lightgcn,GNN_Fully_Connected_satorras2018few,GNN_community_detection-you2019position,Graph-Recommender_ying2018graph} to learn from the ``fine-grained'' user correlations and arm contexts simultaneously.

\vspace{-0.2cm}

\section{GNB: Problem Definition}    \label{sec_problem_def_notation}

Suppose there are a total of $n$ users with the user set $\mathcal{U} = \{1, \cdots, n\}$.
At each time step $t\in [T]$, the learner will receive a target user $u_{t} \in \mathcal{U}$ to serve, along with candidate arms $\mathcal{X}_{t} = \{\vect{x}_{i, t}\}_{i\in [a]}$, $\abs{\mathcal{X}_{t}}=a$.
%The cardinality of this arm set is $\abs{\mathcal{X}_{t}} = a$, and 
Each arm is described by a $d$-dimensional context vector $\vect{x}_{i, t}\in \mathbb{R}^{d}$ with $\norm{\vect{x}_{i, t}}_{2} = 1$, and $\vect{x}_{i, t} \in \mathcal{X}_{t}$ is also associated with a reward $r_{i,t}$. As the user correlation is one important factor in determining the reward, we define the following reward function:  
\begin{equation}
\begin{split}
    r_{i, t} =  h(\vect{x}_{i, t}, u_{t}, \mathcal{G}^{(1), *}_{i, t}) + \epsilon_{i, t}
\end{split}
\label{eq_weighted_reward_func}
\end{equation}
where $h(\cdot)$ is the unknown reward mapping function, and $\epsilon_{i, t}$ stands for some zero-mean noise such that $\mathbb{E}[r_{i,t}] =  h(\vect{x}_{i, t}, u_{t}, \mathcal{G}^{(1), *}_{i, t})$. 
Here, we have $\mathcal{G}^{(1), *}_{i, t} = (\mathcal{U}, E, W^{(1), *}_{i, t})$ being the \textbf{unknown} user graph induced by arm $\vect{x}_{i, t}$, which encodes the ``fine-grained'' user correlations in terms of the \textbf{expected rewards}. In graph $\mathcal{G}^{(1), *}_{i, t}$, each user $u\in \mathcal{U}$ will correspond to a node; meanwhile, $E = \{e(u, u')\}_{ u, u' \in \mathcal{U}}$ refers to the set of edges, and the set $W^{(1), *}_{i, t} = \{w^{(1), *}_{i, t}(u, u')\}_{u, u' \in \mathcal{U}}$ stores the weights for each edge from $E$.
Note that under real-world application scenarios, users sharing the same preference for certain arms (e.g., sports news) may have distinct tastes over other arms (e.g., political news).
Thus, we allow each arm $\vect{x}_{i, t} \in \mathcal{X}_{t}$ to induce different user collaborations $\mathcal{G}^{(1), *}_{i, t}$. 

Then, motivated by various real applications (e.g., online recommendation with normalized ratings), we consider $r_{i,t}$ to be bounded $r_{i,t}\in[0, 1]$, which is standard in existing works (e.g., \cite{club_2014,CAB_2017,local_clustering-ban2021local,Meta-Ban}). Note that as long as $r_{i,t} \in [0, 1]$, we do not impose the distribution assumption (e.g., sub-Gaussian distribution) on noise term $\epsilon_{i,t}$.

\textbf{[Reward Constraint]}
To bridge user collaborative effects with user preferences (i.e., rewards), we consider the following constraint for reward function in \textbf{Eq.} \ref{eq_weighted_reward_func}. The intuition is that for any two users with comparable user correlations, they will incline to share similar tastes for items.
For arm $\vect{x}_{i, t}$, we consider the difference of expected rewards between any two users $u, u' \in \mathcal{U}$ to be governed by
\begin{equation}
\begin{split}
    \big| \mathbb{E}[r_{i,t} | u, \vect{x}_{i, t}] - \mathbb{E}[r_{i,t} | u', \vect{x}_{i, t}] \big| \leq \Psi \big(\mathcal{G}^{(1), *}_{i, t}[u:], \mathcal{G}^{(1), *}_{i, t}[u':] \big)
\end{split}   
\label{eq_assumption_correlation_vector}
\end{equation}
where $\mathcal{G}^{(1), *}_{i, t}[u:]$ represents the normalized adjacency matrix row of $\mathcal{G}^{(1), *}_{i, t}$ that corresponds to user (node) $u$, and $\Psi: \mathbb{R}^{n}\times\mathbb{R}^{n} \mapsto \mathbb{R}$ 
denotes an unknown mapping function. 
The reward function definition (\textbf{Eq.} \ref{eq_weighted_reward_func}) and the constraint (\textbf{Eq.} \ref{eq_assumption_correlation_vector}) motivate us to design the \name\ framework, to be introduced in Section \ref{sec_proposed_framework}.

Then, we proceed to give the formulation of $\mathcal{G}^{(1), *}_{i, t} = (\mathcal{U}, E, W^{(1), *}_{i, t})$ below.
Given arm $\vect{x}_{i, t} \in \mathcal{X}_{t}$, users with strong correlations tend to have similar expected rewards, which will be reflected by $W^{(1), *}_{i, t}$.

%
% For both kinds of user graphs, each user from $\mathcal{U}$ is mapped to a corresponding node in node set $\mathcal{U}$. With $E = \{e(u_{i}, u_{j})\}_{\forall u_{i}, u_{j} \in \mathcal{U}}$ being the set of edges, we have $W^{(1), *}_{i, t}, W^{(2), *}_{i, t}$ to respectively represent the set of edge weights for $\mathcal{G}^{(1), *}_{i, t}, \mathcal{G}^{(2), *}_{i, t}$. Here, the estimated user (exploitation / exploration) correlations are modeled by the edge weights of node (user) pairs. 
% Next, we proceed to give the definitions of two arm-specific user correlations, which are encoded by $\mathcal{G}^{(1), *}_{i, t}, \mathcal{G}^{(2), *}_{i, t}$ respectively.

% ====================================================
\begin{definition}[User Correlation for Exploitation]  \label{def_exploitation_simi}
In round $t$, for any two users $u, u'\in \mathcal{U}$, their exploitation correlation score $w_{i, t}^{(1), *}(u, u')$ w.r.t. a candidate arm $\vect{x}_{i, t} \in \mathcal{X}_{t}$ is defined as
\begin{displaymath}
\begin{aligned}
    w_{i, t}^{(1), *}(u, u') = \Psi^{(1)} \big( \mathbb{E}[r_{i, t}| u,~ \vect{x}_{i, t}],~ \mathbb{E}[r_{i, t}| u',~ \vect{x}_{i, t}] \big)
\end{aligned}
\end{displaymath}
where $\mathbb{E}[r_{i, t}| u,~ \vect{x}_{i, t}], i\in [a]$ is the expected reward in terms of the user-arm pair $(u, \vect{x}_{i, t})$. Given two users $u, u'\in \mathcal{U}$, the function $\Psi^{(1)}: \mathbb{R} \times \mathbb{R} \mapsto \mathbb{R}$ maps from their expected rewards $\mathbb{E}[r_{i, t}| u,~ \vect{x}_{i, t}]$ to their user exploitation score $w_{i, t}^{(1), *}(u, u')$.
\end{definition}

% ----------------------------------------
%Given an arm $\vect{x}_{i, t} \in \mathcal{X}_{t}$, the user correlation for exploitation measures the user preference (i.e., expected reward) correlation between two users $u, u' \in \mathcal{U}$, and the corresponding exploitation score $w_{i, t}^{(1), *}(u, u')$ refers to the edge weight between these two users (nodes) $u, u'$ in exploitation graph $\mathcal{G}_{i, t}^{(1), *}$.

The edge weight $w_{i, t}^{(1), *}(u, u')$ measures the correlation between the two users' preferences. 
When $w_{i, t}^{(1), *}(u, u')$ is large, $u$ and $u'$ tend to have the same taste; Otherwise, these two users' preferences will be different in expectation.
In this paper, we consider the mapping functions $\Psi^{(1)}$ as the prior knowledge. For example,   $\Psi^{(1)}$ can be the radial basis function (RBF) kernel or normalized absolute difference.

\textbf{[Modeling with User Exploration Graph $\mathcal{G}^{(2), *}_{i, t}$]}
Unfortunately, $\mathcal{G}^{(1), *}_{i, t}$ is the \textbf{unknown} prior knowledge in our problem setting. 
Thus, the learner has to estimate $\mathcal{G}^{(1), *}_{i, t}$ by exploiting the current knowledge, denoted by $\mathcal{G}^{(1)}_{i, t} =  (\mathcal{U}, E, W^{(1)}_{i, t})$, 
where $W^{(1)}_{i, t} = \{w^{(1)}_{i, t}(u, u')\}_{u, u' \in \mathcal{U}}$  is the estimation of $W^{(1), *}_{i, t}$ based on the function class $\mathcal{F}= \{ f_u^{(1)} \}_{u \in \mathcal{U}}$ where $f_u^{(1)}$ is the hypothesis specified to user $u$.
However, greedily exploiting $\mathcal{G}^{(1)}_{i, t}$ may lead to the sub-optimal solution, or overlook some correlations that may only be revealed in the future rounds.
Thus, we propose to construct another \textbf{user exploration graph} $\mathcal{G}^{(2), *}_{i, t}$ for principled exploration to measure the estimation gap $\mathcal{G}^{(1), *}_{i, t} - \mathcal{G}^{(1)}_{i, t}$, which refers to the uncertainty of the estimation of graph $\mathcal{G}^{(1)}_{i, t}$.

For each arm $\vect{x}_{i, t} \in \mathcal{X}_{t}$, we formulate the user exploration graph $\mathcal{G}^{(2), *}_{i, t} = (\mathcal{U}, E, W^{(2), *}_{i, t})$,
with the set of edge weights $W^{(2), *}_{i, t} = \{w^{(2), *}_{i, t}(u, u')\}_{\forall u, u' \in \mathcal{U}}$.
Here, $\mathcal{G}^{(2), *}_{i, t}$ models the uncertainty of estimation $\mathcal{G}^{(1)}_{i, t}$ in terms of the true exploitation graph $\mathcal{G}^{(1), \ast}_{i, t}$, and $\mathcal{G}^{(2), *}_{i, t}$ can be thought as the oracle exploration graph, i.e., "perfect exploration".
Then, with the aforementioned hypothesis $f_{u}^{(1)}(\vect{x}_{i, t})$ for estimating the expected reward of arm $\vect{x}_{i, t}$ given $u$, we introduce the formulation of $\mathcal{G}^{(2), *}_{i, t}$ as the user exploration correlation.

% ----------------------------------------
\begin{definition}[User Correlation for Exploration]  \label{def_exploration_simi}
In round $t$, given two users $u, u'\in \mathcal{U}$ and an arm $\vect{x}_{i, t} \in \mathcal{X}_{t}$, their underlying exploration correlation score is defined as 
% $w_{i, t}^{(2), *}(u, u')$ 
\begin{displaymath}
\begin{aligned}
  w_{i, t}^{(2), *}&(u, u') \\
   & = \Psi^{(2)} \bigg( \mathbb{E}[r_{i, t}| u, \vect{x}_{i, t}] - f_{u}^{(1)}(\vect{x}_{i, t}), \mathbb{E}[r_{i, t}| u', \vect{x}_{i, t}] - f_{u'}(\vect{x}_{i, t})\bigg)
\end{aligned}
\end{displaymath}
with $\mathbb{E}[r_{i, t}| u,~ \vect{x}_{i, t}] - f_{u}^{(1)}(\vect{x}_{i, t}), i\in [a]$ being the potential gain of the estimation $f_{u}^{(1)}(\vect{x}_{i, t})$ for the user-arm pair $(u, \vect{x}_{i, t})$. Here, $ f_{u}^{(1)}(\cdot )$ is the reward estimation function specified to user $u$, and
$\Psi^{(2)}: \mathbb{R} \times \mathbb{R} \mapsto \mathbb{R}$ is the mapping from user potential gains $\mathbb{E}[r_{i, t}| u,~ \vect{x}_{i, t}] - f_{u}^{(1)}(\vect{x}_{i, t})$ to their exploration correlation score.
\end{definition}
Here, $w_{i, t}^{(2), *}(u, u')$ is defined based on the potential gain of $f_{u}^{(1)}(\cdot)$, i.e., $\mathbb{E}[r_{i, t}| u,~ \vect{x}_{i, t}] - f_{u}^{(1)}(\vect{x}_{i, t})$, to measure the estimation uncertainty. Note that our formulation is distinct from the formulation in \cite{EE-Net_ban2021ee}, where they only focus on the single-bandit setting with no user collaborations, and all the users will be treated identically.

As we have discussed, $w_{i, t}^{(2), *}(u, u')$ measures the uncertainty of estimation $w_{i, t}^{(1)}(u, u')$. When $w_{i, t}^{(2), *}(u, u')$ is large, the uncertainty of estimated exploitation correlation, i.e., $w_{i, t}^{(1)}(u, u')$, will also be large, and we should explore them more. Otherwise, we have enough confidence towards $w_{i, t}^{(1)}(u, u')$, and we can exploit $w_{i, t}^{(1)}(u, u')$ in a secure way.
Analogous to $\Psi^{(1)}$ in \textbf{Def.} \ref{def_exploitation_simi}, we consider the mapping function $\Psi^{(2)}$ as the known prior knowledge.

% -------------

% -------------
\textbf{[Learning Objective]}
With the received user $u_{t}$ in each round $t\in [T]$, the learner is expected to recommend an arm $x_{t}\in \mathcal{X}_{t}$ (with reward $r_{t}$) in order to minimize the cumulative pseudo-regret
\begin{equation}
\begin{split}
    R(T) = \mathbb{E}[\sum_{t=1}^{T}(r_{t}^{*} - r_{t})] 
\end{split}
\label{eq_pseudo_regret}
\end{equation}
where we have $r_{t}^{*}$ being the reward for the optimal arm, such that 
$ \mathbb{E}[r_{t}^{*}| u_t, \mathcal{X}_{t}] = \max_{\vect{x}_{i,t}\in \mathcal{X}_{t}} h(\vect{x}_{i, t}, u_{t}, \mathcal{G}^{(1), *}_{i, t})$.

% ------------------------
\textbf{[Comparing with Existing Problem Definitions]}
The problem definition of existing user clustering works (e.g., \cite{club_2014,SCLUB_li2019improved,CAB_2017,local_clustering-ban2021local,Meta-Ban}) only formulates "coarse-grained" user correlations. In their settings, for a user group $\mathcal{N} \subseteq \mathcal{U}$ with the mapping function $h_{\mathcal{N}}$, all the users in $\mathcal{N}$ are forced to share the same reward mapping given an arm $\vect{x}_{i, t}$, i.e., $\mathbb{E}[r_{i,t}\mid u, \vect{x}_{i,t}]= h_{\mathcal{N}}(\vect{x}_{i,t}), \forall u \in \mathcal{N}$. In contrast, our definition of the reward function enables us to model the pair-wise ``fine-grained'' user correlations by introducing another two important factors $u$ and $\mathcal{G}^{(1), *}_{i, t}$. With our formulation, each user here is allowed to produce different rewards facing the same arm, i.e.,  $\mathbb{E}[r_{i,t}\mid u, \vect{x}_{i,t}]= h(\vect{x}_{i,t}, u, \mathcal{G}^{(1), *}_{i, t}), \forall u \in \mathcal{N}$. Here, with different users $u$, the corresponding expected reward $h(\vect{x}_{i,t}, u, \mathcal{G}^{(1), *}_{i, t})$ can be different. 
%ur definition of the reward function enables us to model the pair-wise fine-grained user correlations instead of imposing rigid user groups, where users within the same group are forced to share the identical preference. 
%
%Note that compared with previous linear methods (e.g., \cite{GangOfBandits-cesa2013gang,colla_environ_2016}) which simply aggregate the expected rewards and assume user correlations are \textbf{known} to the learner, our assumption is more generic and can fit into non-linear settings. 
Therefore, our definition of the reward function is more generic, and it can also readily generalize to existing user clustering algorithms (with ``coarse-grained'' user correlations) by allowing each single user group to form an isolated sub-graph in $\mathcal{G}^{(1), *}_{i, t}$ with no connections across different sub-graphs (i.e., user groups).

% ------------------------------------------------------------

% ---
\textbf{[Notation]}
Up to round $t$, denoting $\mathcal{T}_{u, t} \subseteq [t]$ as the collection of time steps at which user $u\in \mathcal{U}$ has been served, we use $\mathcal{P}_{u, t} = \{(\vect{x}_{\tau}, r_{\tau})\}_{\tau\in \mathcal{T}_{u, t}}$ to represent the collection of received arm-reward pairs associated with user $u$, and $T_{u, t} = \abs{\mathcal{T}_{u, t}}$ refers to the corresponding number of rounds.
Here, $\vect{x}_{\tau} \in \mathcal{X}_{\tau}, r_{\tau} \in [0, 1]$ separately refer to the chosen arm and actual received reward in round $\tau\in [t]$.
Similarly, we use $\mathcal{P}_{t} = \{(\vect{x}_{\tau}, r_{\tau})\}_{\tau\in [t]}$ to denote all the past records (i.e., arm-reward pairs), up to round $t$. 
For a graph $\mathcal{G}$, we let $\matr{A} \in \mathbb{R}^{n \times n}$ be its adjacency matrix (with self-loops), and $\matr{D} \in \mathbb{R}^{n \times n}$ refers to the corresponding degree matrix.

\begin{figure*}[t]
  \centering
  \vspace{-0.40cm}
  \includegraphics[width=0.8\linewidth]{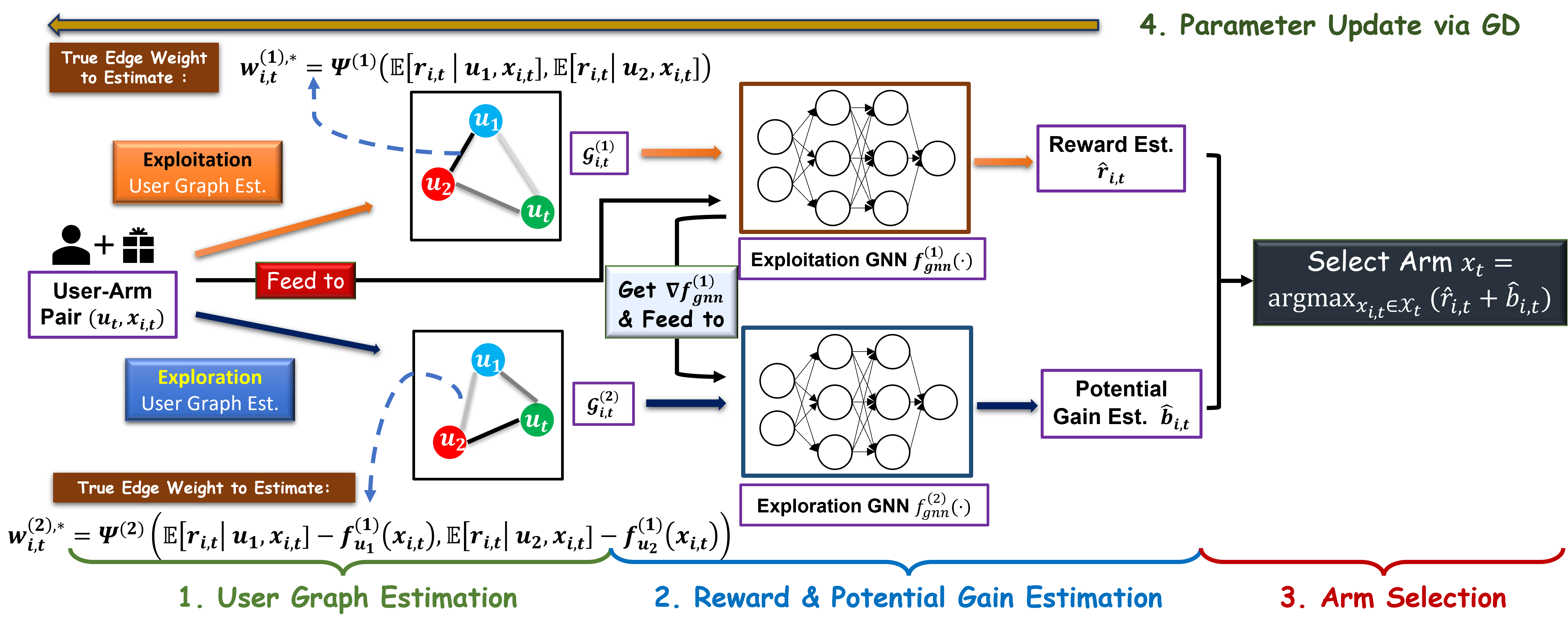}
  \vspace{-0.45cm}
  \caption{ Workflow of the proposed Graph Neural Bandits (\name) framework. }
  \vspace{-0.45cm}
  \label{fig_model_arch}
\end{figure*}

\vspace{-0.1cm}

\begin{algorithm}[t]
\caption{Graph Neural Bandits (\name)}
\label{algo_main}
\textbf{Input:} Number of rounds $T$, network width $m$, information propagation hops $k$. Functions for edge weight estimation $\Psi^{(1)}(\cdot, \cdot), \Psi^{(2)}(\cdot, \cdot): \mathbb{R} \times \mathbb{R} \mapsto \mathbb{R}$. \\
% arm group graph threshold $\beta$.\\ 
\textbf{Output:} Arm recommendation $\vect{x}_{t}$ for each time step $t$.\\
\textbf{Initialization:} Initialize trainable parameters for all models. \\

\For{$t \in \{1, 2, \dots, T\}$}{
    Receive a user $u_{t}$ and a set of arm contexts $\mathcal{X}_{t} = \{ \vect{x}_{i, t} \}_{i\in [a]}$. \\
    %
   % Construct two kinds of user graphs $\{\mathcal{G}_{i, t}^{(1)}\}_{i\in [a]}$, $\{\mathcal{G}_{i, t}^{(2)}\}_{i\in [a]} = $ \textbf{Procedure}~~ \textit{Estimating Arm-Specific User Graphs} $(\mathcal{X}_{t})$ [line 13-27], given the arm set $\mathcal{X}_{t}$. \\
    %
    \For{each candidate arm $\vect{x}_{i, t} \in \mathcal{X}_{t}$}{
        Construct two kinds of user graphs $\mathcal{G}_{i, t}^{(1)}$, $\mathcal{G}_{i, t}^{(2)} = $ \textbf{Procedure}~~ \textit{Estimating Arm-Specific User Graphs} $(\vect{x}_{i, t})$. [line 13-20] \\
        
        %%%%%%%%%%%%%
        Compute reward estimation [\textbf{Eq.} \ref{eq_f_1_user_output}]
        $\hat{r}_{i, t} = f_{gnn}^{(1)}(\vect{x}_{i, t},~\mathcal{G}_{i, t}^{(1)}; [\matr{\Theta}_{gnn}^{(1)}]_{t-1})$, and the potential gain [\textbf{Eq.} \ref{eq_f_2_explore_user_output}]
        $\hat{b}_{i, t} = f_{gnn}^{(2)}(\nabla [f_{gnn}^{(1)}]_{i, t} ,~\mathcal{G}_{i, t}^{(2)}; [\matr{\Theta}_{gnn}^{(2)}]_{t-1})$. \\
    }
    Play arm  
    $\vect{x}_{t} = \arg\max_{\vect{x}_{i, t} \in \mathcal{X}_{t}} \big( \hat{r}_{i, t} + \hat{b}_{i, t} \big)$, and observe its true reward $r_{t}$. \\
    Train the user networks $f_{u_{t}}^{(1)}(\cdot; \matr{\Theta}_{u_{t}}^{(1)})$, $f_{u_{t}}^{(2)}(\cdot; \matr{\Theta}_{u_{t}}^{(2)})$ and GNN models 
    $f_{gnn}^{(1)}(\cdot; \matr{\Theta}_{gnn}^{(1)})$, $f_{gnn}^{(2)}(\cdot; \matr{\Theta}_{gnn}^{(2)})$  with GD.  
    % (\textbf{Algorithm} \ref{algo_training}, Appendix Section \ref{sec_appd_pseudo_code}).
    % \ChangedForCamera
}

%%%%%%%%%%%%%%%%%%%%%%%%%%%%%%%%%%%%%%%%%%%%%%%%%%%%%%
%%%%%%%%%%%%%%%%%%%%%%%%%%%%%%%%%%%%%%%%%%%%%%%%%%%%%%
% \SetKwProg{myproc}{Procedure}{}{end procedure}
% \myproc{Estimating Arm-Specific User Graphs $(\mathcal{X}_{t})$}{
%     Initialize $\{\mathcal{G}_{i, t}^{(1)}\}_{i\in [a]}$, $\{\mathcal{G}_{i, t}^{(2)}\}_{i\in [a]}$. \\
%     \For{each user $u\in\mathcal{U}$}{
%         %
%         \For{each arm $\vect{x}_{i, t} \in \mathcal{X}_{t}, i\in [a]$}{
%             Compute $\hat{r}_{u, i} = f_{u}^{(1)}(\vect{x}_{i, t}; [\matr{\Theta}_{u}^{(1)}]_{t-1})$, and 
%             $\hat{b}_{u, i} = f_{u}^{(2)}(\nabla_{\matr{\Theta}_{u}^{(1)}} f_{u}^{(1)}(\vect{x}_{i, t}; [\matr{\Theta}_{u}^{(1)}]_{t-1}); [\matr{\Theta}_{u}^{(2)}]_{t-1})$. \\
%         }
%         %
%     }
%     \For{each arm $\vect{x}_{i, t}\in \mathcal{X}_{t}$}{
%         \For{each user pair $(u, u') \in\mathcal{U}\times \mathcal{U}$}{
%             For edge weight $w_{i, t}^{(1)}(u, u') \in W_{i, t}^{(1)}$, update $w_{i, t}^{(1)}(u, u') = \Psi^{(1)}(\hat{r}_{u, i}, \hat{r}_{u', i})$. \\
%             For edge weight $w_{i, t}^{(2)}(u, u') \in W_{i, t}^{(2)}$, update $w_{i, t}^{(2)}(u, u') = \Psi^{(2)}(\hat{b}_{u, i}, \hat{b}_{u', i})$. \\
%         }
%     }
%     Return user graphs $\{\mathcal{G}_{i, t}^{(1)}\}_{i\in [a]}$, $\{\mathcal{G}_{i, t}^{(2)}\}_{i\in [a]}$. \\
% }

\SetKwProg{myproc}{Procedure}{}{end procedure}
\myproc{Estimating Arm-Specific User Graphs $(\vect{x}_{i, t})$}{
    Initialize arm graphs $\mathcal{G}_{i, t}^{(1)}$, $\mathcal{G}_{i, t}^{(2)}$. \\
    
    \For{each user pair $(u, u') \in\mathcal{U}\times \mathcal{U}$}{
        % Compute reward estimations $\hat{r}_{u, i} = f_{u}^{(1)}(\vect{x}_{i, t}; [\matr{\Theta}_{u}^{(1)}]_{t-1})$, $\hat{r}_{u', i} = f_{u'}^{(1)}(\vect{x}_{i, t}; [\matr{\Theta}_{u'}^{(1)}]_{t-1})$ and the potential gain
        %     $\hat{b}_{u, i} = f_{u}^{(2)}(\nabla f_{u}^{(1)}(\vect{x}_{i, t}); [\matr{\Theta}_{u}^{(2)}]_{t-1})$, $\hat{b}_{u', i} = f_{u'}^{(2)}(\nabla f_{u'}^{(1)}(\vect{x}_{i, t}); [\matr{\Theta}_{u'}^{(2)}]_{t-1})$. \\
        %
        For edge weight $w_{i, t}^{(1)}(u, u') \in W_{i, t}^{(1)}$, update $w_{i, t}^{(1)}(u, u') = \Psi^{(1)} \big( f_{u}^{(1)}(\vect{x}_{i, t}), f_{u'}^{(1)}(\vect{x}_{i, t}) \big)$. [\textbf{Eq.} \ref{eq_est_exploit_correlation}] \\
        For edge weight $w_{i, t}^{(2)}(u, u') \in W_{i, t}^{(2)}$, based on the [\textbf{Eq.} \ref{eq_est_explore_correlation}], update $w_{i, t}^{(2)}(u, u') = \Psi^{(2)} \big( f_{u}^{(2)}\big( \nabla f_{u}^{(1)}(\vect{x}_{i, t}) ), f_{u'}^{(2)}\big( \nabla f_{u'}^{(1)}(\vect{x}_{i, t}) \big) \big)$. \\
    }
    Return user graphs $\mathcal{G}_{i, t}^{(1)}$, $\mathcal{G}_{i, t}^{(2)}$. \\
}

\end{algorithm}

\vspace{-0.1cm}
% ====================================================

% \vspace{-0.3cm}
\section{GNB: Proposed Framework}   \label{sec_proposed_framework}
% \vspace{-0.2cm}

The workflow of our proposed \name\ framework (Figure \ref{fig_model_arch}) consists of four major components: 
\textbf{First}, we derive the estimation for user exploitation graphs $\mathcal{G}_{i, t}^{(1), *}, i\in [a]$ (denoted by $\mathcal{G}_{i, t}^{(1)}$), and the exploration graphs $\mathcal{G}_{i, t}^{(2), *}, i\in [a]$ (denoted by $\mathcal{G}_{i, t}^{(2)}$), 
to model user correlations in terms of exploitation and exploration respectively; 
\textbf{Second}, to estimate the reward and the potential gain by leveraging the "fine-grained" correlations, we propose the GNN-based models  $f_{gnn}^{(1)}(\cdot)$, $f_{gnn}^{(2)}(\cdot)$  to aggregate the correlations of the target user-arm pair on estimated graphs $\mathcal{G}_{i, t}^{(1)} $ and  $\mathcal{G}_{i, t}^{(2)}$, respectively;
\textbf{Third}, we select the arm $\vect{x}_{t}$, based on the estimated arm reward and potential gain calculated by our GNN-based models; 
\textbf{Finally}, we train the parameters of \name\ using gradient descent (GD) on past records.

% since the model training with GD is standard knowledge, 
% Here, we choose to include all the information for the training process (e.g., training samples, labels, and the loss function) in the main body, and leave the pseudo-code summarizing the training procedure to Appendix Section \ref{sec_appd_pseudo_code} (\textbf{Alg.} \ref{algo_training}) due to page limit.
% \ChangedForCamera

% --------------
\vspace{-0.1cm}
\subsection{User Graph Estimation with User Networks} \label{subsec_user_models} 
Based on the definition of unknown true user graphs $\mathcal{G}^{(1), *}_{i, t}$, $\mathcal{G}^{(2), *}_{i, t}$ w.r.t. arm $\vect{x}_{i, t}\in \mathcal{X}_{t}$ (\textbf{Definition} \ref{def_exploitation_simi} and \ref{def_exploration_simi}), we proceed to derive their estimations $\mathcal{G}^{(1)}_{i, t}$, $\mathcal{G}^{(2)}_{i, t}, i\in [a]$ with individual user networks $f_{u}^{(1)}$, $f_{u}^{(2)}, u\in \mathcal{U}$. 
Afterwards, with these two kinds of estimated user graphs $\mathcal{G}_{i, t}^{(1)}$ and $\mathcal{G}_{i, t}^{(2)}$, 
we will be able to apply our GNN-based models to leverage the user correlations under the exploitation and the exploration settings. The pseudo-code is presented in \textbf{Alg.} \ref{algo_main}.

\textbf{[User Exploitation Network $f_{u}^{(1)}$]}
For each user $u\in\mathcal{U}$, we use a neural network $f_{u}^{(1)}(\cdot) = f_{u}^{(1)}(\cdot; \matr{\Theta}_{u}^{(1)})$ to learn user $u$'s preference for arm $\vect{x}_{i,t}$, i.e.,  $\mathbb{E}[r_{i,t} | u, \vect{x}_{i,t}]$. 
Following the \textbf{Def.} \ref{def_exploitation_simi}, we construct the exploitation graph $\mathcal{G}^{(1)}_{i, t}$ by estimating the user exploitation correlation based on user preferences.
Thus, in $\mathcal{G}^{(1)}_{i, t}$, we consider the edge weight of two user nodes $u, u'$ as
\begin{equation} 
    w_{i, t}^{(1)}(u, u') = \Psi^{(1)} \big( f_{u}^{(1)}(\vect{x}_{i, t}), f_{u'}^{(1)}(\vect{x}_{i, t}) \big)
\label{eq_est_exploit_correlation}
\end{equation}
where $\Psi^{(1)}(\cdot, \cdot)$ is the mapping function applied in \textbf{Def.} \ref{def_exploitation_simi} (line 16, \textbf{Alg.} \ref{algo_main}).
Here, $f_{u}^{(1)}(\cdot)$ will be trained by GD with chosen arms $\{\vect{x}_{\tau}\}_{\tau\in \mathcal{T}_{u, t}}$ as samples,
and received reward $\{ r_{\tau} \}_{\tau\in \mathcal{T}_{u, t}}$ as the labels, where 
$
\mathcal{L}_{u}^{(1)}(\Theta_{u}^{(1)}) = \sum_{\tau\in \mathcal{T}_{u, t}} \big| f_{u}^{(1)}(\vect{x}_{\tau}; 
    \matr{\Theta}_{u}^{(1)}) - r_{\tau}\big|^{2}   
$ will be the corresponding quadratic loss. Recall that $\vect{x}_{\tau}$ and $r_{\tau}$ stand for the chosen arm and the received reward respectively in round $\tau$.

\iffalse
\textbf{[User Exploitation Network $f_{u}^{(1)}$]}
For each user $u\in\mathcal{U}$, we propose to apply an exploitation network $f_{u}^{(1)}(\cdot)$ to learn user $u$'s preference for arm $\vect{x}_{i,t}$, i.e.,  $\mathbb{E}[r_{i,t} | u, \vect{x}_{i,t}]$. 
Here, $f_{u}^{(1)}(\cdot)$ will be trained with GD on the past records (arm contexts and rewards) $\mathcal{P}_{u, t}$ from user $u$, and the loss function will be the quadratic loss between the predicted reward and the actual reward.
%
Following the \textbf{Def.} \ref{def_exploitation_simi}, we will then be able to construct the exploitation graph $\mathcal{G}^{(1)}_{i, t}$ by estimating the user exploitation correlation based on user preferences.
In $\mathcal{G}^{(1)}_{i, t}$, we consider the edge weight between two user nodes $u, u'$ to be
$$
    w_{i, t}^{(1)}(u, u') = \Psi^{(1)} \big( f_{u}^{(1)}(\vect{x}_{i, t}), f_{u'}^{(1)}(\vect{x}_{i, t}) \big),
$$ 
where $\Psi^{(1)}(\cdot, \cdot)$ is the mapping function applied in \textbf{Def.} \ref{def_exploitation_simi} (line 16, \textbf{Alg.} \ref{algo_main}).
\fi
%
\textbf{[User Exploration Network $f_{u}^{(2)}$]}
Given user $u\in \mathcal{U}$, to estimate the potential gain (i.e., the uncertainty for the reward estimation) $\mathbb{E}[r | u, \vect{x}_{i,t}] - f_{u}^{(1)}(\vect{x}_{i,t})$ for arm $\vect{x}_{i,t}$, we adopt the user exploration network $f_{u}^{(2)}(\cdot) = f_{u}^{(2)}(\cdot; \matr{\Theta}_{u}^{(2)})$ inspired by \cite{EE-Net_ban2021ee}. 
As it has proved that the confidence interval (uncertainty) of reward estimation can be expressed as a function of network gradients \citep{Neural-UCB,AGG-UCB_qi2022neural}, we apply $f_{u}^{(2)}(\cdot)$ to directly learn the uncertainty with the gradient of $f_{u}^{(1)}(\cdot)$. Thus, the input of $f_{u}^{(2)}(\cdot)$ will be the network gradient of $f_{u}^{(1)}(\cdot)$ given arm $\vect{x}_{i,t}$, denoted as $\nabla f_{u}^{(1)}(\vect{x}_{i,t}) = \nabla_{\matr{\Theta}} f_{u}^{(1)} (\vect{x}_{i,t}; [\matr{\Theta}_{u}^{(1)}]_{t-1} )$, where $[\matr{\Theta}_{u}^{(1)}]_{t-1}$ refer to the parameters of $f_{u}^{(1)}$ in round $t$ (before training [line 11, \textbf{Alg.} \ref{algo_main}]).
Analogously, given the estimated user exploration graph $\mathcal{G}^{(2)}_{i, t}$ and two user nodes $u, u'$, we let the edge weight be 
\begin{equation} 
    w_{i, t}^{(2)}(u, u') = \Psi^{(2)} \bigg( f_{u}^{(2)}\big( \nabla f_{u}^{(1)}(\vect{x}_{i, t}) ), f_{u'}^{(2)}\big( \nabla f_{u'}^{(1)}(\vect{x}_{i, t}) \big) \bigg)
\label{eq_est_explore_correlation}
\end{equation}
% where $\nabla f_{u}^{(1)}(\vect{x}_{i, t})$ stands for the gradient of $f_{u}^{(1)}(\cdot)$ given arm $\vect{x}_{i, t}$ as the input
as in line 17, \textbf{Alg.} \ref{algo_main}, and $\Psi^{(2)}(\cdot, \cdot)$ is the mapping function that has been applied in \textbf{Def.} \ref{def_exploration_simi}. 
With GD, $f_{u}^{(2)}(\cdot)$ will be trained with the past gradients of $f_{u}^{(1)}$, i.e., $\{\nabla f_{u}^{(1)}(\vect{x}_{\tau})  \}_{\tau\in \mathcal{T}_{u, t}}$ as samples;
and the potential gain (uncertainty) $\{ r_{\tau} - f_{u}^{(1)}(\vect{x}_{\tau}; [\matr{\Theta}_{u}^{(1)}]_{\tau-1}) \}_{\tau\in \mathcal{T}_{u, t}}$ as labels. The quadratic loss is defined as 
$\mathcal{L}_{u}^{(2)}(\Theta_{u}^{(2)}) = \\
    \sum_{\tau\in \mathcal{T}_{u, t}} 
    \big|f_{u}^{(2)}( \nabla f_{u}^{(1)}(\vect{x}_{\tau}); \matr{\Theta}_{u}^{(2)}) - \big(r_{\tau} - f_{u}^{(1)}(\vect{x}_{\tau}; [\matr{\Theta}_{u}^{(1)}]_{\tau-1})\big)
    \big|^{2}.
$

\textbf{[Network Architecture]}
% For the theoretical analysis and experiments, we apply separate $L$-layer ($L \geq 2$) fully-connected (FC) networks for both kinds of user networks.
% %
% Their weight matrix entries for the first $L-1$ layers are drawn from the Gaussian distribution $N(0, 2 / m)$. The entries of the last layer ($L$-th layer) are sampled from $N(0, 1 / m)$. Complementary details are in Appendix Section \ref{sec_appd_user_network_arch}.
Here, we can apply various architectures for $f_{u}^{(1)}(\cdot), f_{u}^{(2)}(\cdot)$ to deal with different application scenarios (e.g., Convolutional Neural Networks [CNNs] for visual content recommendation tasks). For the theoretical analysis and experiments, with user $u\in \mathcal{U}$, we apply separate $L$-layer ($L \geq 2$) fully-connected (FC) networks as the user exploitation and exploration network  
\begin{equation} 
f_{u}(\vect{\chi}; \matr{\Theta}_{u} ) = \matr{\Theta}_{L} \sigma ( \matr{\Theta}_{L-1}  \sigma (\matr{\Theta}_{L-2} \dots  \sigma(\matr{\Theta}_{1} \vect{\chi}) )), ~~ \sigma := \text{ReLU}(\cdot) 
\label{eq_user_model_structure}
\end{equation}
with $\matr{\Theta}_{u} = [\text{vec}(\matr{\Theta}_{1})^{\intercal}, \dots, \text{vec}(\matr{\Theta}_{L})^{\intercal}]^{\intercal}$ being the vector of trainable parameters.
Here, since $f_{u}^{(1)}(\cdot), f_{u}^{(2)} (\cdot)$ are both the $L$-layer FC network  (\textbf{Eq.} \ref{eq_user_model_structure}), the input $\vect{\chi}$ can be substituted with either the arm context $\vect{x}_{i, t}$ or the network gradient $\nabla f_{u}^{(1)}(\vect{x}_{i, t})$ accordingly. 

\textbf{[Parameter Initialization]}
The weight matrices of the first layer are slightly different for two kinds of user networks, as $\matr{\Theta}_{1}^{(1)} \in \mathbb{R}^{m\times d}$, $\matr{\Theta}_{1}^{(2)} \in \mathbb{R}^{m\times p_{u}^{(1)}}$ where $p_{u}^{(1)}$ is the dimensionality of $\matr{\Theta}_{u}^{(1)}$. The weight matrix shape for the rest of the $L-1$ layers will be the same for these two kinds of user networks, which are $\matr{\Theta}_{l} \in \mathbb{R}^{m\times m}, l\in [2, \cdots, L-1]$, and $\matr{\Theta}_{L} \in \mathbb{R}^{1\times m}$. 
To initialize $f_{u}^{(1)}, f_{u}^{(2)}$, the weight matrix entries for their first $L-1$ layers $\{\matr{\Theta}_{1}, \dots \matr{\Theta}_{L-1} \}$ are drawn from the Gaussian distribution $N(0, 2 / m)$, and the entries of the last layer weight matrix $\matr{\Theta}_{L}$ are sampled from $N(0, 1 / m)$.

\subsection{Achieving Exploitation and Exploration with GNN Models on Estimated User Graphs}
% \vspace{-0.1cm}

With derived user exploitation graphs $\mathcal{G}_{i, t}^{(1)}$, and exploitation graphs $\mathcal{G}_{i, t}^{(2)}, i\in [a]$, we apply two GNN models to separately estimate the arm reward and potential gain for a refined arm selection strategy, by utilizing the past interaction records with all the users. 

% --------------
% \vspace{-0.1cm}
\subsubsection{The Exploitation GNN $f_{gnn}^{(1)}(\cdot)$}     \label{subsec_GNN_arch}
% \vspace{-0.05cm}
In round $t$, with the estimated user exploitation graph $\mathcal{G}_{i, t}^{(1)}$ for arm $\vect{x}_{i, t} \in \mathcal{X}_{t}$, we apply the exploitation GNN model $f_{gnn}^{(1)}(\vect{x}_{i, t}, \mathcal{G}_{i, t}^{(1)}; \Theta_{gnn}^{(1)})$ to collaboratively estimate the arm reward $\widehat{r}_{i, t}$ for the received user $u_{t} \in \mathcal{U}$. We start from learning the aggregated representation for $k$ hops, as
\begin{equation}
\begin{split}
    \matr{H}_{agg} = \sigma\big((\matr{S}_{i, t}^{(1)})^{k} \cdot (\matr{X}_{i, t} \matr{\Theta}_{agg}^{(1)})\big) \in \mathbb{R}^{n \times m}
\end{split}
\label{eq_GNN_aggegation}
\end{equation}
where 
$\matr{S}_{i, t}^{(1)} = (\matr{D}_{i, t}^{(1)})^{-\frac{1}{2}} \matr{A}_{i, t}^{(1)} (\matr{D}_{i, t}^{(1)})^{-\frac{1}{2}}$ is the symmetrically normalized adjacency matrix of $\mathcal{G}_{i, t}^{(1)}$, and $\sigma$ represents the ReLU activation function. 
With $m$ being the network width, we have $\matr{\Theta}_{agg}^{(1)} \in \mathbb{R}^{nd\times m}$ as the trainable weight matrix. After propagating the information for $k$ hops over the user graph, each row of $\matr{H}_{agg}$ corresponds to the aggregated $m$-dimensional hidden representation for one specific user-arm pair $(u, \vect{x}_{i, t}), u\in \mathcal{U}$. 
Here, the propagation of multi-hop information can provide a global perspective over the users, since it also involves the neighborhood information of users' neighbors \citep{consistency-2004,SGC_wu2019simplifying}.
To achieve this, we have the embedding matrix $\matr{X}_{i, t}$ (in \textbf{Eq.} \ref{eq_GNN_aggegation}) for arm $\vect{x}_{i, t} \in \mathcal{X}_{t}, i\in [a]$ being
\begin{equation}
\matr{X}_{i, t} = 
\left(\begin{array}{cccc}
\vect{x}_{i, t}^{\intercal} & \matr{0} & \cdots & \matr{0} \\
\matr{0} & \vect{x}_{i, t}^{\intercal} & \cdots & \matr{0} \\
\vdots  &       & \ddots   & \vdots \\
\matr{0} & \matr{0} & \cdots & \vect{x}_{i, t}^{\intercal} \\
\end{array} \right) \in \mathbb{R}^{n\times nd}
\label{eq_new_embedding_matrix}
\end{equation}
which partitions the weight matrix $\matr{\Theta}_{gnn}^{(1)}$ for different users. In this way, it is designed to generate individual representations w.r.t. each user-arm pair $(u, \vect{x}_{i, t}), u\in \mathcal{U}$ before the $k$-hop propagation (i.e., multiplying with $(\matr{S}_{i, t}^{(1)})^{k}$), which correspond to the rows of the matrix multiplication $(\matr{X}_{i, t} \matr{\Theta}_{agg}^{(1)}) \in \mathbb{R}^{n\times m}$.

Afterwards, with $\matr{H}_{0} = \matr{H}_{agg}$, we feed the aggregated representations into the $L$-layer ($L \geq 2$) FC network, represented by  
\begin{equation}
\begin{split}
    &\matr{H}_{l} = \sigma(\matr{H}_{l-1} \cdot \matr{\Theta}_{l}^{(1)}) \in \mathbb{R}^{n\times m} 
    ,~~ l \in [L-1], \\
    &\widehat{\vect{r}}_{all}(\vect{x}_{i, t}) = \matr{H}_{L-1} \cdot  \matr{\Theta}_{L}^{(1)} \in \mathbb{R}^{n}
\end{split}
\label{eq_GNN_estimation}
\end{equation}
where $\widehat{\vect{r}}_{all}(\vect{x}_{i, t}) \in \mathbb{R}^{n}$ represents the reward estimation for all the users in $\mathcal{U}$, given the arm $\vect{x}_{i, t}$. Given the target user $u_{t}$ in round $t$, the reward estimation for the user-arm pair $(u_{t}, \vect{x}_{i, t})$ would be the corresponding element in $\widehat{\vect{r}}_{all}$ (line 8, \textbf{Alg.} \ref{algo_main}), represented by: 
\begin{equation}
\begin{split}
    \widehat{r}_{i, t} = f_{gnn}^{(1)}(\vect{x}_{i, t},~\mathcal{G}_{i, t}^{(1)}; [\matr{\Theta}_{gnn}^{(1)}]_{t-1}) = [\widehat{\vect{r}}_{all}(\vect{x}_{i, t})]_{u_{t}}
\end{split}
\label{eq_f_1_user_output}
\end{equation}
where $\matr{\Theta}_{gnn}^{(1)} = [ \text{vec}(\matr{\Theta}_{agg}^{(1)})^{\intercal}, \text{vec}(\matr{\Theta}_{1}^{(1)})^{\intercal}, \dots, \text{vec}(\matr{\Theta}_{L}^{(1)})^{\intercal} ]^{\intercal} \in \mathbb{R}^{p}$ represent the trainable parameters of the exploitation GNN model, and we have $[\matr{\Theta}_{gnn}^{(1)}]_{t-1}$ being the parameters $\matr{\Theta}_{gnn}^{(1)}$ in round $t$ (before training [line 11, \textbf{Alg.} \ref{algo_main}]).
Here, the weight matrix shapes are $\matr{\Theta}_{l}^{(1)} \in \mathbb{R}^{m\times m}, l\in [1, \cdots, L-1]$, and the $L$-th layer $\matr{\Theta}_{L}^{(1)} \in \mathbb{R}^{m}$.
%
% Meanwhile, we use $\matr{\Theta}_{gnn}^{(1)} = \{\matr{\Theta}_{agg}^{(1)}, \matr{\Theta}_{1}^{(1)}, \dots, \matr{\Theta}_{L}^{(1)} \}$
% to represent the collection of trainable weight matrices for the GNN exploitation model.

\textbf{[Training $f_{gnn}^{(1)}$ with GD]}
The exploitation GNN $f_{gnn}^{(1)}(\cdot)$ will be trained with GD based on the received records $\mathcal{P}_{t}$. Then, we apply the quadratic loss function based on reward predictions $\{f_{gnn}^{(1)}(\vect{x}_{\tau},~\mathcal{G}_{\tau}^{(1)}; \matr{\Theta}_{gnn}^{(1)})\}_{\tau\in [t]}$ of chosen arms $\{\vect{x}_{\tau}\}_{\tau\in [t]}$, the actual received rewards $\{r_{\tau}\}_{\tau\in [t]}$, and the user exploitation graph $\mathcal{G}_{\tau}^{(1)}$ for chosen arms $\vect{x}_{\tau}, \tau\in [t]$. The  corresponding quadratic loss will be
$
    \mathcal{L}_{gnn}^{(1)}(\Theta_{gnn}^{(1)}) = \sum_{\tau\in [t]} \big|f_{gnn}^{(1)}(\vect{x}_{\tau}, \mathcal{G}_{\tau}^{(1)}; 
    \matr{\Theta}_{gnn}^{(1)}) - r_{\tau}\big|^{2}.
$

% ==========================
\textbf{[Connection with the Reward Function Definition (\textit{Eq.} \ref{eq_weighted_reward_func}) and Constraint (\textit{Eq.} \ref{eq_assumption_correlation_vector})]} 
The existing works (e.g.,\citep{conv_theory-allen2019convergence}) show that the FC network is naturally Lipschitz continuous with respect to the input when width $m$ is sufficiently large.
Thus, for \name, with aggregated hidden representations $\matr{H}_{agg}$ being the input to the FC network (\textbf{Eq.} \ref{eq_GNN_estimation}), we will have the difference of reward estimations $\widehat{r}_{i, t}$ bounded by the distance of rows in matrix $\matr{H}_{agg}$ (i.e., aggregated hidden representations). 
Here, given $\vect{x}_{i, t}\in \mathcal{X}_{t}$ and users $u_{1}, u_{2}\in \mathcal{U}$, their estimated reward difference $\abs{[\widehat{\vect{r}}_{all}(\vect{x}_{i, t})]_{u_{1}} - [\widehat{\vect{r}}_{all}(\vect{x}_{i, t})]_{u_{2}}}$ can be bounded by the distance of the corresponding rows in $\matr{S}_{i, t}$ (i.e., $\norm{\matr{S}_{i, t}^{(1)}[u_{1}:]-\matr{S}_{i, t}^{(1)}[u_{2}:]}$) given the exploitation GNN model.
This design matches our definition and the constraint in \textbf{Eq.} \ref{eq_weighted_reward_func}-\ref{eq_assumption_correlation_vector}. 

% ==========================
\subsubsection{The Exploration GNN $f_{gnn}^{(2)}(\cdot)$}
Given a candidate arm $\vect{x}_{i, t}\in \mathcal{X}_{t}$, to achieve adaptive exploration with the user exploration collaborations encoded in $\mathcal{G}^{(2)}_{i, t}$, we apply a second GNN model 
$f_{gnn}^{(2)}(\cdot)$ to evaluate the potential gain $\widehat{b}_{i, t}$ for the reward estimation $\widehat{r}_{i, t} = f_{gnn}^{(1)}(\vect{x}_{i, t},~\mathcal{G}_{i, t}^{(1)}; [\matr{\Theta}_{gnn}^{(1)}]_{t-1})$ [\textbf{Eq.} \ref{eq_f_1_user_output}], denoted by
\begin{equation}
\begin{split}
    \widehat{b}_{i, t} = f_{gnn}^{(2)}( \nabla [f_{gnn}^{(1)}]_{i, t} , \mathcal{G}_{i, t}^{(2)};  [\matr{\Theta}_{gnn}^{(2)}]_{t-1}) 
    = [\widehat{\vect{b}}_{all}(\vect{x}_{i, t})]_{u_{t}}.
\end{split}
\label{eq_f_2_explore_user_output}
\end{equation}
The architecture of $f_{gnn}^{(2)}(\cdot)$ can also be represented by \textbf{Eq.} \ref{eq_GNN_aggegation}-\ref{eq_f_1_user_output}. While $f_{gnn}^{(1)}(\cdot), f_{gnn}^{(2)}(\cdot)$ have the same network width $m$ and number of layers $L$, the dimensionality of $\matr{\Theta}_{agg}^{(1)} \in \mathbb{R}^{nd\times m}, \matr{\Theta}_{agg}^{(2)}\in \mathbb{R}^{np\times m}$ is different. Analogously, $\widehat{\vect{b}}_{all}(\vect{x}_{i, t}) \in \mathbb{R}^{n}$ is the potential gain estimation for all the users in $\mathcal{U}$, w.r.t.
arm $\vect{x}_{i, t}$ and the exploitation GNN $f_{gnn}^{(1)}(\cdot)$.
Here, the inputs are user exploration graph $\mathcal{G}_{i, t}^{(2)}$, and the gradient of the exploitation GNN, represented by $\nabla [f_{gnn}^{(1)}]_{i, t} = \nabla_{ \matr{\Theta}_{gnn}^{(1)}} f_{gnn}^{(1)}(\vect{x}_{i, t}, \mathcal{G}_{i, t}^{(1)} ; [\matr{\Theta}_{gnn}^{(1)}]_{t-1} )$.
The exploration GNN $f_{gnn}^{(2)}(\cdot)$ leverages the user exploration graph $\mathcal{G}_{i, t}^{(2)}$ and the gradients of $f_{gnn}^{(1)}(\cdot)$ to estimate the uncertainty of reward estimations, which stands for our adaptive exploration strategy (downward or upward exploration). More discussions are in Appendix \textbf{Section} \ref{sec_appd_adaptive_exp}.
% \TODO{Update here}

%
\textbf{[Training $f_{gnn}^{(2)}$ with GD]}
% Analogous to previous $f_{gnn}^{(1)}(\cdot)$, the architecture of $f_{gnn}^{(2)}(\cdot)$ can also be represented by \textbf{Eq.} \ref{eq_GNN_aggegation}-\textbf{Eq.} \ref{eq_f_1_user_output}.
%
% Note that while $f_{gnn}^{(1)}(\cdot), f_{gnn}^{(2)}(\cdot)$ have the same network width $m$ and number of layers $L$, the dimensionality of $\matr{\Theta}_{agg}^{(1)} \in \mathbb{R}^{nd\times m}, \matr{\Theta}_{agg}^{(2)}\in \mathbb{R}^{np\times m}$ is different. 
%
Similar to $f_{gnn}^{(1)}$, we train $f_{gnn}^{(2)}$ with GD by minimizing the quadratic loss, denoted by $\mathcal{L}_{gnn}^{(2)}(\Theta_{gnn}^{(2)}) = \sum_{\tau\in [t]} $
$
    \big|f_{gnn}^{(2)}(\nabla [f_{gnn}^{(1)}]_{\tau}, \mathcal{G}_{\tau}^{(2)}; \matr{\Theta}_{gnn}^{(2)}) - 
    \big(r_{\tau} - f_{gnn}^{(1)}(\vect{x}_{\tau}, \mathcal{G}_{\tau}^{(1)}; [\matr{\Theta}_{gnn}^{(1)}]_{\tau-1})\big)
    \big|^{2}.
$
This is defined to measure the difference between the estimated potential gains $\{f_{gnn}^{(2)}( \nabla [f_{gnn}^{(1)}]_{\tau} , \mathcal{G}_{\tau}^{(2)};  \matr{\Theta}_{gnn}^{(2)})\}_{\tau\in [t]}$, and the corresponding labels $\{r_{\tau} - f_{gnn}^{(1)}(\vect{x}_{\tau},~\mathcal{G}_{\tau}^{(1)}; [\matr{\Theta}_{gnn}^{(1)}]_{\tau-1})\}_{\tau\in [t]}$.

% --------------------------------------------
% Remark gradient dimension reduction
\vspace{-0.1cm}
\begin{remark}[Reducing Input Complexity]
    The input of $f_{gnn}^{(2)}(\cdot)$ is the gradient $\nabla_{\matr{\Theta}} f_{gnn}^{(1)}(\vect{x})$ given the arm $\vect{x}$, and its dimensionality is naturally $p = (nd\times m) + (L-1)\times m^{2} + m$, which can be a large number. Inspired by CNNs, e.g., \cite{CNN_avg_pooling_radenovic2018fine}, we apply the \textbf{average pooling} to approximate the original gradient vector in practice. In this way, we can save the running time and reduce space complexity simultaneously. Note this approach is also compatible with user networks in Subsection \ref{subsec_user_models}. To prove its effectiveness, we will apply this approach on \name~for all the experiments in Section \ref{sec_experiments}.
\label{remark_avg_pool}
\end{remark}

% --------------------------------------------
% Remark large number of users
\vspace{-0.1cm}
\begin{remark}[Working with Large Systems]
    When facing a large number of users, we can apply the ``approximated user neighborhood'' to reduce the running time in practice. 
    Given user graphs $\mathcal{G}_{i, t}^{(1)}, \mathcal{G}_{i, t}^{(2)}$ in terms of arm $\vect{x}_{i, t}$, we derive approximated user neighborhoods $\widetilde{\mathcal{N}}^{(1)}(u_{t}),~ \widetilde{\mathcal{N}}^{(2)}(u_{t}) \subset \mathcal{U}$ for the target user $u_{t}$, with size $\abs{\widetilde{\mathcal{N}}^{(1)}(u_{t})} = \abs{\widetilde{\mathcal{N}}^{(2)}(u_{t})} = \widetilde{n}$,  where $\widetilde{n} << n$.
    For instance, we can choose $\widetilde{n}$ ``representative users'' (e.g., users posting high-quality reviews on e-commerce platforms) to form $\widetilde{\mathcal{N}}^{(1)}(u_{t}), \widetilde{\mathcal{N}}^{(2)}(u_{t})$, and apply the corresponding approximated user sub-graphs for downstream GNN models to reduce the computation cost and space cost in practice. Related experiments are provided in Subsection \ref{subsec_exp_approx_neighborhood}.
    % Here, we also include complementary discussions for the space and time complexity in Appendix Section \ref{sec_appx_additional_discussion_on_time_space_complexity}.
    % \ChangedForCamera
\label{remark_numerous_users}
\end{remark}
\vspace{-0.1cm}

% --------------
\textbf{[Parameter Initialization]}
For the parameters of both GNN models (i.e., $\matr{\Theta}_{gnn}^{(1)}$ and $\matr{\Theta}_{gnn}^{(2)}$), the matrix entries of the aggregation weight matrix $\matr{\Theta}_{agg}$ and the first $L-1$ FC layers $\{\matr{\Theta}_{1}, \dots, \matr{\Theta}_{L-1} \}$ are drawn from the Gaussian distribution $N(0, 2 / m)$. Then, for the last layer weight matrix $\matr{\Theta}_{L}$, we draw its entries from $N(0, 1 / m)$.

% --------------------------------------------

\subsubsection{Arm Selection Mechanism and Model Training}
In round $t$, with the current parameters $[\Theta_{gnn}^{(1)}]_{t-1}, [\Theta_{gnn}^{(2)}]_{t-1}$ for GNN models before model training, the selected arm is chosen as 
\begin{displaymath}
\begin{split}
    \vect{x}_{t} = & \arg\max_{\vect{x}_{i, t} \in \mathcal{X}_{t}} 
    \bigg[ f_{gnn}^{(1)}(\vect{x}_{i, t},~\mathcal{G}_{i, t}^{(1)}; [\Theta_{gnn}^{(1)}]_{t-1}) \\
    & + f_{gnn}^{(2)}(\nabla_{\Theta_{gnn}^{(1)}} f_{gnn}^{(1)}(\vect{x}_{i, t},~\mathcal{G}_{i, t}^{(1)}; [\Theta_{gnn}^{(1)}]_{t-1}),~\mathcal{G}_{i, t}^{(2)}; [\Theta_{gnn}^{(2)}]_{t-1}) \bigg]
\end{split}
\end{displaymath}
based on the estimated reward and potential gain (line 10, \textbf{Alg.} \ref{algo_main}). 
After receiving reward $r_{t}$, we update user networks $f_{u_{t}}^{(1)}, f_{u_{t}}^{(2)}$ of user $u_{t}$, and GNN models based on GD (line 11, \textbf{Alg.} \ref{algo_main}).
% and \textbf{Alg.} \ref{algo_training} in Appendix Section \ref{sec_appd_pseudo_code}).
% \ChangedForCamera

\vspace{-0.1cm}
% ====================================================
% \vspace{-0.2cm}

\section{Theoretical Analysis}    \label{sec_theoretical_analysis}

In this section, we present the theoretical analysis for the proposed \name.
% To model the user behaviors, we adopt two $L$-layer FC networks with the input $\vect{\chi}$ for user networks $f_{u}^{(1)}(\cdot), f_{u}^{(2)}(\cdot)$ respectively as 
% \begin{equation} 
% f_{u}(\vect{\chi}; \matr{\Theta}_{u} ) = \matr{\Theta}_{L} \sigma ( \matr{\Theta}_{L-1}  \sigma (\matr{\Theta}_{L-2} \dots  \sigma(\matr{\Theta}_{1} \vect{\chi}) ))
% \label{eq_user_model_structure}
% \end{equation}
% with $\matr{\Theta}_{u} = \{\matr{\Theta}_{1}, \dots, \matr{\Theta}_{L}\}$ and $\sigma$ being the ReLU activation. 
% Here, since $f_{u}^{(1)}(\cdot), f_{u}^{(2)}(\cdot)$ are both $L$-layer networks shown in \textbf{Eq.}\ref{eq_user_model_structure}, the input $\vect{\chi}$ can be either the arm $\vect{x}$ or the network gradient $\nabla_{\matr{\Theta}_{u}^{(1)}}f_{u}^{(1)}(\cdot; \matr{\Theta}_{u}^{(1)})$.
% Then, the weight matrix of the input layer is different for two user networks where $\matr{\Theta}_{1}^{(1)} \in \mathbb{R}^{m\times d}$ and $\matr{\Theta}_{1}^{(2)} \in \mathbb{R}^{m\times p}$. The rest of the layers will be the same comparing the two kinds of user networks, which are $\matr{\Theta}_{l} \in \mathbb{R}^{m\times m}, l\in [2, \cdots, L-1]$, and $\matr{\Theta}_{L} \in \mathbb{R}^{1\times m}$. 
% Comparably, for both user networks, the entries of $\{\matr{\Theta}_{1}, \dots \matr{\Theta}_{L-1} \}$ are drawn from the Gaussian distribution $N(0, 2 / m)$. The entries of the last layer $\matr{\Theta}_{L}$ are sampled from $N(0, 1 / m)$.
%
Here, we consider each user $u\in \mathcal{U}$ to be evenly served $T / n$ rounds up to time step $T$, i.e., $\abs{\mathcal{T}_{u, t}} = T_{u, t} = T / n$, which is standard in closely related works 
(e.g., \cite{club_2014,local_clustering-ban2021local}). 
To ensure the neural models are able to efficiently learn the underlying reward mapping, we have the following assumption regarding the arm separateness.

\begin{assumption} [$\rho$-Separateness of Arms]  \label{assumption_separateness}
After a total of $~T$ rounds, for every pair $\vect{x}_{i, t}, \vect{x}_{i', t'}$ with $t, t' \in[T]$ and $i, i' \in [a]$,  if $(t, i) \neq (t', i')$, we have $\norm{\vect{x}_{i, t} - \vect{x}_{i', t'}}_{2} \geq \rho$ where $0 < \rho \leq \mathcal{O}(\frac{1}{L})$.
\end{assumption}
% \begin{assumption} [$\rho$-Separateness of Arms]  \label{assumption_separateness}
% After a total of $~T$ rounds, for every pair of chosen arms $(\vect{x}_{t}, \vect{x}_{t'})$ with $t, t' \in[T]$ ,  if $t \neq t'$, we have $\norm{\vect{x}_{t} - \vect{x}_{t'}}_{2} \geq \rho$ where $0 < \rho \leq \mathcal{O}(\frac{1}{L})$.
% \end{assumption}
%
Note that the above assumption is mild, and it has been commonly applied in existing works on neural bandits \citep{EE-Net_ban2021ee} and over-parameterized neural networks \citep{conv_theory-allen2019convergence}. 
Since $L$ can be manually set (e.g., $L = 2$), we can easily satisfy the condition $ 0 < \rho \leq \mathcal{O}(\frac{1}{L})$ as long as no two arms are identical. 
Meanwhile, Assumption 4.2 in \cite{Neural-UCB} and Assumption 3.4 from \cite{neural_thompson-zhang2020neural} also imply that no two arms are the same, and they measure the arm separateness in terms of the minimum eigenvalue $\lambda_{0}$ (with $\lambda_{0} > 0$) of the Neural Tangent Kernel (NTK) \citep{NTK_jacot2018neural} matrix, which is comparable with our Euclidean separateness $\rho$.
%Here, as long as we have a finite number of rounds $T$ and no duplicate arms, there will exist $\rho > 0$ such that the arms received will satisfy the separateness condition. 
%
Based on \textbf{Def.} \ref{def_exploitation_simi} and \textbf{Def.} \ref{def_exploration_simi}, given an arm $\vect{x}_{i, t} \in \mathcal{X}_{t}$, we denote the adjacency matrices as $\matr{A}^{(1), *}_{i, t}$ and $\matr{A}^{(2), *}_{i, t}$ for the true arm graphs $\mathcal{G}^{(1), *}_{i, t}$, $\mathcal{G}^{(2), *}_{i, t}$.
For the sake of analysis, given any adjacency matrix $\matr{A}$, 
we derive the normalized adjacency matrix $\matr{S}$ by scaling the elements of $\matr{A}$ with $1 / n$. 
We also set the propagation parameter $k=1$, and define the mapping functions $\Psi^{(1)}(a, b), \Psi^{(2)}(a, b) := \exp( -\norm{ a - b } )$ given the inputs $a, b$. Note that our results can be readily generalized to other mapping functions with the Lipschitz-continuity properties.

Next, we proceed to show the regret bound $R(T)$ after $T$ time steps [\textbf{Eq.} \ref{eq_pseudo_regret}]. 
Here, the following \btheorem \ref{theorem_regret_bound} covers two types of error: (1) the estimation error of user graphs; and (2) the approximation error of neural models. 
Let $\eta_{1}, J_{1}$ be the learning rate and GD training iterations for user networks, and $\eta_{2}, J_{2}$ denote the learning rate and iterations for GNN models.
The proof sketch of \btheorem \ref{theorem_regret_bound} is presented in Appendix \textbf{Section} \ref{sec_appd_regre_bound_proof}.
\begin{theorem}[Regret Bound]
Define $\delta \in (0, 1)$, $0 < \xi_{1}, \xi_{2} \leq \mathcal{O}(1 / T)$ and $0 < \rho \leq \mathcal{O}(1 / L)$, $c_{\xi} > 0$, $\xi_{L} = (c_{\xi})^{L}$. 
With the user networks defined in \textbf{Eq.} \ref{eq_user_model_structure} and the GNN models defined in \textbf{Eq.} \ref{eq_GNN_aggegation}$-$\ref{eq_GNN_estimation} with $L$ FC-layers, let $m \geq \Omega \big( \text{Poly}(T, L, a, \frac{1}{\rho}) \cdot \xi_{L}\log(1 / \delta) \big) $, $n \geq \widetilde{\Omega} \big(\text{Poly}(L)\big)$.
% With the GD training process in \textbf{Algorithm} \ref{algo_training} (Appendix \textbf{Section} \ref{sec_appd_pseudo_code}), set parameters
% \ChangedForCamera
Set the learning rates and GD iterations
\vspace{-0.2cm}
\begin{displaymath}
\begin{split}
    & \eta_{1} = \Theta \big( \frac{\rho}{m\cdot \text{Poly}(T, n, a, L)} \big), \quad
    \eta_{2} = \Theta \big( \frac{\rho}{m\cdot \text{Poly}(T, a, L)} \big), \\
    & J_{1} = \Theta \big( \frac{\text{Poly}(T, n, a, L)}{\rho\cdot \delta^{2}} \cdot \log(\frac{1}{\xi_{1}}) \big), ~~
    J_{2} = \Theta \big( \frac{\text{Poly}(T, a, L)}{\rho\cdot \delta^{2}} \cdot \log(\frac{1}{\xi_{2}}) \big).
\end{split}
\end{displaymath}
Then, following \textbf{Algorithm} \ref{algo_main}, with probability at least $1 - \delta$, the $T$-round pseudo-regret $R(T)$ of \name\ can be bounded by
\vspace{-0.0cm}
\begin{displaymath}
\begin{split}
    R(T) \leq  \sqrt{T} \cdot  \big( 
        \mathcal{O}(L\xi_{L})  \cdot \sqrt{2\log(\frac{Tn\cdot a}{\delta})}
         \big) +  \sqrt{T}\cdot \mathcal{O}(L) + \mathcal{O}(\xi_{L}) + \mathcal{O}(1).
\end{split}
\end{displaymath}
% where $n$ is the number of users, and $\abs{\mathcal{X}_{t}} = a, \forall t\in [T]$ is the number of candidate arms for each round.
\label{theorem_regret_bound}
\end{theorem}
\vspace{-0.3cm}
Recall that $L$ is generally a small integer (e.g., we set $L=2$ for experiments in \textbf{Section} \ref{sec_experiments}), which makes the condition on number of users reasonable as $n$ is usually a gigantic number in real-world recommender systems. We also have $m$ to be sufficiently large under the over-parameterization regime, which makes the regret bound hold. Here, we have the following remarks.
\vspace{-0.15cm}
\begin{remark} [Dimension terms $d, \tilde{d}$]
    Existing neural single-bandit (i.e., with no user collaborations) algorithms \citep{Neural-UCB,neural_thompson-zhang2020neural,AGG-UCB_qi2022neural} keep the bound of  $\mathcal{O}(\Tilde{d} \sqrt{T}\log(T))$ based on gradient mappings and ridge regression. $\Tilde{d}$ is the effective dimension of the NTK matrix, which can grow along with the number of parameters $p$ and rounds $T$. 
    The linear user clustering algorithms (e.g., \cite{SCLUB_li2019improved,local_clustering-ban2021local,CAB_2017}) have the bound $\mathcal{O}(d \sqrt{T}\log(T))$ with context dimension $d$, which can be large with a high-dimensional context space. Alternatively, the regret bound in \textbf{Theorem} \ref{theorem_regret_bound} is free of terms $d$ and $\Tilde{d}$, as we apply the generalization bounds of over-parameterized networks instead \citep{conv_theory-allen2019convergence,generalization_bound_cao2019generalization}, which are unrelated to dimension terms $d$ or $\Tilde{d}$.
\label{remark_dimension_term}
\end{remark}

\vspace{-0.2cm}
\begin{remark} [From $\sqrt{n}$ to $\sqrt{\log(n)}$]
    % While our $\mathcal{O}(\sqrt{T\log(T)})$ bound matches theoretical bound of state-of-the-art EE-Net \citep{EE-Net_ban2021ee}, EE-Net only considers the single-bandit setting with no collaboration among users. 
    % Compared with Meta-Ban \citep{Meta-Ban}, we provide the theoretical analysis from a new perspective regarding the fine-grained user collaborative effect and GNNs. 
    With $n$ being the number of users, existing user clustering works (e.g., \cite{Meta-Ban,club_2014,SCLUB_li2019improved,local_clustering-ban2021local}) involve a $\sqrt{n}$ factor in the regret bound as the cost of leveraging user collaborative effects. Instead of applying separate estimators for each user group, our proposed \name\ only ends up with a $\sqrt{\log(n)}$ term to incorporate user collaborations by utilizing dual GNN models for estimating the arm rewards and potential gains correspondingly.
\label{remark_bound_sqrt_n}
\end{remark}

\vspace{-0.2cm}
\begin{remark} [Arm i.i.d. Assumption]
   Existing clustering of bandits algorithms (e.g., \cite{club_2014,SCLUB_li2019improved,CAB_2017,Meta-Ban}) and the single-bandit algorithm EE-Net \citep{EE-Net_ban2021ee} typically require the arm i.i.d. assumption for the theoretical analysis, which can be strong since the candidate arm pool $\mathcal{X}_{t}, t\in [T]$ is usually conditioned on the past records. Here, instead of using the regression-based analysis as in existing works, our proof of \btheorem \ref{theorem_regret_bound} applies the martingale-based analysis instead to help alleviate this concern.
\label{remark_bound_iid}
\end{remark}

\vspace{-0.1cm}
% ====================================================
% \vspace{-0.3cm}
\vspace{-0.2cm}
\section{Experiments}   \label{sec_experiments}

% \vspace{-0.1cm}

In this section, we evaluate the proposed \name\ framework on multiple real data sets against nine state-of-the-art algorithms, including
% \textbf{CLUB} \citep{club_2014}, \textbf{SCLUB} \citep{SCLUB_li2019improved}, \textbf{LOCB} \citep{local_clustering-ban2021local}, \textbf{DynUCB} \citep{Dyn-UCB_nguyen2014dynamic}, \textbf{COFIBA} \citep{co_filter_bandits_2016}, \textbf{Neural-UCB-Pool (Neural-Pool)} \citep{Neural-UCB}, \textbf{Neural-UCB-Ind (Neural-Ind)} \citep{Neural-UCB}, \textbf{EE-Net} \citep{EE-Net_ban2021ee}, and \textbf{Meta-Ban} \citep{Meta-Ban}.
% We will include the descriptions for the baselines in the Appendix Section \ref{sec_appd_experiments}.
% The descriptions for our nine baseline methods are:
the linear user clustering algorithms: (1) \textbf{CLUB} \citep{club_2014}, (2) \textbf{SCLUB} \citep{SCLUB_li2019improved}, (3) \textbf{LOCB} \citep{local_clustering-ban2021local}, (4) \textbf{DynUCB} \citep{Dyn-UCB_nguyen2014dynamic}, (5) \textbf{COFIBA} \citep{co_filter_bandits_2016}; the neural single-bandit algorithms: (6) \textbf{Neural-Pool} adopts one single Neural-UCB \citep{Neural-UCB} model for all the users with the UCB-type exploration strategy; (7) \textbf{Neural-Ind} assigns each user with their own separate Neural-UCB \citep{Neural-UCB} model; (8) \textbf{EE-Net} \citep{EE-Net_ban2021ee}; and, the neural user clustering algorithm: (9) \textbf{Meta-Ban} \citep{Meta-Ban}.
We leave the implementation details and data set URLs to Appendix \textbf{Section} \ref{sec_appd_experiments}.
% \begin{itemize}[leftmargin=*]
%     \item \textbf{CLUB} \citep{club_2014} regards connected components as user groups out of the estimated user graph, and adopts a UCB-type exploration strategy;
%     \item \textbf{SCLUB} \citep{SCLUB_li2019improved} estimates dynamic user sets as user groups, and allows set operations for group updates;
%     \item \textbf{LOCB} \citep{local_clustering-ban2021local} applies soft-clustering among users with random seeds and choose the best user group for reward and confidence bound estimations;
%     \item \textbf{DynUCB} \citep{Dyn-UCB_nguyen2014dynamic} dynamically assigns users to its nearest estimated cluster.
%     \item  \textbf{COFIBA} \citep{co_filter_bandits_2016} estimates user clustering and arm clustering simultaneously, and ensembles linear estimators for  reward and confidence bound estimations;
%     \item  \textbf{Neural-Pool} adopts one single Neural-UCB \citep{Neural-UCB} model for all the users with UCB-type exploration strategy;
%     \item  \textbf{Neural-Ind} assigns each user with their own separate Neural-UCB \citep{Neural-UCB} model;
%     \item  \textbf{EE-Net} \citep{EE-Net_ban2021ee} achieves adaptive exploration by applying additional neural models for the exploration and decision making;
%     \item  \textbf{Meta-Ban} \citep{Meta-Ban} utilizes individual neural models for each user's behavior, and applies a meta-model to adapt to estimated user groups.
% \end{itemize}

% % \vspace{-0.2cm}
% \vspace{-0.2cm}
\subsection{Real Data Sets}
In this section, we compare the proposed \name\ with baselines on six data sets with different specifications.

\textbf{[Recommendation Data Sets]}
% First, we conduct the experiments for two recommendation data sets with different specifications, which are the ``MovieLens'' data set and the ``Yelp'' data set. Given one user $u_{t}$ to serve in each round $t$, our goal is to recommend the optimal arm (movie / restaurant) from the candidate pool $\mathcal{X}_{t}$ to the user.
``MovieLens rating dataset'' includes reviews from $1.6 \times 10^5$ users towards $6 \times 10^4$ movies. 
Here, we select 10 genome-scores with the highest variance across movies to generate the movie features $\vect{v}_{i} \in \mathbb{R}^{d}, d=10$. The user features $\vect{v}_{u}\in\mathbb{R}^{d}, u\in \mathcal{U}$ are obtained through singular value decomposition (SVD) on the rating matrix. We use K-means to divide users into $n=50$ groups based on $\vect{v}_{u}$, and consider each group as a node in user graphs. 
In each round $t$, a user $u_t$ will be drawn from a randomly sampled group. 
For the candidate pool $\mathcal{X}_{t}$ with $\abs{\mathcal{X}_{t}}=a=10$ arms, we choose one bad movie ($\leq$ two stars, out of five) rated by $u_t$ with reward $1$, and randomly pick the other $9$ good movies with reward $0$. The target here is to help users avoid bad movies.
For ``Yelp'' data set, we build the rating matrix w.r.t. the top $2,000$ users and top $10,000$ arms with the most reviews.
Then, we use SVD to extract the $10$-dimensional representation for each user and restaurant.
For an arm, if the user's rating $\geq$ three stars (out of five stars), the reward is set to $1$; otherwise, the reward is $0$. 
Similarly, we apply K-means to obtain $n=50$ groups based on user features.
In round $t$, a target $u_t$, is sampled from a randomly selected group. 
For $\mathcal{X}_{t}$, we choose one good restaurant rated by $u_t$ with reward $1$, and randomly pick the other $9$ bad restaurants with reward $0$.

\textbf{[Classification Data Sets]}
We also perform experiments on four real classification data sets under the recommendation settings, 
which are ``MNIST'' (with the number of classes $\mathcal{C} = 10$), ``Shuttle'' ($\mathcal{C} = 7$), the ``Letter'' ($\mathcal{C} = 26$), and the ``Pendigits'' ($\mathcal{C} = 10$) data sets. Each class will correspond to one node in user graphs.
Similar to previous works \citep{Neural-UCB,Meta-Ban}, given a sample $\vect{x}\in \mathbb{R}^{d}$, we transform it into $\mathcal{C}$ different arms, denoted by $\vect{x}_1 = (\vect{x}, 0, \dots, 0), \vect{x}_2 = (0, \vect{x}, \dots, 0), \dots, \vect{x}_{\mathcal{C}} = (0, 0, \dots, \vect{x}) \in \mathbb{R}^{d+\mathcal{C}-1}$ where we add $\mathcal{C}-1$ zero digits as the padding. The received reward $r_{t}=1$ if we select the arm of the correct class, otherwise $r_{t}=0$.

\vspace{-0.2cm}
\subsubsection{Experiment Results}
\label{subsec_exp_results_main}
% \vspace{-0.1cm}

\begin{figure}[t!]
  \centering
  \vspace{-0.25cm}
  \includegraphics[width=\linewidth]{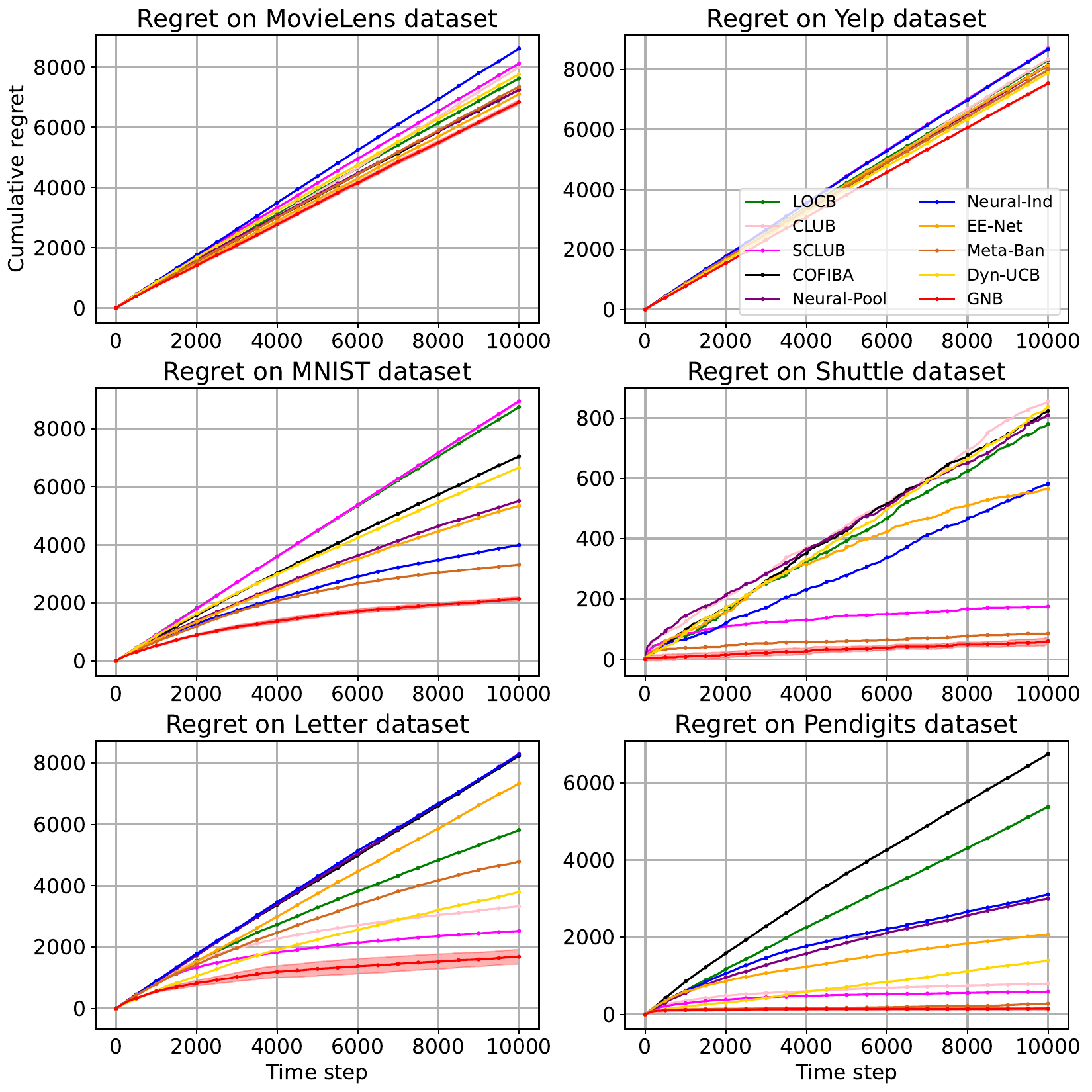}
  \vspace{-0.8cm}
  \caption{Cumulative regrets on the recommendation and classification data sets.}
  \vspace{-0.5cm}
  \label{fig_experiment_regret_results}
\end{figure}

Figure \ref{fig_experiment_regret_results} illustrates the cumulative regret results on the six data sets, and the red shade represents the standard deviation of \name. Here, our proposed \name\ manages to achieve the best performance against all these strong baselines. Since the MovieLens data set involves real arm features (i.e., genome-scores), the performance of different algorithms on the MovieLens data set tends to have larger divergence. 
Note that due to the inherent noise within these two recommendation data sets, we can observe the ``linear-like'' regret curves, which are common as in existing works (e.g., \cite{Meta-Ban}).
In this case, to show the model convergence, we will present the convergence results for the recommendation data sets in Appendix \textbf{Subsec.} \ref{subsec_appx_conv_GNN}. 
Among the baselines, the neural algorithms (Neural-Pool, EE-Net, Meta-Ban) generally perform better than linear algorithms due to the representation power of neural networks. However, as Neural-Ind considers no correlations among users, it tends to perform the worst among all baselines on these two data sets. 
For classification data sets, Meta-Ban performs better than the other baselines by modeling user (class) correlations with the neural network.
Since the classification data sets generally involve complex reward mapping functions, it can lead to the poor performances of linear algorithms. Our proposed \name\ outperforms the baselines by modeling fine-grained correlations and utilizing the adaptive exploration strategy simultaneously. In addition, \name\ only takes at most $75\%$ of Meta-Ban's running time for experiments, since Meta-Ban needs to train the framework individually for each arm before making predictions. We will discuss more about the running time in \textbf{Subsec.} \ref{subsec_running_time}.

\subsection{Effects of Propagation Hops $k$}
We also include the experiments on the MovieLens data set with 100 users to further investigate the effects of the propagation hyper-parameter $k$. Recall that given two input vectors $w, v$, we apply the RBF kernel as the mapping functions $\Psi^{(1)}(w, v) = \Psi^{(2)}(w, v) = \exp(-\gamma\cdot \|w - v\|^{2})$ where $\gamma$ is the kernel bandwidth.
The experiment results are shown in the \textbf{Table} \ref{table_different_hops_100_users} below, and the value in the brackets "[]" is the element standard deviation of the normalized adjacency matrix of user exploitation graphs.

\begin{table}[h]
    \centering
    \vspace{-0.3cm}
    \begin{tabular}{ |p{0.1cm}|p{1.8cm}p{1.8cm}p{1.8cm}p{1.8cm}|  }
     \hline
     &\multicolumn{4}{|c|}{ Bandwidth $\gamma$ } \\
     \hline
     \textbf{$k$} & 0.1& 1& 2& 5 \\
     \hline
     1    & 7276 [$1.6\times 10^{-4}$]   &  7073 [$1.4\times 10^{-3}$] &  7151 [$2.2\times 10^{-3}$] &  7490 [$3.9\times 10^{-3}$] \\
     \hline
     2  &   6968 [$1.0\times 10^{-4}$] & 6966 [$7.7\times 10^{-4}$] & 7074 [$1.3\times 10^{-3}$] &  7087 [$2.5\times 10^{-3}$] \\
     \hline
     3  &  7006 [$7.1\times 10^{-5}$] & 7018 [$7.0\times 10^{-4}$] &  6940 [$1.2\times 10^{-3}$] &  7167 [$1.9\times 10^{-3}$] \\
     \hline
    \end{tabular}
    % \vspace{+0.1in}
    \caption{ Cumulative regrets on MovieLens dataset with 100 users (different k / kernel bandwidth). The value in the brackets "[]" is the element standard deviation of the corresponding normalized adjacency matrix. }
    \vspace{-0.6cm}
    \label{table_different_hops_100_users}
\end{table}

Here, increasing the value of parameter $k$ will generally make the normalized adjacency matrix elements "smoother", as we can see from the decreasing standard deviation values. This matches the low-pass nature of graph multi-hop feature propagation \cite{SGC_wu2019simplifying}. 
With larger $k$ values, \name\ will be able to propagate the information for mores hops.
In contrast, with a smaller $k$ value, it is possible that the target user will be "heavily influenced" by only several specific users. 
% Therefore, the practitioner may need to choose $k$ value properly under different application scenarios.
% For a larger user graph, propagating multiple hops (larger $k$ values) seems to be more beneficial. One possible explanation can be that with a larger user graph, the user correlations will become more complex. 
% Thus, instead of only utilizing the information directly from the target user's neighbors, we need to involve the neighborhood information of the target user's neighbors for a global perspective over the users. In this case, properly increasing the $k$ value will indeed enable \name\ to obtain a more comprehensive neighborhood information around the target user, which leads to a better performance when working with the large user graph.
However, overly large $k$ values can also lead to the ``over-smoothing'' problem \citep{JK_Net_xu2018representation,linearized-GNN_xu2021optimization}, which can impair the model performance. Therefore, the practitioner may need to choose $k$ value properly under different application scenarios.

\subsection{Effects of the Approximated Neighborhood}   \label{subsec_exp_approx_neighborhood}

In this subsection, we conduct experiments to support our claim that applying approximated user neighborhoods is a feasible solution to reduce the computational cost, when facing the increasing number of users (\textbf{Remark} \ref{remark_numerous_users}).
We consider three scenarios where the number of users $n \in \{200, 300, 500\}$. Meanwhile, we let the size of the approximated user neighborhood $\Tilde{\mathcal{N}}^{(1)}(u_{t}), \Tilde{\mathcal{N}}^{(2)}(u_{t})$ fix to $\tilde{n} = \abs{\Tilde{\mathcal{N}}^{(1)}(u_{t})} = \abs{\Tilde{\mathcal{N}}^{(2)}(u_{t})} = 50$ for all these three experiment settings, and the neighborhood users are sampled from the user pool $\mathcal{U}$ in the experiments. 

\begin{figure}[!h]
  \centering
  \vspace{-0.3cm}
  \includegraphics[width=\linewidth]{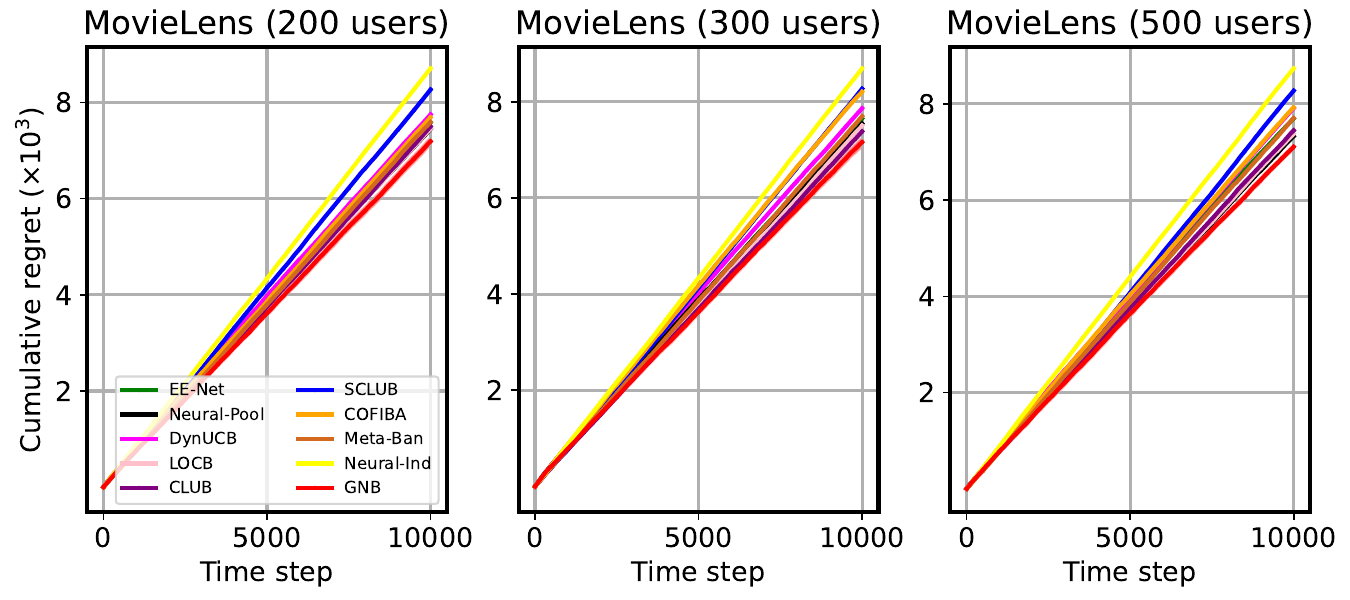}
  \vspace{-0.7cm}
  \caption{Cumulative regrets for different number of users with approximated user neighborhood (MovieLens data set).}
  \vspace{-0.6cm}
  \label{fig_MORE_USERS_MovieLens}
\end{figure}

\begin{table}[h]
    \centering
    \begin{tabular}{ |p{2cm}|p{0.8cm}p{0.8cm}p{0.8cm}p{0.8cm}p{0.8cm}|  }
     \hline
     &\multicolumn{5}{|c|}{ Avg. regret per round at different $t$ } \\
     
     \hline
     \textbf{Algorithm} & 2000  & 4000 & 6000 & 8000 & 10000 \\
     \hline
      CLUB   & 0.7691 & 0.7513 & 0.7464 & 0.7468 & 0.7496\\
     Neural-Ind  &   0.8901 &  0.8808 &  0.8790  &  0.8754 & 0.8741\\
     Neural-Pool   & 0.7681 & 0.7526 & 0.7405 & 0.7362 & 0.7334\\
     EE-Net  &   0.7886 & 0.7723 & 0.7642 & 0.7618 & 0.7582\\
     Meta-Ban  &  0.7811 & 0.7761 & 0.7754 & 0.7729 & 0.7708\\
     \hline
     \name\ ($\widetilde{n}=50$)  &  0.7760 & 0.7245 & 0.7190 & 0.7265 & 0.7140\\
     \name\ ($\widetilde{n}=100$)   & 0.7406 & 0.7178 & 0.7172 & 0.7110 & 0.7104\\
     \name\ ($\widetilde{n}=150$)  &  0.7291 & 0.7228 & 0.7129 & 0.7105 & 0.7085\\
     \hline
    \end{tabular}
    \caption{ Running for 10000 rounds and with the number of users $n=500$ for the MovieLens data set, the comparison between \name\ and baselines on average regret per round. }
    \vspace{-0.5cm}
    \label{table_different_size_appx_neighborhood}
\end{table}

Here, we see that the proposed \name\ still outperforms the baselines with increasing number of users. In particular, given a total of 500 users, the approximated neighborhood is only $10\%$ (50 users) of the overall user pool. These results can show that applying approximated user neighborhoods (Remark \ref{remark_numerous_users}) is a practical way to scale-up \name\ in real-world application scenarios.
%
% Meanwhile, setting $n=500$ and with different approximated neighborhood sizes $\widetilde{n}\in \{50, 100, 150\}$, the experiment results on the MovieLens data set are shown in \textbf{Table} \ref{table_different_size_appx_neighborhood}.
In addition, in \textbf{Table} \ref{table_different_size_appx_neighborhood}, we also include the average regret per round across different time steps.
With the number of users $n=500$ on the MovieLens data set, we include the experiments given different numbers of “representative users” $\widetilde{n}\in \{50, 100, 150\}$ to better show the model performance when applying the approximated neighborhood.
Here, increasing the number of “representative users” $\widetilde{n}$ can lead to better performances of \name, while it also shows that a small number of “representative users” will be enough for \name\ to achieve satisfactory performances.

\vspace{-0.15cm}
\subsection{Effects of the Adaptive Exploration}

To show the necessity of the adaptive exploration strategy, we consider an alternative arm selection mechanism (different from line 10, \textbf{Alg.} \ref{algo_main}) in round $t\in [T]$, as
% \begin{displaymath}
% \begin{split}
%     \vect{x}_{t} &= \arg\max_{\vect{x}_{i, t} \in \mathcal{X}_{t}}  
%     \bigg[ f_{gnn}^{(1)} \big( \vect{x}_{i, t},~\mathcal{G}_{i, t}^{(1)}; [\Theta_{gnn}^{(1)}]_{t-1} \big) \\
%     & 
%     + \alpha\cdot f_{gnn}^{(2)} \big( \nabla_{\Theta_{gnn}^{(1)}} f_{gnn}^{(1)}(\vect{x}_{i, t},~\mathcal{G}_{i, t}^{(1)}; [\Theta_{gnn}^{(1)}]_{t-1}),~\mathcal{G}_{i, t}^{(2)}; [\Theta_{gnn}^{(2)}]_{t-1} \big) \bigg]
% \end{split}
% \end{displaymath}
$
    \vect{x}_{t} = \arg\max_{\vect{x}_{i, t} \in \mathcal{X}_{t}} \big( \hat{r}_{i, t} + \alpha\cdot \hat{b}_{i, t} \big),
$
given the estimated reward and potential gain. Here, we introduce an additional parameter $\alpha\in [0, 1]$ as the exploration coefficient to control the exploration levels (i.e., larger $\alpha$ values will lead to higher levels of exploration). Here, we show the experiment results with $\alpha \in \{0, 0.1, 0.3, 0.7, 1.0\}$ on the ``MNIST'' and ``Yelp'' data sets.

% \begin{figure}[ht]
%   \centering
%   \includegraphics[width=\linewidth]{Figure/ablation_exp_coef_figure.pdf}
%   \caption{Results with different exploration coefficients $\alpha$.}
%   \label{fig_exp_coef_regret_results}
% \end{figure}

\begin{table}[h]
    \centering
    \vspace{-0.2cm}
    \begin{tabular}{ |p{1cm}|p{0.8cm}p{1.2cm}p{1.2cm}p{1.2cm}p{0.8cm}|  }
     \hline
     &\multicolumn{5}{|c|}{ Regret results with different $\alpha$ values } \\
     
     \hline
     \textbf{Dataset} & $\alpha=0$  & $\alpha=0.1$ & $\alpha=0.3$ & $\alpha=0.7$ & $\alpha=1$ \\
     \hline
      Yelp  & 7612 & 7444 & 7546 & 7509 & 7457\\
     MNIST  &   2323 &  2110 &  2170  & 2151 & 2141\\
     \hline
    \end{tabular}
    \caption{ Results with different exploration coefficients $\alpha$. }
    \vspace{-0.7cm}
    \label{table_exp_coef_regret_results}
\end{table}

In Table \ref{table_exp_coef_regret_results}, regarding the results of the ``Yelp'' data set, although the performances of \name\ do not differ significantly with different $\alpha$ values, our adaptive exploration strategy based on user exploration graphs is still helpful to improve \name's performances, which is validated by the fact that setting $\alpha\in (0, 1]$ will lead to better results compared with the situation where no exploration strategies are involved (setting $\alpha =0$).
On the other hand, as for the results of the ``MNIST'' data set, different $\alpha$ values will lead to relatively divergent results. One reason can be that with larger context dimension $d$ in the ``MNIST'' data set, the underlying reward mapping inclines to be more complicated compared with that of the ``Yelp'' data set. In this case, leveraging the exploration correlations will be more beneficial. 
Thus, the adaptive exploration strategy is necessary to improve the performance of \name\ by estimating the potential gains based on ``fine-grained'' user (class) correlations.

\vspace{-0.2cm}
\subsection{Running Time vs. Performance}      \label{subsec_running_time}

\begin{figure}[ht]
  \centering
  \vspace{-0.10cm}
  \includegraphics[width=0.8\linewidth]{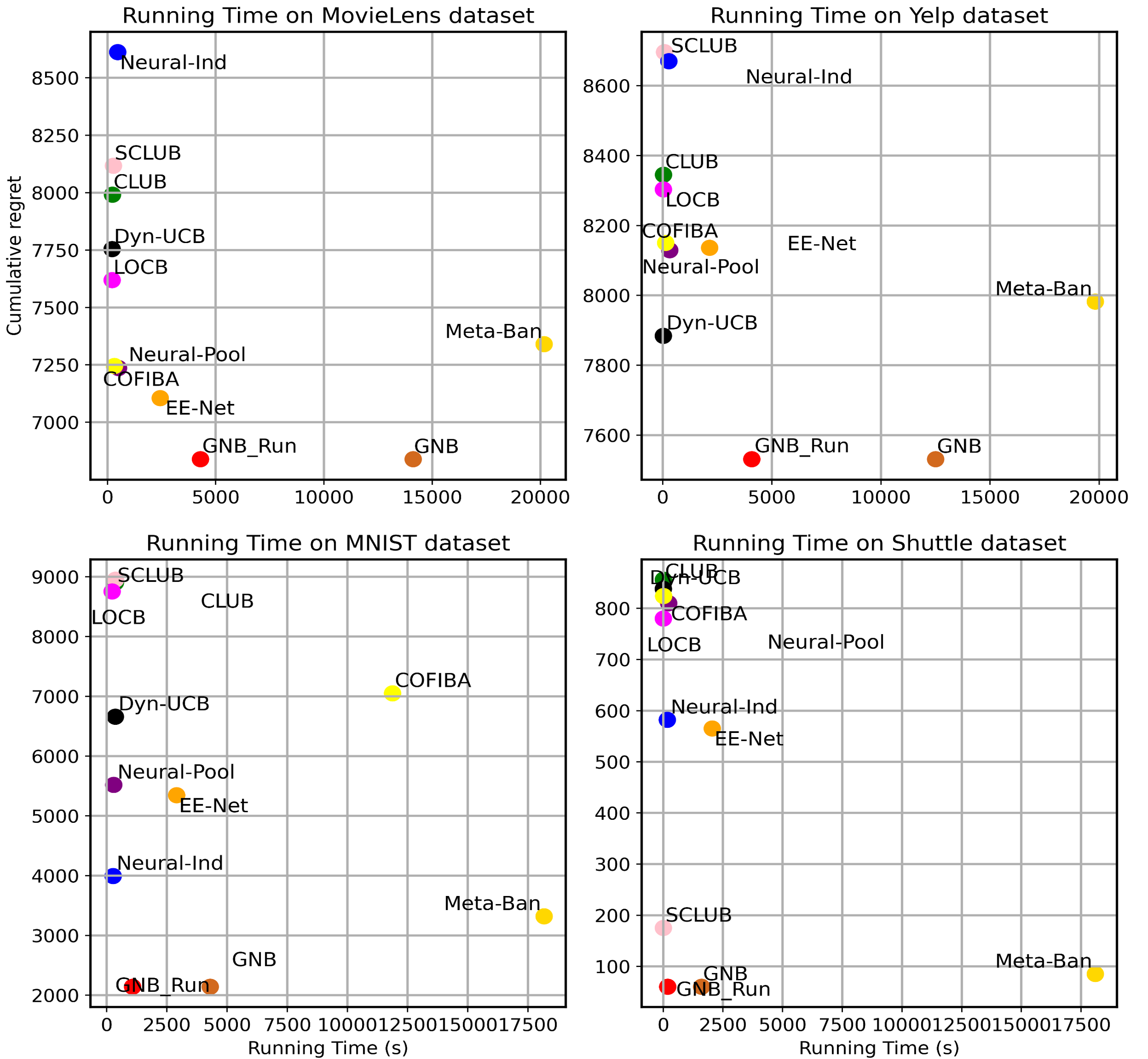}
  \vspace{-0.3cm}
  \caption{Running time vs. performance with baselines.}
  \label{fig_running_time_main}
  \vspace{-0.3cm}
\end{figure}

In Figure \ref{fig_running_time_main}, we show the results in terms of cumulative regret [y-axis, smaller $ = $ better] and running time [x-axis, smaller $ = $ better]. Additional results are in Appendix \textbf{Subsec.} \ref{subsec_appx_time_complex}. Each colored point here refers to one single method.
The point labeled as ``\name\_Run'' refers to the time consumption of \name\ on the arm recommendation process only, and the point ``\name'' denotes the overall running time of \name, including the recommendation and model training process.

Although the linear baselines tend to run faster compared with our proposed \name, their experiment performances (Subsec. \ref{subsec_exp_results_main}) are not comparable with \name, as their linear assumption can be too strong for many application scenarios. 
In particular, for the data set with high context dimension $d$, the mapping from the arm context to the reward will be much more complicated and more difficult to learn. For instance, as shown by the experiments on the MNIST data set ($d=784$), the neural algorithms manage to achieve a significant improvement over the linear algorithms (and the other baselines) while enjoying the reasonable running time.
Meanwhile, we also have the following remarks: (1) We see that for the two recommendation tasks, \name\ takes approximately $0.4$ second per round to make the arm recommendation with satisfactory performances for the received user;
(2) In all the experiments, we train the \name\ framework per 100 rounds after $t > 1000$ and still manage to achieve the good performance. In this case, the running time of \name\ in a long run can be further improved considerably by reducing the training frequency when we already have sufficient user interaction data and a well-trained framework;
(3) Moreover, since we are actually predicting the rewards and potential gains for all the nodes within the user graph (or the approximated user graph as in Remark \ref{remark_numerous_users}), \name\ is able to serve multiple users in each round simultaneously without running the recommendation procedure for multiple times, which is efficient in real-world cases.

\vspace{-0.2cm}
\subsection{Supplementary Experiments}  \label{subsec_supplement_exps}
% \vspace{-0.1cm}

Due to page limit, we present supplementary experiments in Appendix Section \ref{sec_appd_experiments}, including: (1) [\textbf{Subsec.} \ref{subsec_appx_underlying_groups}] experiments showing the potential impact on \name\ when there exist underlying user clusters; (2) [\textbf{Subsec.} \ref{subsec_appx_time_complex}] complementary contents for \textbf{Subsec.} \ref{subsec_running_time} regarding the ``Letter" and ``Pendigits'' data sets; (3) [\textbf{Subsec.} \ref{subsec_appx_conv_GNN}] the convergence results of \name\ on recommendation data sets.

% ====================================================
% \vspace{-0.35cm}
% \vspace{-0.1cm}
\vspace{-0.10cm}
\section{Conclusion}    \label{sec_conclusion}
% \vspace{-0.25cm}
In this paper, we propose a novel framework named \name~to model the fine-grained user collaborative effects. Instead of modeling user correlations through the estimation of rigid user groups, we estimate the user graphs to preserve the pair-wise user correlations for exploitation and exploration respectively, and utilize individual GNN-based models to achieve the adaptive exploration with respect to the arm selection. Under standard assumptions, we also demonstrate the improvement of the regret bound over existing methods from new perspectives of ``fine-grained'' user collaborative effects and GNNs.
Extensive experiments are conducted to show the effectiveness of our proposed framework against strong baselines.

\vspace{-0.10cm}
\begin{acks}
This work is supported by National Science Foundation under Award No. IIS-1947203, IIS-2117902, IIS-2137468, and Agriculture and Food Research Initiative (AFRI) grant no. 2020-67021-32799/project accession no.1024178 from the USDA National Institute of Food and Agriculture. The views and conclusions are those of the authors and should not be interpreted as representing the official policies of the funding agencies or the government.
\end{acks}

\clearpage

\bibliographystyle{ACM-Reference-Format}
\bibliography{User_GNN_Bandits}

\appendix

\onecolumn

\clearpage
\section{Experiments (Cont.)}    \label{sec_appd_experiments}

% -----------------------------------------------------
\subsection{Baselines and Experiment Settings}
The brief introduction for the nine baseline methods:
\begin{itemize}
    \item \textbf{CLUB} \citep{club_2014} regards connected components as user groups out of the estimated user graph, and adopts a UCB-type exploration strategy;
    \item \textbf{SCLUB} \citep{SCLUB_li2019improved} estimates dynamic user sets as user groups, and allows set operations for group updates;
    \item \textbf{LOCB} \citep{local_clustering-ban2021local} applies soft-clustering among users with random seeds and choose the best user group for reward and confidence bound estimations;
    \item \textbf{DynUCB} \citep{Dyn-UCB_nguyen2014dynamic} dynamically assigns users to its nearest estimated cluster.
    \item  \textbf{COFIBA} \citep{co_filter_bandits_2016} estimates user clustering and arm clustering simultaneously, and ensembles linear estimators for  reward and confidence bound estimations;
    \item  \textbf{Neural-Pool} adopts one single Neural-UCB \citep{Neural-UCB} model for all the users with UCB-type exploration strategy;
    \item  \textbf{Neural-Ind} assigns each user with their own separate Neural-UCB \citep{Neural-UCB} model;
    \item  \textbf{EE-Net} \citep{EE-Net_ban2021ee} achieves adaptive exploration by applying additional neural models for the exploration and decision making;
    \item  \textbf{Meta-Ban} \citep{Meta-Ban} utilizes individual neural models for each user's behavior, and applies a meta-model to adapt to estimated user groups.
\end{itemize}

% \subsection{Experiment Settings} \label{subsec_appx_baseline_settings}
% First, we add more descriptions about our baselines:
% \begin{itemize}[leftmargin=*]
%     \item \textbf{CLUB} \citep{club_2014} regards connected components as user groups out of the estimated user graph, and adopts a UCB-type exploration strategy;
%     \item \textbf{SCLUB} \citep{SCLUB_li2019improved} estimates dynamic user sets as user groups, and allows set operations for group updates;
%     \item \textbf{LOCB} \citep{local_clustering-ban2021local} applies soft-clustering among users with random seeds and choose the best user group for reward and confidence bound estimations;
%     \item \textbf{DynUCB} \citep{Dyn-UCB_nguyen2014dynamic} dynamically assigns users to its nearest estimated cluster.
%     \item  \textbf{COFIBA} \citep{co_filter_bandits_2016} estimates user clustering and arm clustering simultaneously, and ensembles linear estimators for  reward and confidence bound estimations;
%     \item  \textbf{Neural-Pool} adopts one single Neural-UCB \citep{Neural-UCB} model for all the users with UCB-type exploration strategy;
%     \item  \textbf{Neural-Ind} assigns each user with their own separate Neural-UCB \citep{Neural-UCB} model;
%     \item  \textbf{EE-Net} \citep{EE-Net_ban2021ee} achieves adaptive exploration by applying additional neural models for the exploration and decision making;
%     \item  \textbf{Meta-Ban} \citep{Meta-Ban} utilizes individual neural models for each user's behavior, and applies a meta-model to adapt to estimated user groups.
% \end{itemize}

Here, for all the UCB-based baselines, we choose their exploration parameter from the range $\{0.01, 0.1, 1\}$ with grid search. We set the $L=2$ for all the deep learning models including our proposed \name, and set the network width $m=100$. The learning rate of all neural algorithms are selected by grid search from the range $\{0.0001, 0.001, 0.01\}$. For EE-Net \cite{EE-Net_ban2021ee}, we follow the default settings in their paper by using a hybrid decision maker, where the estimation is $f_{1} + f_{2}$ for the first 500 time steps, and then we apply an additional neural network for decision making afterwards. For Meta-Ban, we follow the settings in their paper by tuning the clustering parameter $\gamma$ through the grid search on $\{0.1, 0.2, 0.3, 0.4\}$. For \name, we choose the $k$-hop user neighborhood $k\in \{1, 2, 3\}$ with grid search.
Reported results are the average of 3 runs.
The URLs for the data sets are:
% \begin{itemize}
MovieLens: \url{https://www.grouplens.org/datasets/movielens/20m/}.
Yelp: \url{https://www.yelp.com/dataset}.
MNIST/Shuttle/Letter/Pendigits: \url{https://archive.ics.uci.edu/ml/datasets}.

\vspace{-0.2cm}
\subsection{Experiments with Underlying User Groups}    \label{subsec_appx_underlying_groups}

\begin{figure}[th]
    \includegraphics[width=1\columnwidth]{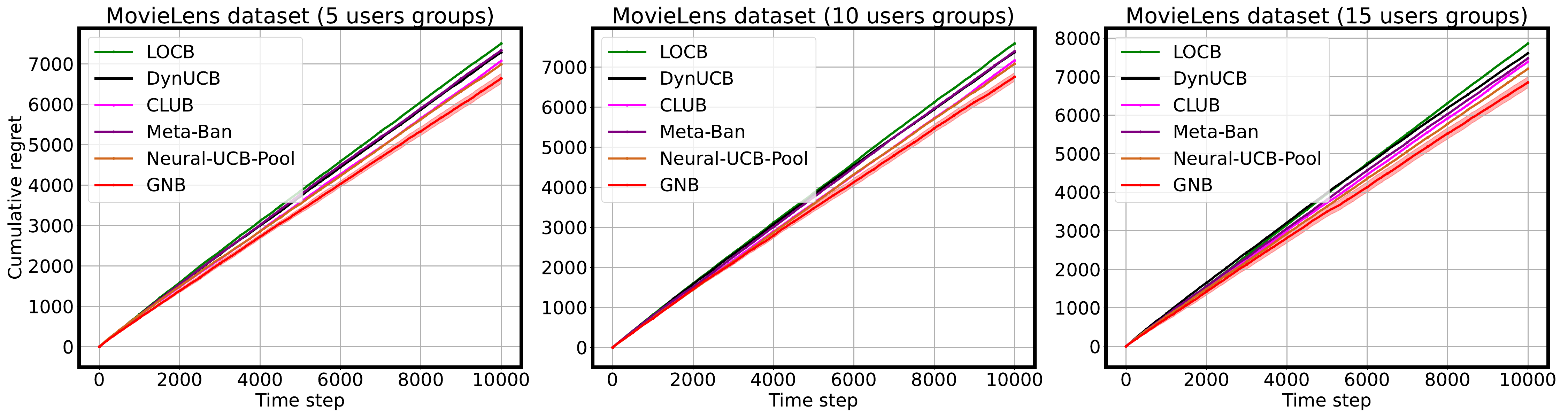}\hfill 
    \includegraphics[width=1\columnwidth]{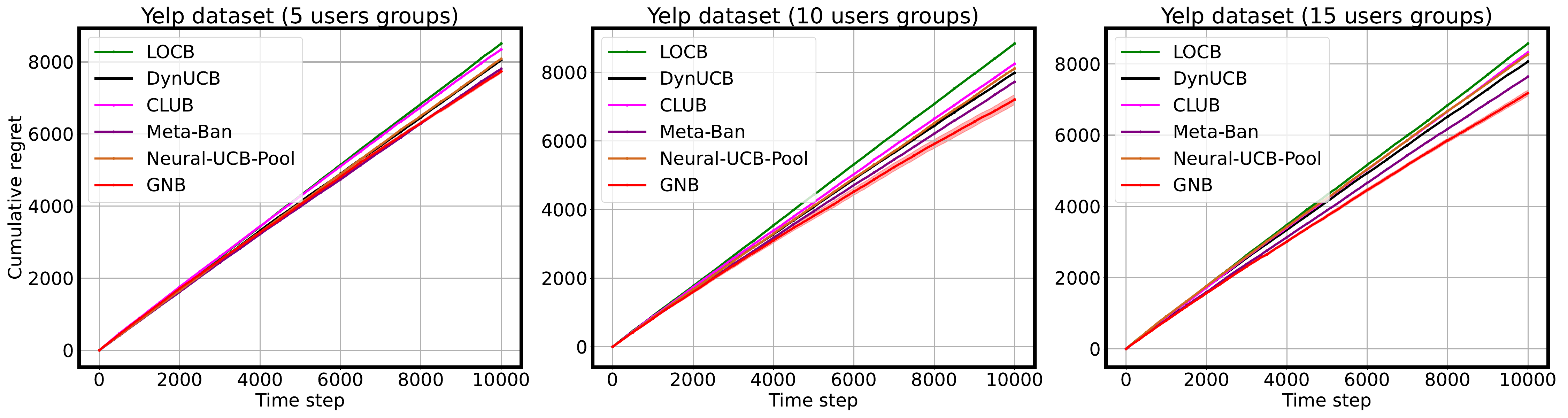}\hfill
    \caption{ Cumulative regrets for different number of underlying user groups (on MovieLens and Yelp data sets)}
       \label{fig_rebuttal_group_size}
\end{figure}

To understand the influence of potential underlying user clusters, we conduct the experiments on the MovieLens and the Yelp data sets, with controlled number of underlying user groups. The underlying user groups are derived by using hierarchical clustering on the user features, with a total of 50 users approximately. Here, we compare \name\ with four representative baselines with relatively good performances, including DynUCB \citep{Dyn-UCB_nguyen2014dynamic} [fixed number of user clusters], LOCB \citep{local_clustering-ban2021local} [fixed number of user clusters with local clustering], CLUB \citep{club_2014} [distance-based user clustering], Neural-UCB-Pool \citep{Neural-UCB} [neural single-bandit algorithm], and Meta-Ban \citep{Meta-Ban} [neural user clustering bandits]. DynUCB and LOCB are given the \textbf{true cluster number} as the prior knowledge to determine the quantity of user clusters or random seeds. Results are shown in Fig. \ref{fig_rebuttal_group_size}.

% \begin{figure}[ht]
%   \centering
%   \includegraphics[width=\linewidth]{Figure/cumu_regret_rebuttal_movielens_datasets_figure.pdf}
%   \caption{Cumulative regrets for different number of underlying user groups (MovieLens data set).}
%   \label{fig_rebuttal_group_size_MovieLens}
% \end{figure}

% \begin{figure}[ht]
%   \centering
%   \includegraphics[width=\linewidth]{Figure/cumu_regret_rebuttal_yelp_datasets_figure.pdf}
%   \caption{Cumulative regrets for different number of underlying user groups (Yelp data set).}
%   \label{fig_rebuttal_group_size_Yelp}
% \end{figure}

As we can see from the results, our proposed \name\ outperforms other baselines across different data sets and number of user groups. In particular, with more underlying user groups, the performance improvement of \name\ over the baselines will slightly increase, which can be the result of the increasingly complicated user correlations. 
The modeling of fine-grained user correlations, the adaptive exploration strategy with the user exploration graph, and the representation power of our GNN-based architecture can be the reasons for \name's good performances.

\subsection{Running Time vs. Performance (Cont.)}  
\label{subsec_appx_time_complex}

\begin{figure}[h]
  \centering
  \includegraphics[width=0.5\linewidth]{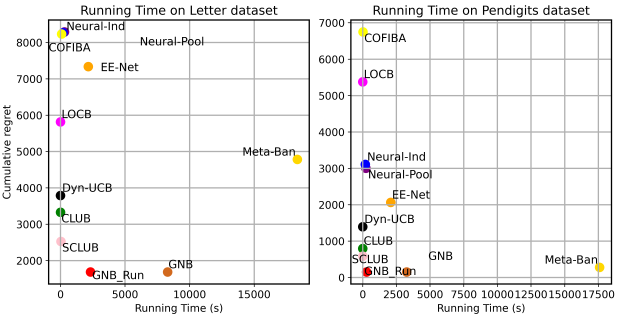}
  \caption{Running time \& performance comparison.}
  \label{fig_running_time_appx}
\end{figure}

In Figure \ref{fig_running_time_appx}, we include the comparison with baselines in terms of the running performance and running time on the last two classification data sets (``Letter'' and ``Pendigits'' data sets).
Analogously, the results match our conclusions that by applying the GNN models and the ``fine-grained" user correlations, \name\ can find a good balance between the computational cost and recommendation performance.

% From \textbf{Table} \ref{table_running_time}, we see that compared with the most closely related work, Meta-Ban, our proposed GNB is generally faster, since GNB does not required to re-train the model for each candidate arm. 
% \begin{table}[h]
%     \centering
%     \begin{tabular}{ |p{1.7cm}|p{1.5cm}p{1cm}p{1cm}p{1cm}|  }
%      \hline
%      &\multicolumn{4}{|c|}{ \textbf{Data sets} ($10^{4}$ rounds) } \\
     
%      \hline
%      \textbf{Methods} & MovieLens  & Yelp & MNIST & Shuttle \\
%      \hline
%      CLUB    & 230    &  36 &  359 & 4     \\
%      SCLUB    & 274    &  82 &  363 & 4     \\
%      LOCB    & 215    &  31 &  223 & 3    \\
%      DynUCB    & 214    &  29 &  357 & 3     \\
%      Neural-Pool    & 509    &  321 &  289 & 226     \\
%      Neural-Ind    & 472    &  281 &  265 & 169   \\
%      EE-Net    & 2435    &  2149 &  2903 & 2052    \\
%      COFIBA    & 321    &  135 &  11874 & 13     \\
%      Meta-Ban    & 20170    &  19825 &  18172 & 18101    \\
%      \hline
%      \textbf{GNB} (Ours)    &  14121 [4295]    &  12506 [4082] &  4299 [1072] &  1606 [185]    \\
%      \hline
%     \end{tabular}
%     \vspace{+0.1in}
%     \caption{Average running time results (seconds) on real data sets. The running time \textbf{in the brackets ``[]''} is the actual time consumption for recommendation w/o the time consumption for training.}
%     \label{table_running_time}
% \end{table}

\subsection{Convergence of GNN Models}  \label{subsec_appx_conv_GNN}

Here, for each separate time step interval of 2000 rounds, we show the average cumulative regrets results on MovieLens and Yelp data sets within this interval. We also include the baselines (EE-Net \cite{EE-Net_ban2021ee}, Neural-UCB-Pool \cite{Neural-UCB} and Meta-Ban \cite{Meta-Ban}) for comparison.

\begin{table}[h!]
    \centering
    \begin{tabular}{ |p{2.1cm}|p{1cm}p{1cm}p{1cm}p{1cm}p{1cm}|  }
     \hline
     &\multicolumn{5}{|c|}{ Time step intervals } \\
     
     \hline
     \textbf{Data (Algo.)} & 0-2k  & 2k-4k & 4k-6k & 6k-8k & 8k-10k  \\
     \hline
     \textbf{M (GNB)}   & 0.7050 &  0.6895 & 0.6795  &  0.6770 & 0.6685\\
     \textbf{Y  (GNB)}  &   0.7705 & 0.7685 & 0.7485 & 0.7450 & 0.7330\\
     \hline
     M (EE-Net)    & 0.7320 & 0.7120 & 0.7072 & 0.6945 & 0.7070 \\
     Y  (EE-Net)  &   0.8480 & 0.8420 & 0.7965 & 0.7950 & 0.7865\\
     M (Neural-UCB)    & 0.8090 & 0.7115 & 0.7165 & 0.6945 & 0.6865\\
      Y  (Neural-UCB)  &   0.8380 & 0.8115 & 0.8185 & 0.7940 & 0.8025\\
     M (Meta-Ban)    & 0.7760 & 0.7245 & 0.7190 & 0.7265 & 0.7140 \\
      Y  (Meta-Ban)  &   0.8525& 0.8035 & 0.7930 & 0.7600 & 0.7620\\
     \hline
    \end{tabular}
    % \vspace{+0.1in}
    \caption{ In different time step intervals, the average regrets per round on MovieLens (M) and Yelp (Y) data sets.}
    % \vspace{-0.7cm}
    \label{table_convergence_GNN}
\end{table}

As in Table \ref{table_convergence_GNN}, the average regret per round of GNB is decreasing, along with more time steps. Compared with baselines, GNB manages to achieve the best prediction accuracy across different time step intervals by modeling the fine-grained user correlations and applying the adaptive exploration strategy.
Here, one possible reason for the ``linear-like'' curves of the cumulative regrets is that these two recommendation data sets contain considerable inherent noise, which makes it hard for algorithms to learn the underlying reward mapping function. In this case, achieving experimental improvements on these two data sets is non-trivial. 
\clearpage
\section{Pseudo-code for Training the \name~Framework}   \label{sec_appd_pseudo_code}
In this section, we present the pseudo-code for training \name\ with GD with the following \textbf{Alg.} \ref{algo_training}.

\begin{algorithm}[H]
\caption{Model Training}
\label{algo_training}
\textbf{Input:} Initial parameter $\vect{\Theta_{0}}$, step size $\eta_{1}, \eta_{2}$, training steps $J_{1}, J_{2}$, network width $m$. Updated user graphs $\mathcal{G}_{t}^{(1)}$, $\mathcal{G}_{t}^{(2)}$. Served user $u_{t}$. \\
\textbf{Output:} Updated model parameters $[\vect{\Theta}_{u_{t}}^{(1)}]_{t}$, $[\vect{\Theta}_{u_{t}}^{(2)}]_{t}$, $[\vect{\Theta}_{gnn}^{(1)}]_{t}$ and $[\vect{\Theta}_{gnn}^{(2)}]_{t}$. \\

% ---------------------------------

\BlankLine

$[\matr{\Theta}_{u_{t}}^{(1)}]_{t}$, $[\matr{\Theta}_{u_{t}}^{(2)}]_{t}$ = User-Model-Training $\big( u_{t}, [\matr{\Theta}_{u_{t}}^{(1)}]_{0}, [\matr{\Theta}_{u_{t}}^{(2)}]_{0} \big)$. \\
\For{$\forall u'\in \mathcal{U}$, $u' \neq u_{t}$}{ 
    $[\matr{\Theta}_{u'}^{(1)}]_{t} \leftarrow [\matr{\Theta}_{u'}^{(1)}]_{t-1}$, \quad 
    $[\matr{\Theta}_{u'}^{(2)}]_{t} \leftarrow [\matr{\Theta}_{u'}^{(2)}]_{t-1}$\\
} 
$[\matr{\Theta}_{gnn}^{(1)}]_{t}$, $[\matr{\Theta}_{gnn}^{(2)}]_{t}$ = GNN-Model-Training $\big( [\matr{\Theta}_{gnn}^{(1)}]_{0}, [\matr{\Theta}_{gnn}^{(2)}]_{0} \big)$. \\
Return $[\matr{\Theta}_{u_{t}}^{(1)}]_{t}$, $[\matr{\Theta}_{u_{t}}^{(2)}]_{t}$, $[\matr{\Theta}_{gnn}^{(1)}]_{t}$, $[\matr{\Theta}_{gnn}^{(2)}]_{t}$.

\BlankLine
% ---------------------------------
\SetKwProg{myproc}{Procedure}{}{end procedure}
\myproc{User-Model-Training $\big( u_{t}, [\matr{\Theta}_{u_{t}}^{(1)}]_{0}, [\matr{\Theta}_{u_{t}}^{(2)}]_{0} \big)$}{
    $[\matr{\Theta}_{u_{t}}^{(1)}]^{0} \xleftarrow{} [\matr{\Theta}_{u_{t}}^{(1)}]_{0}$, \quad
    $[\matr{\Theta}_{u_{t}}^{(2)}]^{0} \xleftarrow{} [\matr{\Theta}_{u_{t}}^{(2)}]_{0}$. \\

    \# Training of $f_{u_{t}}^{(1)}(\cdot )$ \\
    Let $\mathcal{L}_{u}^{(1)}(\Theta_{u_{t}}^{(1)}) := \sum_{\tau\in \mathcal{T}_{u_{t}, t}} \abs{f_{u_{t}}^{(1)}(\vect{x}_{\tau}; 
    \matr{\Theta}_{u_{t}}^{(1)}) - r_{\tau}}^{2}$ \\
    \For{$j = 1, 2, \dots, J_{1}$}{ 
        $[\matr{\Theta}_{u_{t}}^{(1)}]^{j} = [\matr{\Theta}_{u_{t}}^{(1)}]^{j-1} - \eta_{1} \cdot\nabla_{\vect{\Theta}} \mathcal{L}_{u}^{(1)}([\matr{\Theta}_{u_{t}}^{(1)}]^{j-1})$ \\
    } 

    \# Training of $f_{u_{t}}^{(2)}(\cdot )$ \\
    Let $\mathcal{L}_{u}^{(2)}(\Theta_{u_{t}}^{(2)}) := \sum_{\tau\in \mathcal{T}_{u_{t}, t}} 
    \abs{f_{u_{t}}^{(2)}(\nabla_{\matr{\Theta}_{u_{t}}^{(1)}} f_{u_{t}}^{(1)}(\vect{x}_{\tau}; [\matr{\Theta}_{u_{t}}^{(1)}]_{\tau-1}); \matr{\Theta}_{u_{t}}^{(2)}) - 
    \big(r_{\tau} - f_{u_{t}}^{(1)}(\vect{x}_{\tau}; [\matr{\Theta}_{u_{t}}^{(1)}]_{\tau-1})\big)
    }^{2}$ \\
    \For{$j = 1, 2, \dots, J_{1}$}{ 
        $[\matr{\Theta}_{u_{t}}^{(2)}]^{j} = [\matr{\Theta}_{u_{t}}^{(2)}]^{j-1} - \eta_{1}\cdot\nabla_{\vect{\Theta}} \mathcal{L}_{u}^{(2)}([\matr{\Theta}_{u_{t}}^{(2)}]^{j-1})$ \\
    } 
    Let $[\widehat{\vect{\Theta}}_{u_{t}}^{(1)}]_{t} \leftarrow [\matr{\Theta}_{u_{t}}^{(1)}]^{J_{1}}$, 
    $[\widehat{\vect{\Theta}}_{u_{t}}^{(2)}]_{t} \leftarrow [\matr{\Theta}_{u_{t}}^{(2)}]^{J_{1}}$ \\
    Sample and return new parameters $([\vect{\Theta}_{u_{t}}^{(1)}]_{t}, [\vect{\Theta}_{u_{t}}^{(2)}]_{t}) \sim \{ ([\widehat{\vect{\Theta}}_{u_{t}}^{(1)}]_{\tau}, [\widehat{\vect{\Theta}}_{u_{t}}^{(2)}]_{\tau}) \}_{\tau\in [t]}$.\\
    %
    % Sample and Return new parameters $[\vect{\Theta}_{u_{t}}^{(1)}]_{t}$ and $[\vect{\Theta}_{u_{t}}^{(2)}]_{t}$.
    % Return new parameters $[\matr{\Theta}_{u_{t}}^{(1)}]^{J}$, $[\matr{\Theta}_{u_{t}}^{(2)}]^{J}$. \\
}
\BlankLine
% ---------------------------------
\myproc{GNN-Model-Training $\big( [\vect{\Theta}_{gnn}^{(1)}]_{0}$, $[\vect{\Theta}_{gnn}^{(2)}]_{0} \big)$}{
    $[\matr{\Theta}_{gnn}^{(1)}]^{0} \xleftarrow{} [\matr{\Theta}_{gnn}^{(1)}]_{0}$, \quad
    $[\matr{\Theta}_{gnn}^{(2)}]^{0} \xleftarrow{} [\matr{\Theta}_{gnn}^{(2)}]_{0}$. \\

    \# Training of $f_{gnn}^{(1)}(\cdot )$ \\
    Let $\mathcal{L}_{gnn}^{(1)}(\Theta_{gnn}^{(1)}) := \sum_{\tau\in [t]} \abs{f_{gnn}^{(1)}(\vect{x}_{\tau}, \mathcal{G}_{\tau}^{(1)}; 
    \matr{\Theta}_{gnn}^{(1)}) - r_{\tau}}^{2}$ \\
    \For{$j = 1, 2, \dots, J_{2}$}{ 
        $[\matr{\Theta}_{gnn}^{(1)}]^{j} = [\matr{\Theta}_{gnn}^{(1)}]^{j-1} - \eta_{2} \cdot\nabla_{\vect{\Theta}} \mathcal{L}_{gnn}^{(1)}([\matr{\Theta}_{gnn}^{(1)}]^{j-1})$ \\
    } 

    \# Training of $f_{gnn}^{(2)}(\cdot )$ \\
    Apply $f_{gnn}^{(1)}(\vect{x}_{\tau})$ to denote $f_{gnn}^{(1)}(\vect{x}_{\tau}, \mathcal{G}_{\tau}^{(1)}; [\matr{\Theta}_{gnn}^{(1)}]_{\tau-1})$. \\
    Let $\mathcal{L}_{gnn}^{(2)}(\Theta_{gnn}^{(2)}) := \sum_{\tau\in [t]} 
    \abs{f_{gnn}^{(2)}(\nabla_{\matr{\Theta}_{gnn}^{(1)}} f_{gnn}^{(1)}(\vect{x}_{\tau}), \mathcal{G}_{\tau}^{(2)}; \matr{\Theta}_{gnn}^{(2)}) - 
    \big(r_{\tau} - f_{gnn}^{(1)}(\vect{x}_{\tau}, \mathcal{G}_{\tau}^{(1)}; [\matr{\Theta}_{gnn}^{(1)}]_{\tau-1})\big)
    }^{2}$ \\
    \For{$j = 1, 2, \dots, J_{2}$}{ 
        $[\matr{\Theta}_{gnn}^{(2)}]^{j} = [\matr{\Theta}_{gnn}^{(2)}]^{j-1} - \eta_{2} \cdot\nabla_{\vect{\Theta}} \mathcal{L}_{gnn}^{(2)}([\matr{\Theta}_{gnn}^{(2)}]^{j-1})$ \\
    }
    Let $[\widehat{\vect{\Theta}}_{gnn}^{(1)}]_{t} \leftarrow [\matr{\Theta}_{gnn}^{(1)}]^{J_{2}}$, 
    $[\widehat{\vect{\Theta}}_{gnn}^{(2)}]_{t} \leftarrow [\matr{\Theta}_{gnn}^{(2)}]^{J_{2}}$ \\
    Sample and return new parameters $([\vect{\Theta}_{gnn}^{(1)}]_{t}, [\vect{\Theta}_{gnn}^{(2)}]_{t}) \sim \{ ([\widehat{\vect{\Theta}}_{gnn}^{(1)}]_{\tau}, [\widehat{\vect{\Theta}}_{gnn}^{(2)}]_{\tau}) \}_{\tau\in [t]}$. \\
    %
    % Return new parameters $[\vect{\Theta}_{gnn}^{(1)}]_{t}$ and $[\vect{\Theta}_{gnn}^{(2)}]_{t}$.
}
\end{algorithm}

\vspace{-0.15cm}
\section{Intuition of Adaptive Exploration}
\label{sec_appd_adaptive_exp}

Recall that GNB adopts a second GNN model ($f_{gnn}^{(2)}$) to adaptively learn the potential gain, which can be either positive or negative. 
The intuition is that the exploitation model (i.e., $f_{gnn}^{(1)}(\cdot)$) can "provide the excessively high estimation of the reward", and applying the UCB based exploration (e.g., \cite{Neural-UCB}) can amplify the mistake as the UCB is non-negative. 
For the simplicity of notation, let us denote the expected reward of an arm $x$ as $\mathbb{E}[r] = h(x)$, where $h$ is the unknown reward mapping function. The reward estimation is denoted as $\hat{r} = f_{1}(x)$ where $f_{1}(\cdot)$ is the exploitation model. 
For \name,
when the estimated reward is lower than the expected reward ($f_{1}(x) < h(x)$), we will apply the "upward" exploration (i.e., positive exploration score) to increase the chance of arm $x$ being explored. Otherwise, if the estimated reward is higher than the expected reward ($f_{1}(x) > h(x)$), we will apply the "downward" exploration  (i.e., negative exploration score)  instead to tackle the excessively high reward estimation. 
Here, we apply the exploration GNN $f_{gnn}^{(2)}$ to adaptively learn the relationship between the gradients of $f_{1}$ and the reward estimation residual $h(x) - f_{1}(x)$, based on the user exploration graph for a refined exploration strategy.
Readers can also refer to \cite{EE-Net_ban2021ee} for additional insights of neural adaptive exploration.

% ------------------------------------------------------------------------------
% ------------------------------------------------------------------------------
% \clearpage
\vspace{-0.2cm}
\section{ Proof Sketch of \textbf{Theorem} \ref{theorem_regret_bound} }    \label{sec_appd_regre_bound_proof}

% Due to the page limit, we include the proof sketch for \textbf{Theorem} \ref{theorem_regret_bound} in this section.
% For the full proof of the regret bound, please refer to our arXiv version of the paper.
%
Apart from two kinds of estimated user graphs $\{\mathcal{G}_{i, t}^{(1)}\}_{i\in [a]}$, $\{\mathcal{G}_{i, t}^{(2)}\}_{i\in [a]}$ at each time step $t$, we can also define true user exploitation graph $\{\mathcal{G}_{i, t}^{(1), *}\}_{i\in [a]}$ and true user exploration graph $\{\mathcal{G}_{i, t}^{(2), *}\}_{i\in [a]}$ based on \textbf{Def.} \ref{def_exploitation_simi} and \textbf{Def.} \ref{def_exploration_simi} respectively. Comparably, the true normalized adjacency matrices of $\mathcal{G}_{i, t}^{(1), *}, i\in [a]$ are represented as $\vect{S}_{i, t}^{(1), *}$.
With $r_{t}, r_{t}^{*}$ separately being the rewards for the selected arm $\vect{x}_{t}\in \mathcal{X}_{t}$ and the optimal arm $\vect{x}_{t}^{*}\in \mathcal{X}_{t}$, we formulate the pseudo-regret for a single round $t$, as
$
    R_{t} = \mathbb{E} [r_{t}^{*} |  u_{t}, \mathcal{X}_{t}] - \mathbb{E} [r_{t} |  u_{t}, \mathcal{X}_{t}]
$
w.r.t. the candidate arms $\mathcal{X}_{t}$ and served user $u_{t}$. Here, with \textbf{Algorithm} \ref{algo_main}, we denote 
$
    f_{gnn}(\vect{x}_{t}) = \ f_{gnn}^{(1)}(\vect{x}_{t},~\mathcal{G}_{t}^{(1)}; [\matr{\Theta}_{gnn}^{(1)}]_{t-1}) 
    + f_{gnn}^{(2)}(\nabla f_{t}^{(1)} (\vect{x}_{t}) ,~\mathcal{G}_{t}^{(2)}; [\matr{\Theta}_{gnn}^{(2)}]_{t-1}),
$
with the arm $\vect{x}_{t}$, gradients $\nabla f_{t}^{(1)} (\vect{x}_{t}) = \frac{\nabla_{\matr{\Theta}_{gnn}^{(1)}} f_{gnn}^{(1)}(\vect{x}_{t},~\mathcal{G}_{t}^{(1)}; [\matr{\Theta}_{gnn}^{(1)}]_{t-1})}{c_{g}L}$ ($c_{g} > 0$ is the normalization factor, such that $\norm{\nabla f_{t}^{(1)} (\vect{x}_{t})}_{2} \leq 1$), as well as the estimated user graphs $\mathcal{G}_{t}^{(1)}$, $\mathcal{G}_{t}^{(2)}$ related to chosen arm $\vect{x}_{t}$. 
On the other hand, with the true graph $\mathcal{G}_{t}^{(1), *}$ of arm $\vect{x}_{t}$,
the corresponding gradients will be $\nabla f_{t}^{(1), *} (\vect{x})$.
Analogously, we also have the estimated user graphs $\mathcal{G}_{t, *}^{(1)}, \mathcal{G}_{t, *}^{(2)}$ for the optimal arm $\vect{x}_{t}^{*}$.
Afterwards, in round $t \in [T]$, the single-round regret $R_{t}$ will be
\begin{displaymath}
\begin{split}
    R_{t} & = \mathbb{E} [r_{t}^{*} |  u_{t}, \mathcal{X}_{t}] - \mathbb{E} [r_{t} |  u_{t}, \mathcal{X}_{t}] \\
    & = \mathbb{E} [r_{t}^{*} |  u_{t}, \mathcal{X}_{t}]  - f_{gnn}(\vect{x}_{t}) + f_{gnn}(\vect{x}_{t}) - \mathbb{E} [r_{t} |  u_{t}, \mathcal{X}_{t}] \\
    & \underset{(\text{i})}{\leq} \mathbb{E} [r_{t}^{*} |  u_{t}, \mathcal{X}_{t}]  - f_{gnn}(\vect{x}_{t}^{*}) + f_{gnn}(\vect{x}_{t}) - \mathbb{E} [r_{t} |  u_{t}, \mathcal{X}_{t}] \\
    & \leq  \mathbb{E} \big[ \abs{r_{t}^{*} - f_{gnn}(\vect{x}_{t}^{*})} \big|  u_{t}, \mathcal{X}_{t}  \big] + 
    \mathbb{E} \big[ \abs{r_{t} - f_{gnn}(\vect{x}_{t})} \big|  u_{t}, \mathcal{X}_{t}  \big] \\
    & = \mathbb{E} \bigg[ \abs{f_{gnn}^{(2)}(\nabla f_{t}^{(1)} (\vect{x}_{t}^{*}),~\mathcal{G}_{t, *}^{(2)}; [\matr{\Theta}_{gnn}^{(2)}]_{t-1}) - (r_{t}^{*} -  f_{gnn}^{(1)}(\vect{x}_{t}^{*},~\mathcal{G}_{t, *}^{(1)}; [\matr{\Theta}_{gnn}^{(1)}]_{t-1}))}  \bigg| u_{t}, \mathcal{X}_{t} \bigg] + \\
    & \quad \mathbb{E} \bigg[ \abs{f_{gnn}^{(2)}(\nabla f_{t}^{(1)} (\vect{x}_{t}),~\mathcal{G}_{t}^{(2)}; [\matr{\Theta}_{gnn}^{(2)}]_{t-1})  - (r_{t} -  f_{gnn}^{(1)}(\vect{x}_{t},~\mathcal{G}_{t}^{(1)}; [\matr{\Theta}_{gnn}^{(1)}]_{t-1}))}  \bigg| u_{t}, \mathcal{X}_{t} \bigg] \\
    & = \mathsf{CB}_{t}(\vect{x}_{t}) + \mathsf{CB}_{t}(\vect{x}_{t}^{*})
\end{split}
\end{displaymath}
where inequality $(\text{i})$ is due to the arm pulling mechanism, i.e., $f_{gnn}(\vect{x}_{t}) \geq f_{gnn}(\vect{x}_{t}^{*})$, and $\mathsf{CB}_{t}(\cdot)$ is the regret bound function in round $t$, formulated by the last equation. Then, given arm $\vect{x} \in \mathcal{X}_{t}$ and its reward $r$, with the aforementioned notation, we have 
% \begin{equation}
\vspace{-0.2cm}
\begin{equation}
\begin{split}
     \mathsf{CB}_{t}(\vect{x}) & = \mathbb{E} \bigg[ \abs{f_{gnn}^{(2)}(\nabla f_{t}^{(1)} (\vect{x}),~\mathcal{G}^{(2)}; [\matr{\Theta}_{gnn}^{(2)}]_{t-1}) \
    - (r -  f_{gnn}^{(1)}(\vect{x},~\mathcal{G}^{(1)}; [\matr{\Theta}_{gnn}^{(1)}]_{t-1}))}  \bigg| u_{t}, \mathcal{X}_{t} \bigg] \\
    & \leq \underbrace{\mathbb{E} \bigg[ \abs{f_{gnn}^{(2)}(\nabla f_{t}^{(1), *} (\vect{x}),~\mathcal{G}^{(2), *}; [\matr{\Theta}_{gnn}^{(2)}]_{t-1})  - (r -  f_{gnn}^{(1)}(\vect{x},~\mathcal{G}^{(1), *}; [\matr{\Theta}_{gnn}^{(1)}]_{t-1}))} \ \bigg| u_{t}, \mathcal{X}_{t} \bigg] }_{I_{1}}  \\
    & + \underbrace{\mathbb{E} \bigg[ \abs{f_{gnn}^{(1)}(\vect{x},~\mathcal{G}^{(1), *}; [\matr{\Theta}_{gnn}^{(1)}]_{t-1}) - f_{gnn}^{(1)}(\vect{x},~\mathcal{G}^{(1)}; [\matr{\Theta}_{gnn}^{(1)}]_{t-1})}  \bigg| u_{t}, \mathcal{X}_{t}\bigg]}_{I_{2}} \\
    & + \underbrace{\mathbb{E} \bigg[ \abs{f_{gnn}^{(2)}(\nabla f_{t}^{(1), *} (\vect{x}), ~\mathcal{G}^{(2), *}; [\matr{\Theta}_{gnn}^{(2)}]_{t-1}) -   f_{gnn}^{(2)}(\nabla f_{t}^{(1), *} (\vect{x}),~\mathcal{G}^{(2)}; [\matr{\Theta}_{gnn}^{(2)}]_{t-1})} \ \bigg| u_{t}, \mathcal{X}_{t} \bigg] }_{I_{3}} \\
    & + \underbrace{ \mathbb{E} \bigg[ \abs{f_{gnn}^{(2)}(\nabla f_{t}^{(1), *} (\vect{x}),~\mathcal{G}^{(2)}; [\matr{\Theta}_{gnn}^{(2)}]_{t-1})    - f_{gnn}^{(2)}(\nabla f_{t}^{(1)} (\vect{x}),~\mathcal{G}^{(2)}; [\matr{\Theta}_{gnn}^{(2)}]_{t-1})} \ \bigg| u_{t}, \mathcal{X}_{t}  \bigg] }_{I_{4}}.
\end{split}
\label{eq_single_CB}
\end{equation}
\vspace{-0.2cm}
Here, we have the term $I_{1}$ representing the estimation error induced by the GNN model parameters $\{[\matr{\Theta}_{gnn}^{(1)}]_{t-1}, [\matr{\Theta}_{gnn}^{(2)}]_{t-1}\}$, the term $I_{2}$ denoting the error caused by the estimation of user exploitation graph. Then, error term $I_{3}$ is caused by the estimation of user exploration graph, and term $I_{4}$ is the output difference given input gradients $\nabla f_{t}^{(1), *} (\vect{x})$ and $ \nabla f_{t}^{(1)} (\vect{x})$, which are individually associated with the true user exploitation graph $\mathcal{G}^{(1), *}$ and the estimation $\mathcal{G}^{(1)}$. 
These four terms $I_{1}, I_{2}, I_{3}, I_{4}$ can be respectively bounded by \blemma \ref{lemma_term_I_1_selected_arm} (\textbf{Corollary} \ref{corollary_term_I_1_optimal_arm} and the bounds in Subsection \ref{subsec_bounding_I_1}), \blemma \ref{lemma_I_2_bound}, \blemma \ref{lemma_I_3_bound}, and \blemma \ref{lemma_I_4_overall_bound} in the appendix. 
Afterwards, with the notation from \btheorem \ref{theorem_regret_bound}, we have the pseudo regret after $T$ rounds, i.e., $R(T)$, as

\begin{displaymath}
\begin{split}
    %  R(t) & = \sum_{t\in [T]} R_{t} \\
    % & \leq  2\cdot \bigg( 
    % %
    % \sqrt{t} \cdot \big( \sqrt{2\xi_{2}} + \frac{3L}{\sqrt{2}} + (1 + \gamma_{2}) \sqrt{2\log(\frac{Tn\cdot a}{\delta})}\big) \\
    % %
    % & \quad + \big(1 + \mathcal{O}(\frac{tL^{3} \log^{5/6} (m)}{\rho^{1/3}m^{1/6}})\big)\cdot \mathcal{O}(\frac{t^{3}L}{\rho\sqrt{m}}\log(m)) +
    % \mathcal{O}\big( \frac{t^{4}L^{2} \log^{11/6} (m)}{\rho^{4/3}m^{1/6}} \big) \\
    % %
    % &\quad + \mathcal{O}(\xi_{L}) \cdot \sqrt{8t} \cdot \big( \sqrt{2\xi_{1}} + \frac{3L}{\sqrt{2}} + (1 + \gamma_{1}) \sqrt{2\log(\frac{Tn\cdot a}{\delta})}\big) 
    % %
    % +  \mathcal{O} ( \frac{\xi_{L}tL^{7/2}  \sqrt{\log m}}{\sqrt{mn}}) + 4\Gamma_{t}
    % \bigg) \implies  \\
    %
    R(T) = & \sum_{t\in [T]} R_{t} \\
    & \leq 2 \cdot \sqrt{T} \big( \sqrt{2\xi_{2}} + \frac{3L}{\sqrt{2}} + (1 + \gamma_{2}) \sqrt{2\log(\frac{Tn\cdot a}{\delta})}\big)  +
    \sqrt{T}\cdot \mathcal{O}(\xi_{L}) \cdot \big( \sqrt{2\xi_{1}} + \frac{3L}{\sqrt{2}} + (1 + \gamma_{1}) \sqrt{2\log(\frac{Tn\cdot a}{\delta})}\big) + \mathcal{O}(1) 
    %
    % & \leq \sqrt{T} \cdot  \bigg( \mathcal{O}(\sqrt{\xi_{2}} + \xi_{L}\sqrt{\xi_{1}} ) 
    %     + \mathcal{O}(L\xi_{L}) + \mathcal{O}(\xi_{L}) \cdot \sqrt{2\log(\frac{Tn\cdot a}{\delta})}
    %      \bigg) + \\
    % & \qquad  \sqrt{T}\cdot \mathcal{O}(L) + \mathcal{O}(\sqrt{T}) + \mathcal{O}(1)
\end{split}
\end{displaymath}
where the inequality is because we have
sufficient large network width $m \geq \Omega ( \text{Poly}(T, L, a, 1 / \rho) \cdot \log(1 / \delta) ) $ as indicated in \textbf{Theorem} \ref{theorem_regret_bound}. Meanwhile, with sufficient $m \geq \Omega(\text{Poly}(T, \rho^{-1}))$, the terms $\gamma_{1}, \gamma_{2}$ can also be upper bounded by $\mathcal{O}(1)$, which leads to
\begin{displaymath}
\begin{split}
   & R(T)  \leq \sqrt{T} \cdot  \big( \mathcal{O}(\sqrt{\xi_{2}} + \xi_{L}\sqrt{\xi_{1}} ) 
        + \mathcal{O}(L\xi_{L}) + \mathcal{O}(\xi_{L}) \cdot \sqrt{2\log(\frac{Tn\cdot a}{\delta})}
         \big) + \\
     & \qquad\qquad \sqrt{T}\cdot \mathcal{O}(L) + \mathcal{O}(\xi_{L}) + \mathcal{O}(1) \\
    & \leq  \sqrt{T} \cdot  \big( 
        \mathcal{O}(L\xi_{L}) + \mathcal{O}(\xi_{L}) \cdot \sqrt{2\log(\frac{Tn\cdot a}{\delta})}
         \big) +  \sqrt{T} \mathcal{O}(L) + \mathcal{O}(\xi_{L}) + \mathcal{O}(1)  \\
\end{split}
\end{displaymath}
since we have $\xi_{1}, \xi_{2} \leq \mathcal{O}(\frac{1}{T})$.
This will complete the proof.

\section{Generalization of User Networks after GD}     \label{sec_appd_analysis_user}

In this section, we present the generalization results of user networks $f_{u}^{(1)}(\cdot; \matr{\Theta}_{u}^{(1)}), f_{u}^{(2)}(\cdot; \matr{\Theta}_{u}^{(2)}), u\in \mathcal{U}$. 
Up to a certain time step $t$ and for a given user $u\in \mathcal{U}$, we have all its past arm-reward pairs $\mathcal{P}_{u, t-1} = \{(\vect{x}_{\tau}, r_{\tau})\}_{\tau\in \mathcal{T}_{u, t}}$. 
Without loss of generality, following an analogous approach as in \citep{conv_theory-allen2019convergence}, we let the last coordinate of each arm context $\vect{x}$ to be a fixed small constant $c_{x} > 0$. 
This can be achieved by simplify normalizing the input $\vect{x}$ as $\vect{x} \leftarrow \big( \vect{x}\cdot \sqrt{1-c_{x}^{2}}, c_{x} \big)$.
Meanwhile, inspired by \citep{conv_theory-allen2019convergence}, with two vectors $\Tilde{\vect{x}}, \vect{x}$ such that $\norm{\Tilde{\vect{x}}}_{2} \leq 1, \norm{\vect{x}}_{2} = 1$, we first define the the following operator
\begin{equation}
\begin{split}
    \phi(\Tilde{\vect{x}}, \vect{x}) = 
    (\frac{\Tilde{\vect{x}}}{\sqrt{2}}, \frac{\vect{x}}{2}, c)
\end{split}
\label{eq_concat_transformation}
\end{equation}
as the concatenation of the two vectors $\frac{\Tilde{\vect{x}}}{\sqrt{2}}, \frac{\vect{x}}{2}$ and one constant $c$, where $c = \sqrt{\frac{3}{4} - (\frac{\norm{\Tilde{\vect{x}}}_{2}}{\sqrt{2}})^{2}} \geq \frac{1}{2}$. And this operator makes the transformed vector $\norm{\phi(\Tilde{\vect{x}}, \vect{x})}_{2} = 1$.
The idea of this operator is to make the gradients $\nabla_{\matr{\Theta}_{u}^{(1)}} f_{u}^{(1)}(\cdot; \matr{\Theta}_{u}^{(1)})$ of the user exploitation model, which is the input of the user exploration model $f_{u}^{(2)}(\cdot)$, comply with the normalization requirement and the separateness assumption (\textbf{Assumption} \ref{assumption_separateness}). For the sake of analysis, we will adopt this operation in the following proof. Note that this operator is just one possible solution, and our results could be easily generalized to other forms of input gradients under the unit-length and separateness assumption. Similar ideas are also applied in previous works \citep{EE-Net_ban2021ee}.

% ==================================================================
\subsection{User Exploitation Model}

% -------------------------------------
With the convergence result presented in \blemma \ref{lemma_convergence_user_f_1}, we could bound the output of the user exploitation model $f_{u}^{(1)}(\cdot)$ after GD with the following lemma.

% -------------------------------------
\begin{lemma} \label{lemma_user_f_1_output_bound}
For the constants $\rho \in (0, \mathcal{O}(\frac{1}{L}))$ and $\xi_{1} \in (0, 1)$, given user $u\in\mathcal{U}$ and its past records $\mathcal{P}_{u, t-1}$ up to time step $t$, we suppose $m, \eta_1, J_1$ satisfy the conditions in \btheorem \ref{theorem_regret_bound}.
Then, with probability at least $1 - \delta$, given an arm-reward pair $(\vect{x}, r)$, we have
\begin{displaymath}
    \abs{f_{u}^{(1)} (\vect{x}; [\widehat{\matr{\Theta}}_{u}^{(1)}]_{t})} \leq \gamma_{1}
\end{displaymath}
where 
\begin{displaymath}
    \gamma_{1} = 
    2 + \mathcal{O} \left( \frac{ t^3 L}{ n^{3} \rho\sqrt{m}} \log m \right)
    + \mathcal{O} \left(  \frac{ L^{2}t^4 }{ n^{4} \rho^{4 / 3} m^{1 / 6}} \log^{11 / 6} (m) \right).
\end{displaymath}
\end{lemma}

\textbf{Proof.}
For brevity, we use $\widehat{\matr{\Theta}}_{u}^{(1)}$ to denote $[\widehat{\matr{\Theta}}_{u}^{(1)}]_{t}$.
The LHS of the inequality could be written as
\begin{displaymath}
\begin{split}
    \abs{f_{u}^{(1)} (\vect{x}; \widehat{\matr{\Theta}}_{u}^{(1)})} \leq & 
    \abs{f_{u}^{(1)} (\vect{x}; \widehat{\matr{\Theta}}_{u}^{(1)}) -  f_{u}^{(1)} (\vect{x}; [\widehat{\matr{\Theta}}_{u}^{(1)}]_{0}) - \inp{\nabla_{[\widehat{\matr{\Theta}}_{u}^{(1)}]_{0}} f_{u}^{(1)} (\vect{x}; [\widehat{\matr{\Theta}}_{u}^{(1)}]_{0})}{\widehat{\matr{\Theta}}_{u}^{(1)} - [\widehat{\matr{\Theta}}_{u}^{(1)}]_{0}}} \\
    & + \abs{f_{u}^{(1)} (\vect{x}; [\widehat{\matr{\Theta}}_{u}^{(1)}]_{0}) + \inp{\nabla_{[\widehat{\matr{\Theta}}_{u}^{(1)}]_{0}} f_{u}^{(1)} (\vect{x}; [\widehat{\matr{\Theta}}_{u}^{(1)}]_{0})}{\widehat{\matr{\Theta}}_{u}^{(1)} - [\widehat{\matr{\Theta}}_{u}^{(1)}]_{0}}}.
\end{split}
\end{displaymath}
Here, we could bound the first term on the RHS with \blemma \ref{lemma_FC_network_inner_product}. Applying \blemma \ref{lemma_FC_network_output_and_gradient} on the second term, and recalling $\norm{\widehat{\matr{\Theta}}_{u}^{(1)} - [\matr{\Theta}_{u}^{(1)}]_{0}}_{2} \leq \omega$, would give
\begin{displaymath}
\begin{split}
    \abs{f_{u}^{(1)} (\vect{x}; \widehat{\matr{\Theta}}_{u}^{(1)})} &\leq 
    2 +  
    \norm{\nabla_{[\widehat{\matr{\Theta}}_{u}^{(1)}]_{0}} f_{u}^{(1)} (\vect{x}; [\widehat{\matr{\Theta}}_{u}^{(1)}]_{0})}_{2}  \norm{\widehat{\matr{\Theta}}_{u}^{(1)} - [\matr{\Theta}_{u}^{(1)}]_{0}}_{2} + \\
    &\qquad \mathcal{O}(\omega^{1/3}L^{2}\sqrt{m\log(m)}) \cdot \norm{\widehat{\matr{\Theta}}_{u}^{(1)} - [\matr{\Theta}_{u}^{(1)}]_{0}}_{2} \\
    & \leq 2 + \mathcal{O}(L) \cdot \norm{\widehat{\matr{\Theta}}_{u}^{(1)} - [\matr{\Theta}_{u}^{(1)}]_{0}}_{2}
    + \mathcal{O}(L^{2}\sqrt{m\log(m)}) (\norm{\widehat{\matr{\Theta}}_{u}^{(1)} - [\matr{\Theta}_{u}^{(1)}]_{0}}_{2})^{\frac{4}{3}}.
\end{split}
\end{displaymath}
Then, with $T_{u, t} = \frac{t}{n}$, applying the conclusion of \blemma \ref{lemma_convergence_user_f_1} would lead to 
\begin{displaymath}
\begin{split}
    \abs{f_{u}^{(1)} (\vect{x}; \widehat{\matr{\Theta}}_{u}^{(1)})} &\leq 
    2 + \mathcal{O}(L) \cdot  \mathcal{O} \left( \frac{ (T_{u, t})^3}{ \rho \sqrt{m}} \log m \right)
    + \mathcal{O}(L^{2}\sqrt{m\log(m)}) \left( \mathcal{O} ( \frac{ (T_{u, t})^3}{ \rho \sqrt{m}} \log m ) \right)^{\frac{4}{3}} \\
    & = 2 + \mathcal{O} \left( \frac{ t^3 L}{ n^{3} \rho\sqrt{m}} \log m \right)
    + \mathcal{O} \left(  \frac{ L^{2}t^4}{ n^{4} \rho^{4 / 3} m^{1 / 6}} \log^{11 / 6} (m) \right) = \gamma_{1}.
\end{split}
\end{displaymath}
\qed

% -------------------------------------
Then, under the assumption of arm separateness (\textbf{Assumption} \ref{assumption_separateness}), we proceed to bound the reward estimation error of the user exploitation network  $f_{u}^{(1)} (\cdot; [\matr{\Theta}_{u}^{(1)}]_{t})$ in the current round $t$.

\begin{lemma} \label{lemma_user_f_1_est_error}
For the constants $\rho \in (0, \mathcal{O}(\frac{1}{L}))$ and $\xi_{1} \in (0, 1)$, given user $u\in\mathcal{U}$ and its past records $\mathcal{P}_{u, t-1}$, we suppose $m, \eta_1, J_1$ satisfy the conditions in \btheorem \ref{theorem_regret_bound}, and randomly draw the parameter $[\matr{\Theta}_{u}^{(1)}]_{t} \sim \{[\widehat{\matr{\Theta}}_{u}^{(1)}]_{\tau}\}_{\tau\in \mathcal{T}_{u, t}}$. Consider the past records $\mathcal{P}_{u, t}$ up to round $t$ are generated by a fixed policy when witness the candidate arms $\{\mathcal{X}_{\tau}\}_{\tau\in \mathcal{T}_{u, t}}$.
Then, with probability at least $1 - \delta$ given an arm-reward pair $(\vect{x}_{t}, r_{t})$, we have
\begin{displaymath}
    \sum_{\tau\in \mathcal{T}_{u, t} \cup \{t\}} \mathbb{E} \big[ \abs{f_{u}^{(1)} (\vect{x}_{\tau}; [\matr{\Theta}_{u}^{(1)}]_{\tau}) - r_{\tau}} ~\big| \mathcal{X}_{\tau} \big] \leq 
    \sqrt{\frac{t}{n}} \cdot \bigg( \sqrt{2\xi_{1}} + \frac{3L}{\sqrt{2}} + (1 + \gamma_{1}) \sqrt{2\log(\frac{tn\cdot a}{\delta})}\bigg)
\end{displaymath}
where $r_{\tau}$ is the corresponding reward generated by the reward mapping function given an arm $\vect{x}_{\tau}$.
\end{lemma}

\textbf{Proof.}
We proof this Lemma following a similar approach as in Lemma C.1 from \citep{EE-Net_ban2021ee} and Lemma D.1 from \citep{Meta-Ban}. First, for the LHS and with $\tau \in \mathcal{T}_{u, t} \cup \{t\}$, we have 
\begin{displaymath}
    \abs{f_{u}^{(1)} (\vect{x}_{t}; [\widehat{\matr{\Theta}}_{u}^{(1)}]_{\tau}) - r_{t}} \leq 
    \abs{f_{u}^{(1)} (\vect{x}_{t}; [\widehat{\matr{\Theta}}_{u}^{(1)}]_{\tau})} + \abs{r_{t}} \leq 1 + \gamma_{1}
\end{displaymath}
based on the conclusion from \blemma \ref{lemma_user_f_1_output_bound}.
Then, for user $u$, we define the following martingale difference sequence with regard to the previous records $\mathcal{P}_{u, \tau}$ up to round $\tau$ as
\begin{displaymath}
    V_{\tau}^{(1)} = \mathbb{E}[ \abs{f_{u}^{(1)}(\vect{x}_{\tau}; [\widehat{\matr{\Theta}}_{u}^{(1)}]_{\tau}) - r_{\tau}} ] 
    - \abs{ f_{u}^{(1)}(\vect{x}_{\tau}; [\widehat{\matr{\Theta}}_{u}^{(1)}]_{\tau}) - r_\tau}.
\end{displaymath}
Since the records in set $\mathcal{P}_{u, \tau}$ are sharing the same reward mapping function, we have the expectation
\begin{displaymath}
    \mathbb{E}[V_{\tau}^{(1)} \big| F_{u, \tau} ] = \mathbb{E}
    [ \abs{f_{u}^{(1)}(\vect{x}_{\tau}; [\widehat{\matr{\Theta}}_{u}^{(1)}]_{\tau}) - r_{\tau}} ] - 
    \mathbb{E}[\abs{ f_{u}^{(1)}(\vect{x}_{\tau}; [\widehat{\matr{\Theta}}_{u}^{(1)}]_{\tau}) - r_\tau}  \big| F_{u, \tau} ] = 0.
\end{displaymath}
where $F_{u, \tau}$ denotes the filtration given the past records $\mathcal{P}_{u, \tau}$. And we have the mean value of $V_{\tau}^{(1)}$ across different time steps as
\begin{displaymath}
    \frac{1}{T_{u, t}} \sum_{\tau\in \mathcal{T}_{u, t}} [V_{\tau}^{(1)}] = 
    \frac{1}{T_{u, t}} \sum_{\tau\in \mathcal{T}_{u, t}} \mathbb{E}
    [ \abs{f_{u}^{(1)}(\vect{x}_{\tau}; [\widehat{\matr{\Theta}}_{u}^{(1)}]_{\tau}) - r_{\tau}} ] 
    - \frac{1}{T_{u, t}} \sum_{\tau\in \mathcal{T}_{u, t}} \abs{ f_{u}^{(1)}(\vect{x}_{\tau}; [\widehat{\matr{\Theta}}_{u}^{(1)}]_{\tau}) - r_\tau}.
\end{displaymath}
with the expectation of zero.
Then, we proceed to bound the expected estimation error of the exploitation model with the estimation error from existing samples
following the Proposition 1 from \citep{general_bound_online-cesa2004generalization}. Applying the Azuma-Hoeffding inequality, with a constant $\delta \in (0, 1)$, it leads to
\begin{displaymath}
\begin{split}
    \mathbb{P}\bigg[ 
        \frac{1}{T_{u, t}} &\sum_{\tau\in \mathcal{T}_{u, t}} \mathbb{E}
        [ \abs{f_{u}^{(1)}(\vect{x}_{\tau}; [\widehat{\matr{\Theta}}_{u}^{(1)}]_{\tau}) - r_{t}} ] - 
        \frac{1}{T_{u, t}} \sum_{\tau\in \mathcal{T}_{u, t}} \abs{ f_{u}^{(1)}(\vect{x}_{\tau}; [\widehat{\matr{\Theta}}_{u}^{(1)}]_{\tau}) - r_\tau} \\
        & \geq (1+\gamma_{1}) \cdot \sqrt{\frac{2}{T_{u, t}}\ln(\frac{1}{\delta})}
    \bigg] \leq \delta.
\end{split}  
\end{displaymath}
As we have the parameter $[\matr{\Theta}_{u}^{(1)}]_{t} \sim \{[\widehat{\matr{\Theta}}_{u}^{(1)}]_{\tau}\}_{\tau\in \mathcal{T}_{u, t}}$, with the probability at least $1-\delta$, the expected loss on $[\matr{\Theta}_{u}^{(1)}]_{t}$ could be bounded as
\begin{displaymath}
\begin{split}
    \frac{1}{T_{u, t}} \sum_{\tau\in \mathcal{T}_{u, t}} \mathbb{E}
        [ \abs{f_{u}^{(1)}(\vect{x}_{t}; [\widehat{\matr{\Theta}}_{u}^{(1)}]_{\tau}) - r_{t}} ] 
    \leq (1+\gamma_{1}) \cdot \sqrt{\frac{2}{T_{u, t}}\ln(\frac{1}{\delta})} 
        + \bigg(\frac{1}{T_{u, t}} \sum_{\tau\in \mathcal{T}_{u, t}} \abs{ f_{u}^{(1)}(\vect{x}_{\tau}; [\widehat{\matr{\Theta}}_{u}^{(1)}]_{\tau}) - r_\tau} \bigg)
\end{split}  
\end{displaymath}
where for the second term on the RHS, we have
\begin{displaymath}
\begin{split}
    \frac{1}{T_{u, t}} \sum_{\tau\in \mathcal{T}_{u, t}} & \abs{ f_{u}^{(1)}(\vect{x}_{\tau}; [\widehat{\matr{\Theta}}_{u}^{(1)}]_{\tau}) - r_\tau} \leq
    \frac{1}{T_{u, t}} \sum_{\tau\in \mathcal{T}_{u, t}} \abs{ f_{u}^{(1)}(\vect{x}_{\tau}; [\widehat{\matr{\Theta}}_{u}^{(1)}]_{t}) - r_\tau} + 
    \frac{3L\sqrt{2T_{u, t}}}{2} \cdot \frac{1}{T_{u, t}} \\
    & \leq \frac{1}{T_{u, t}} \sqrt{T_{u, t}\cdot \sum_{\tau\in \mathcal{T}_{u, t}} \abs{ f_{u}^{(1)}(\vect{x}_{\tau}; [\widehat{\matr{\Theta}}_{u}^{(1)}]_{t}) - r_\tau}^{2}} + 
    \frac{3L}{\sqrt{2T_{u, t}}} \\
    & \leq \sqrt{\frac{\xi_{1}}{T_{u, t}}} + 
    \frac{3L}{\sqrt{2T_{u, t}}} 
\end{split}  
\end{displaymath}
where the first inequality is the application of \blemma \ref{lemma_FC_network_bound_sampled_parameters}, and the last inequality is due to \blemma \ref{lemma_convergence_user_f_1}. Summing up all the components and applying the union bound for all $a$ arms, all $n$ users and $t$ time steps would complete the proof.
\qed

% -------------------------------------
Then, we also have the following Corollary for the rest of the candidate arms $\vect{x}_{i, t} \in \big( \mathcal{X}_{t} \setminus \{\vect{x}_{t}\} \big)$.

\begin{corollary} \label{corollary_user_f_1_est_error_other_arms}
For the constants $\rho \in (0, \mathcal{O}(\frac{1}{L}))$ and $\xi_{1} \in (0, 1)$, given user $u\in\mathcal{U}$ and its past records $\mathcal{P}_{u, t-1}$, we suppose $m, \eta_1, J_1$ satisfy the conditions in \btheorem \ref{theorem_regret_bound}, and randomly draw the parameter $[\matr{\Theta}_{u}^{(1)}]_{t} \sim \{[\widehat{\matr{\Theta}}_{u}^{(1)}]_{\tau}\}_{\tau\in \mathcal{T}_{u, t}}$. For an arm $\vect{x}_{i, t} \in \mathcal{X}_{t}$, consider its union set with the the collection of arms $\widetilde{\mathcal{P}}_{u, t} \cup \{\vect{x}_{i, t}, r_{i, t}\}$ are generated by a fixed policy when witness the candidate arms $\{\mathcal{X}_{\tau}\}_{\tau\in \mathcal{T}_{u, t}}$, with $\widetilde{\mathcal{P}}_{u, t} = \{\vect{x}_{i_{\tau}, \tau}, r_{i_{\tau}, \tau}\}_{\tau}$ being the collection of arms chosen by this policy.
Then, with probability at least $1 - \delta$, we have
\begin{displaymath}
\begin{split}
    \sum_{\tau\in \mathcal{T}_{u, t}\cup \{t\}} \mathbb{E}  \big[& \abs{f_{u}^{(1)} (\vect{x}_{i_{\tau}, \tau}; [\matr{\Theta}_{u}^{(1)}]_{t}) - r_{i_{\tau}, \tau}} ~\big| \mathcal{X}_{\tau} \big] \leq 
    \sqrt{T_{u, t}} \cdot \bigg( \sqrt{2\xi_{1}} + \frac{3L}{\sqrt{2}} + (1 + \gamma_{1}) \sqrt{2\log(\frac{tn\cdot a}{\delta})}\bigg) + \Gamma_{t}
\end{split}
\end{displaymath}
where $r_{i, \tau}$ is the corresponding reward generated by the mapping function given an arm $\vect{x}_{i, \tau}$, and 
\begin{displaymath}
\begin{split}
    \Gamma_{t} \leq 
    \bigg(1 + \mathcal{O}(\frac{tL^{3} \log^{5/6} (m)}{\rho^{1/3}m^{1/6}})\bigg)\cdot \mathcal{O}(\frac{t^{4}L}{\rho\sqrt{m}}\log(m)) + 
    \mathcal{O}\bigg( \frac{t^{5}L^{2} \log^{11/6} (m)}{\rho^{4/3}m^{1/6}} \bigg).
\end{split}  
\end{displaymath}
\end{corollary}

\textbf{Proof.}
The proof of this Corollary follows an analogous approach as in Lemma \ref{lemma_user_f_1_est_error}. 
First, suppose a shadow model $f_{u}^{(1)} (\cdot; [\widetilde{\matr{\Theta}}_{u}^{(1)}]_{t})$, which is trained on the alternative trajectory $\widetilde{\mathcal{P}}_{u, t}$. Analogous to the proof of Lemma \ref{lemma_user_f_1_est_error}, for user $u$, we can define the following martingale difference sequence with regard to the previous records $\widetilde{\mathcal{P}}_{u, \tau}$ up to round $\tau\in [t]$ as
\begin{displaymath}
    \tilde{V}_{\tau}^{(1)} = \mathbb{E}[ \abs{f_{u}^{(1)}(\vect{x}_{i_{\tau}, \tau}; [\widetilde{\matr{\Theta}}_{u}^{(1)}]_{\tau}) - r_{\tau}} ] 
    - \abs{ f_{u}^{(1)}(\vect{x}_{i_{\tau}, \tau}; [\widetilde{\matr{\Theta}}_{u}^{(1)}]_{\tau}) - r_{i_{\tau}, \tau}}.
\end{displaymath}
Since the records in set $\widetilde{\mathcal{P}}_{u, \tau}$ are sharing the same reward mapping function, we have the expectation
\begin{displaymath}
    \mathbb{E}[\tilde{V}_{\tau}^{(1)} \big| \tilde{F}_{u, \tau} ] = \mathbb{E}
    [ \abs{f_{u}^{(1)}(\vect{x}_{i_{\tau}, \tau}; [\widetilde{\matr{\Theta}}_{u}^{(1)}]_{\tau}) - r_{i_{\tau}, \tau}} ] - 
    \mathbb{E}[\abs{ f_{u}^{(1)}(\vect{x}_{i_{\tau}, \tau}; [\widetilde{\matr{\Theta}}_{u}^{(1)}]_{\tau}) - r_{i_{\tau}, \tau}}  \big| \tilde{F}_{u, \tau} ] = 0.
\end{displaymath}
where $\tilde{F}_{u, \tau}$ denotes the filtration given the past records $\widetilde{\mathcal{P}}_{u, \tau}$. The mean value of $\tilde{V}_{\tau}^{(1)}$ across different time steps will be
\begin{displaymath}
    \frac{1}{T_{u, t}} \sum_{\tau\in \mathcal{T}_{u, t}} [\tilde{V}_{\tau}^{(1)}] = 
    \frac{1}{T_{u, t}} \sum_{\tau\in \mathcal{T}_{u, t}} \mathbb{E}
    [ \abs{f_{u}^{(1)}(\vect{x}_{i_{\tau}, \tau}; [\widetilde{\matr{\Theta}}_{u}^{(1)}]_{\tau}) - r_{i_{\tau}, \tau}} ] 
    - \frac{1}{T_{u, t}} \sum_{\tau\in \mathcal{T}_{u, t}} \abs{ f_{u}^{(1)}(\vect{x}_{i_{\tau}, \tau}; [\widetilde{\matr{\Theta}}_{u}^{(1)}]_{\tau}) - r_{i_{\tau}, \tau}}.
\end{displaymath}
with the expectation of zero. Afterwards, applying the Azuma-Hoeffding inequality, with a constant $\delta \in (0, 1)$, it leads to
\begin{displaymath}
\begin{split}
    \mathbb{P}\bigg[ 
        \frac{1}{T_{u, t}} &\sum_{\tau\in \mathcal{T}_{u, t}} \mathbb{E}
        [ \abs{f_{u}^{(1)}(\vect{x}_{i_{\tau}, \tau}; [\widetilde{\matr{\Theta}}_{u}^{(1)}]_{\tau}) - r_{i_{\tau}, \tau}} ] - 
        \frac{1}{T_{u, t}} \sum_{\tau\in \mathcal{T}_{u, t}} \abs{ f_{u}^{(1)}(\vect{x}_{i_{\tau}, \tau}; [\widetilde{\matr{\Theta}}_{u}^{(1)}]_{\tau}) - r_{i_{\tau}, \tau}} \\
        & \geq (1+\gamma_{1}) \cdot \sqrt{\frac{2}{T_{u, t}}\ln(\frac{1}{\delta})}
    \bigg] \leq \delta.
\end{split}  
\end{displaymath}
To bound the output difference between the shadow model $f_{u}^{(1)} (\cdot; [\widetilde{\matr{\Theta}}_{u}^{(1)}]_{t})$ and the model we trained based on received records $f_{u}^{(1)} (\cdot; [\widehat{\matr{\Theta}}_{u}^{(1)}]_{t})$, we apply the conclusion from \textbf{Lemma} \ref{lemma_FC_network_different_parameters}, which leads to that given the same input $\vect{x}$, we have
\begin{displaymath}
\begin{split}
    \abs{f_{u}^{(1)} (\vect{x}; [\widetilde{\matr{\Theta}}_{u}^{(1)}]_{t}) & - f_{u}^{(1)} (\vect{x}; [\widehat{\matr{\Theta}}_{u}^{(1)}]_{t})} \leq \\
    & \bigg(1 + \mathcal{O}(\frac{tL^{3} \log^{5/6} (m)}{\rho^{1/3}m^{1/6}})\bigg)\cdot \mathcal{O}(\frac{t^{3}L}{\rho\sqrt{m}}\log(m)) + 
    \mathcal{O}\bigg( \frac{t^{4}L^{2} \log^{11/6} (m)}{\rho^{4/3}m^{1/6}} \bigg).
\end{split}  
\end{displaymath}
Finally, assembling all the components together will finish the proof.

\qed

% ==================================================================
\subsection{User Exploration Model}

To ensure the unit length of $f_{u}^{(2)}(\cdot)$'s input, we normalize the gradient $\frac{\nabla_{[\matr{\Theta}_{u}^{(1)}]_{t}} f_{u}^{(1)} (\vect{x}; [\matr{\Theta}_{u}^{(1)}]_{t})}{c'_{g}L}$ with \blemma \ref{lemma_FC_network_output_and_gradient}, \blemma \ref{lemma_FC_network_gradient_difference} and a normalization constant $c'_{g} > 0$. 
Then, to satisfy the separateness (\textbf{Assumption} \ref{assumption_separateness}) assumption, we adopt the operation mentioned in \textbf{Eq.} \ref{eq_concat_transformation} to derive the transformation 
$\phi(\frac{\nabla_{[\matr{\Theta}_{u}^{(1)}]_{t}} f_{u}^{(1)} (\vect{x}; [\matr{\Theta}_{u}^{(1)}]_{t})}{c'_{g}L}, \vect{x})$ to make sure the transformed input gradient is of the norm of $1$, and the separateness of at least $\frac{\rho}{\sqrt{2}}$.

% --------------
Analogous to the user exploitation model, regarding the convergence result for FC networks in \blemma \ref{lemma_convergence_user_f_1}, we proceed to present the generalization result of the user exploration model $f_{u}^{(2)}(\cdot)$ after GD with the following lemma.

\begin{lemma} \label{lemma_user_f_2_est_error}
For the constants $c'_{g} > 0$, $\rho \in (0, \mathcal{O}(\frac{1}{L}))$ and $\xi_{1} \in (0, 1)$, given user $u\in\mathcal{U}$ and its past records $\mathcal{P}_{u, t-1}$, we suppose $m, \eta_1, J_1$ satisfy the conditions in \btheorem \ref{theorem_regret_bound}, and randomly draw the parameter $[\matr{\Theta}_{u}^{(2)}]_{t} \sim \{[\widehat{\matr{\Theta}}_{u}^{(2)}]_{\tau}\}_{\tau\in \mathcal{T}_{u, t}}$.
Consider the past records $\mathcal{P}_{u, t}$ up to round $t$ are generated by a fixed policy when witness the candidate arms $\{\mathcal{X}_{\tau}\}_{\tau\in \mathcal{T}_{u, t}}$.
Then, with probability at least $1 - \delta$ given an arm-reward pair $(\vect{x}_{t}, r_{t})$, we have
\begin{displaymath}
\begin{split}
    \sum_{\tau\in \mathcal{T}_{u, t}\cup \{t\}} \mathbb{E} 
    \bigg[ 
    \abs{ f_{u}^{(2)}&\bigg( \phi(\frac{\nabla_{[\matr{\Theta}_{u}^{(1)}]_{\tau}} f_{u}^{(1)} (\vect{x}_{\tau}; [\matr{\Theta}_{u}^{(1)}]_{\tau})}{c'_{g}L}, \vect{x}_{\tau}); [\matr{\Theta}_{u}^{(2)}]_{\tau} \bigg)  - \bigg(r_{\tau} - f_{u}^{(1)} (\vect{x}_{\tau}; [\matr{\Theta}_{u}^{(1)}]_{\tau}) \bigg) } 
    ~\big| \mathcal{X}_{\tau} \bigg] \\
    & \leq 
    \sqrt{T_{u, t}} \cdot \bigg( \sqrt{2\xi_{1}} + \frac{3 L}{\sqrt{2}} + (1 + 2\gamma_{1}) \sqrt{2\log(\frac{tn\cdot a}{\delta})}\bigg)
\end{split}
\end{displaymath}
\end{lemma}

\textbf{Proof.}
The proof of this lemma is inspired by Lemma C.1 from \citep{EE-Net_ban2021ee}. Following the same procedure as in the proof of \blemma \ref{lemma_user_f_1_est_error}, we bound 
\begin{displaymath}
\begin{split}
    \bigg| f_{u}^{(2)}\bigg( & \phi(\frac{\nabla_{[\matr{\Theta}_{u}^{(1)}]_{t}} f_{u}^{(1)} (\vect{x}_{t}; [\matr{\Theta}_{u}^{(1)}]_{t})}{c'_{g}L}, \vect{x}_{t}); [\matr{\Theta}_{u}^{(2)}]_{t} \bigg)  - \bigg(r_{t} - f_{u}^{(1)} (\vect{x}_{t}; [\matr{\Theta}_{u}^{(1)}]_{t}) \bigg) \bigg| \\
     &\qquad \leq  \bigg| f_{u}^{(2)}\bigg( \phi(\frac{\nabla_{[\matr{\Theta}_{u}^{(1)}]_{t}} f_{u}^{(1)} (\vect{x}_{t}; [\matr{\Theta}_{u}^{(1)}]_{t})}{c'_{g}L}, \vect{x}_{t}); [\matr{\Theta}_{u}^{(2)}]_{t} \bigg) \bigg|
     + \bigg| f_{u}^{(1)} (\vect{x}_{t}; [\matr{\Theta}_{u}^{(1)}]_{t})  \bigg|  + 1 \\
     &\qquad \leq  1 + 2\gamma_{1}
\end{split}
\end{displaymath}
by triangle inequality and applying the generalization result of FC networks (\blemma \ref{lemma_user_f_1_output_bound}) on $f_{u}^{(1)}(\cdot; \matr{\Theta}_{u}^{(1)}), f_{u}^{(2)}(\cdot; \matr{\Theta}_{u}^{(2)})$.

For brevity, we use $\nabla f_{u, \tau}^{(1)} (\vect{x}_{t})$ to denote $\phi(\frac{\nabla_{[\matr{\Theta}_{u}^{(1)}]_{\tau}} f_{u}^{(1)} (\vect{x}_{t}; [\matr{\Theta}_{u}^{(1)}]_{\tau})}{c'_{g}L}, \vect{x}_{t})$ for the following proof.
Define the difference sequence as
\begin{displaymath}
\begin{split}
    V_{\tau}^{(2)} = \mathbb{E}
    \bigg[ \bigg| f_{u}^{(2)}\bigg( \nabla f_{u, \tau}^{(1)} (\vect{x}_{\tau})&; [\matr{\Theta}_{u}^{(2)}]_{\tau} \bigg)  - \bigg(r_{\tau} - f_{u}^{(1)} (\vect{x}_{\tau}; [\matr{\Theta}_{u}^{(1)}]_{\tau}) \bigg) \bigg| \bigg] \\
    & - \bigg| f_{u}^{(2)}\bigg( \nabla f_{u, \tau}^{(1)}(\vect{x}_{\tau}); [\matr{\Theta}_{u}^{(2)}]_{\tau} \bigg)  - \bigg(r_{\tau} - f_{u}^{(1)} (\vect{x}_{\tau}; [\matr{\Theta}_{u}^{(1)}]_{\tau}) \bigg) \bigg|.
\end{split}
\end{displaymath}
Since the reward mapping is fixed given the specific user $u$, which means that the past rewards and the received arm-reward pairs $(\vect{x}_{\tau}, r_{\tau})$ are generated by the same reward mapping function, we have the expectation
\begin{displaymath}
\begin{split}
    \mathbb{E}[V_{\tau}^{(2)} \big| F_{u, \tau} ] = & \mathbb{E}
    \bigg[ \bigg| f_{u}^{(2)}\bigg( \nabla f_{u, \tau}^{(1)} (\vect{x}_{\tau}); [\matr{\Theta}_{u}^{(2)}]_{\tau} \bigg)  - \bigg(r_{\tau} - f_{u}^{(1)} (\vect{x}_{\tau}; [\matr{\Theta}_{u}^{(1)}]_{\tau}) \bigg) \bigg| \bigg] \\
    & - \mathbb{E}\bigg[ \bigg| f_{u}^{(2)}\bigg( \nabla f_{u, \tau}^{(1)}(\vect{x}_{\tau}); [\matr{\Theta}_{u}^{(2)}]_{\tau} \bigg)  - \bigg(r_{\tau} - f_{u}^{(1)} (\vect{x}_{\tau}; [\matr{\Theta}_{u}^{(1)}]_{\tau}) \bigg) \bigg| \big| F_{u, \tau} \bigg] = 0.
\end{split}
\end{displaymath}
where $F_{u, \tau}$ denotes the filtration given the past records $\mathcal{P}_{u, \tau}$, up to round $\tau\in [t]$. This also gives the fact that $V_{\tau}^{(2)}$ is a martingale difference sequence.
Then, after applying the martingale difference sequence over $\mathcal{T}_{u, t}$, we have
\begin{displaymath}
\begin{split}
    \frac{1}{T_{u, t}} \sum_{\tau \in \mathcal{T}_{u, t}} V_{\tau}^{(2)} = & \frac{1}{T_{u, t}} \sum_{\tau \in \mathcal{T}_{u, t}} \mathbb{E}
    \bigg[ \bigg| f_{u}^{(2)}\bigg( \nabla f_{u, \tau}^{(1)} (\vect{x}_{\tau}); [\matr{\Theta}_{u}^{(2)}]_{\tau} \bigg)  - \bigg(r_{\tau} - f_{u}^{(1)} (\vect{x}_{\tau}; [\matr{\Theta}_{u}^{(1)}]_{\tau}) \bigg) \bigg| \bigg] \\
    & - \frac{1}{T_{u, t}} \sum_{\tau \in \mathcal{T}_{u, t}} \bigg| f_{u}^{(2)}\bigg( \nabla f_{u, \tau}^{(1)}(\vect{x}_{\tau}); [\matr{\Theta}_{u}^{(2)}]_{\tau} \bigg)  - \bigg(r_{\tau} - f_{u}^{(1)} (\vect{x}_{\tau}; [\matr{\Theta}_{u}^{(1)}]_{\tau}) \bigg) \bigg|. 
\end{split}
\end{displaymath}
Analogous to the proof of \blemma \ref{lemma_user_f_1_est_error}, by applying the Azuma-Hoeffding inequality, it leads to
\begin{displaymath}
\begin{split}
    \mathbb{P} \bigg[ 
    \frac{1}{T_{u, t}} \sum_{\tau \in \mathcal{T}_{u, t}} V_{\tau}^{(2)} - \frac{1}{T_{u, t}}\sum_{\tau \in \mathcal{T}_{u, t}} \mathbb{E} [V_{\tau}^{(2)}] \geq (1+2\gamma_{1}) \sqrt{\frac{2\log(1 / \delta)}{T_{u, t}}}
    \bigg] \leq \delta
\end{split}
\end{displaymath}
Since the expectation of $V_{\tau}^{(2)}$ is zero, with the probability at least $1-\delta$ and an existing set of parameters $\widetilde{\matr{\Theta}}_{u}^{(2)}$ s.t. $\norm{\widetilde{\matr{\Theta}}_{u}^{(2)} - [\matr{\Theta}_{u}^{(2)}]_{\tau}} \leq \mathcal{O} \left( \frac{ t^3}{  n^{3}\rho \sqrt{m}} \log m \right)$, the above inequality implies
\begin{displaymath}
\begin{split}
    & \frac{1}{T_{u, t}} \sum_{\tau \in \mathcal{T}_{u, t}} V_{\tau}^{(2)} \leq (1+2\gamma_{1}) \sqrt{\frac{2\log(1 / \delta)}{T_{u, t}}} \implies \\
    & \frac{1}{T_{u, t}} \sum_{\tau \in \mathcal{T}_{u, t}} \mathbb{E}
    \bigg[ \bigg| f_{u}^{(2)}\bigg( \nabla f_{u, \tau}^{(1)} (\vect{x}_{t}); [\matr{\Theta}_{u}^{(2)}]_{\tau} \bigg)  - \bigg(r_{t} - f_{u}^{(1)} (\vect{x}_{t}; [\matr{\Theta}_{u}^{(1)}]_{\tau}) \bigg) \bigg| \bigg] \\
    & \leq \frac{1}{T_{u, t}} \sum_{\tau \in \mathcal{T}_{u, t}} \bigg| f_{u}^{(2)}\bigg( \nabla f_{u, \tau}^{(1)}(\vect{x}_{\tau}); [\matr{\Theta}_{u}^{(2)}]_{\tau} \bigg)  - \bigg(r_{\tau} - f_{u}^{(1)} (\vect{x}_{\tau}; [\matr{\Theta}_{u}^{(1)}]_{\tau}) \bigg) \bigg| + (1+2\gamma_{1}) \sqrt{\frac{2\log(1 / \delta)}{T_{u, t}}} \\
    & \underset{(\text{i})}{\leq} \frac{1}{T_{u, t}} \sum_{\tau \in \mathcal{T}_{u, t}} \bigg| f_{u}^{(2)}\bigg( \nabla f_{u, \tau}^{(1)}(\vect{x}_{\tau}); \widetilde{\matr{\Theta}}_{u}^{(2)} \bigg)  - \bigg(r_{\tau} - f_{u}^{(1)} (\vect{x}_{\tau}; [\matr{\Theta}_{u}^{(1)}]_{\tau}) \bigg) \bigg| + (1+2\gamma_{1}) \sqrt{\frac{2\log(1 / \delta)}{T_{u, t}}} \\
    & \leq \frac{1}{\sqrt{T_{u, t}}} \sqrt{ \sum_{\tau \in \mathcal{T}_{u, t}} \bigg| f_{u}^{(2)}\bigg( \nabla f_{u, \tau}^{(1)}(\vect{x}_{\tau}); \widetilde{\matr{\Theta}}_{u}^{(2)} \bigg)  - \bigg(r_{\tau} - f_{u}^{(1)} (\vect{x}_{\tau}; [\matr{\Theta}_{u}^{(1)}]_{\tau}) \bigg) \bigg|^{2} } + (1+2\gamma_{1}) \sqrt{\frac{2\log(1 / \delta)}{T_{u, t}}} \\
    & \underset{(\text{ii})}{\leq} \sqrt{\frac{2\xi_{1}}{T_{u, t}}} + (1+2\gamma_{1}) \sqrt{\frac{2\log(1 / \delta)}{T_{u, t}}}.
\end{split}
\end{displaymath}
Here, the upper bound (i) is derived by applying the conclusions of \blemma \ref{lemma_convergence_user_f_1} and \blemma \ref{lemma_FC_network_bound_sampled_parameters}, and the inequality (ii) is derived by adopting \blemma \ref{lemma_convergence_user_f_1} while defining the empirical loss to be $\frac{1}{2} \sum_{\tau \in \mathcal{T}_{u, t}} \bigg| f_{u}^{(2)}\bigg( \nabla f_{u, \tau}^{(1)}(\vect{x}_{\tau}); \widetilde{\matr{\Theta}}_{u}^{(2)} \bigg)  - \bigg(r_{\tau} - f_{u}^{(1)} (\vect{x}_{\tau}; [\matr{\Theta}_{u}^{(1)}]_{\tau}) \bigg) \bigg|^{2}$.
Finally, applying the union bound would give the aforementioned results.

\qed

% ==================================================================
\begin{corollary} \label{corollary_user_f_2_est_error_other_arms}
For the constants $\rho \in (0, \mathcal{O}(\frac{1}{L}))$ and $\xi_{1} \in (0, 1)$, given user $u\in\mathcal{U}$ and its past records $\mathcal{P}_{u, t-1}$, we suppose $m, \eta_1, J_1$ satisfy the conditions in \btheorem \ref{theorem_regret_bound}, and randomly draw the parameter $[\matr{\Theta}_{u}^{(1)}]_{t} \sim \{[\widehat{\matr{\Theta}}_{u}^{(1)}]_{\tau}\}_{\tau\in \mathcal{T}_{u, t}}$. For an arm $\vect{x}_{i, t} \in \mathcal{X}_{t}$, consider its union set with the the collection of arms $\widetilde{\mathcal{P}}_{u, t} \cup \{\vect{x}_{i, t}, r_{i, t}\}$ are generated by a fixed policy when witness the candidate arms $\{\mathcal{X}_{\tau}\}_{\tau\in \mathcal{T}_{u, t}}$, with $\widetilde{\mathcal{P}}_{u, t} = \{\vect{x}_{i_{\tau}, \tau}, r_{i_{\tau}, \tau}\}_{\tau}$ being the collection of arms chosen by this policy.
Then, with probability at least $1 - \delta$, we have
\begin{displaymath}
\begin{split}
    \sum_{\tau\in \mathcal{T}_{u, t}\cup \{t\}} \mathbb{E}  
    \big[& \abs{ f_{u}^{(2)}\bigg( \phi(\frac{\nabla_{[\matr{\Theta}_{u}^{(1)}]_{\tau}} f_{u}^{(1)} (\vect{x}_{i_{\tau}, \tau}; [\matr{\Theta}_{u}^{(1)}]_{\tau})}{c'_{g}L}, \vect{x}_{i_{\tau}, \tau}); [\matr{\Theta}_{u}^{(2)}]_{\tau} \bigg)  - \bigg(r_{i_{\tau}, \tau} - f_{u}^{(1)} (\vect{x}_{i_{\tau}, \tau}; [\matr{\Theta}_{u}^{(1)}]_{\tau}) \bigg) } ~\big| \mathcal{X}_{\tau} \big] \\
    & \leq 
    \sqrt{T_{u, t}} \cdot \bigg( \sqrt{2\xi_{1}} + \frac{3L}{\sqrt{2}} + (1 + \gamma_{1}) \sqrt{2\log(\frac{tn\cdot a}{\delta})}\bigg) + \Gamma_{t}
\end{split}
\end{displaymath}
where $r_{i_{\tau}, \tau}$ is the corresponding reward generated by the mapping function given an arm $\vect{x}_{i_{\tau}, \tau}$, and 
\begin{displaymath}
\begin{split}
    \Gamma_{t} =
    \bigg(1 + \mathcal{O}(\frac{tL^{3} \log^{5/6} (m)}{n\rho^{1/3}m^{1/6}})\bigg)\cdot \mathcal{O}(\frac{t^{4}L}{n^{4}\rho\sqrt{m}}\log(m)) + 
    \mathcal{O}\bigg( \frac{t^{5}L^{2} \log^{11/6} (m)}{n^{5}\rho^{4/3}m^{1/6}} \bigg).
\end{split}  
\end{displaymath}
\end{corollary}
This corollary is the direct application of Lemma \ref{lemma_user_f_2_est_error}, and the proof is analogous to that of Corollary \ref{corollary_user_f_1_est_error_other_arms}.

% ==================================================================
\subsection{Lemmas for Over-parameterized User Networks} \label{subsec_user_over_param_nets}

Applying $\mathcal{P}_{u, t-1}$ as the training data, we have the following convergence result for the user exploitation network $f_{u}^{(1)}(\cdot; \matr{\Theta}_{u}^{(1)})$ after GD.

\begin{lemma} [Theorem 1 from \citep{conv_theory-allen2019convergence}] \label{lemma_convergence_user_f_1}
For any $ 0 < \xi_1 \leq 1$, $ 0 < \rho \leq \mathcal{O}(\frac{1}{L})$. Given user $u\in\mathcal{U}$ and its past records $\mathcal{P}_{u, t-1}$, suppose $m, \eta_1, J_1$ satisfy the conditions in \btheorem \ref{theorem_regret_bound}, then with probability at least $1 - \delta$, we could have
\begin{enumerate}
    \item $\mathcal{L}(\matr{\Theta}_{u}^{(1)}) \leq \xi_{1}$ after $J_1$ iterations of GD.
    \item For any $j \in [J_1]$, $\norm{ [\matr{\Theta}_{u}^{(1)}]^{j}  -  [\matr{\Theta}_{u}^{(1)}]^{0}  } \leq \mathcal{O} \left( \frac{ (T_{u, t})^3}{ \rho \sqrt{m}} \log m \right)
    = \mathcal{O} \left( \frac{ t^3}{  n^{3}\rho \sqrt{m}} \log m \right)$.
\end{enumerate}
\end{lemma}
In particular, \blemma \ref{lemma_convergence_user_f_1} above provides the convergence guarantee for $f_{u}^{(1)}(\cdot; \matr{\Theta}_{u}^{(1)})$ after certain rounds of GD training on the past records $\mathcal{P}_{u, t-1}$.

% -------------------
\begin{lemma} [Lemma 4.1 in \citep{generalization_bound_cao2019generalization}] \label{lemma_FC_network_inner_product}
Assume a constant $\omega$ such that $ \mathcal{O}( m^{-3/2} L^{-3/2} [ \log (TnL^2/\delta) ]^{3/2}  )   \leq \omega \leq  \mathcal{O} (L^{-6}[\log m]^{-3/2} )$ and $n$ training samples. With randomly initialized $[\matr{\Theta}_{u}^{(1)}]_{0}$, for parameters $\matr{\Theta}, \matr{\Theta}'$ satisfying $\norm{\matr{\Theta} - [\matr{\Theta}_{u}^{(1)}]_{0}}, \norm{\matr{\Theta} - [\matr{\Theta}_{u}^{(1)}]_{0}} \leq \omega$, we have
\begin{displaymath}
    \abs{f_{u}^{(1)} (\vect{x}; \matr{\Theta}) -  f_{u}^{(1)} (\vect{x}; \matr{\Theta}') - \inp{\nabla_{\matr{\Theta}'} f_{u}^{(1)} (\vect{x}; \matr{\Theta}')}{\matr{\Theta} - \matr{\Theta}'}} \leq 
    \mathcal{O}(\omega^{1/3}L^{2}\sqrt{m\log(m)}) \norm{\matr{\Theta} - \matr{\Theta}'}
\end{displaymath}
with the probability at least $1-\delta$.
\end{lemma}

% -------------------
\begin{lemma} \label{lemma_FC_network_output_and_gradient}
Assume $m, \eta_1, J_1$ satisfy the conditions in \btheorem \ref{theorem_regret_bound} and $[\matr{\Theta}_{u}^{(1)}]_{0}$ being randomly initialized.
Then, with probability at least $1 - \delta$ and given an arm $\norm{\vect{x}}_{2}=1$, we have
\begin{enumerate}
    \item $\abs{ f_{u}^{(1)} (\vect{x}; [\matr{\Theta}_{u}^{(1)}]_{0}) } \leq 2$,
    \item $\norm{\nabla_{[\matr{\Theta}_{u}^{(1)}]_{0}} f_{u}^{(1)} (\vect{x}; [\matr{\Theta}_{u}^{(1)}]_{0})}_{2} \leq \mathcal{O}(L) $.
\end{enumerate}
\end{lemma}

\textbf{Proof.}
The conclusion (1) is a direct application of Lemma 7.1 in \citep{conv_theory-allen2019convergence}. For conclusion (2), applying Lemma 7.3 in \citep{conv_theory-allen2019convergence}, for each layer $\matr{\Theta}_{l} \in \{\matr{\Theta}_{1}, \dots, \matr{\Theta}_{L}\}$, we have 
\begin{displaymath}
    \norm{\nabla_{\matr{\Theta}_{l}} f_{u}^{(1)} (\vect{x}; [\matr{\Theta}_{u}^{(1)}]_{0})}_{2} = \norm{ (\matr{\Theta}_{L} \matr{D}_{L-1} \cdots \matr{D}_{l+1}\matr{\Theta}_{l+1})\cdot (\matr{D}_{l+1} \matr{\Theta}_{l+1} \cdots \matr{D}_{1}\matr{\Theta}_{1}) \cdot \vect{x}^{\intercal} }_{2} = \mathcal{O}(\sqrt{L}).
\end{displaymath}
Then, we could have the conclusion that 
\begin{displaymath}
    \norm{\nabla_{[\matr{\Theta}_{u}^{(1)}]_{0}} f_{u}^{(1)} (\vect{x}; [\matr{\Theta}_{u}^{(1)}]_{0})}_{2} = \sqrt{\sum_{l\in[L]} \norm{\nabla_{\matr{\Theta}_{l}} f_{u}^{(1)} (\vect{x}; [\matr{\Theta}_{u}^{(1)}]_{0})}_{2}^
    {2} } = \mathcal{O}(L).
\end{displaymath}

\qed

% -------------------
\begin{lemma} [Theorem 5 in \citep{conv_theory-allen2019convergence}] \label{lemma_FC_network_gradient_difference}
Assume $m, \eta_1, J_1$ satisfy the conditions in \btheorem \ref{theorem_regret_bound} and $[\matr{\Theta}_{u}^{(1)}]_{0}$ being randomly initialized.
Then, with probability at least $1 - \delta$, and for all parameter $\matr{\Theta}_{u}^{(1)}$ such that $\norm{\matr{\Theta}_{u}^{(1)} - [\matr{\Theta}_{u}^{(1)}]_{0}}_{2} \leq \omega $, we have
\begin{displaymath}
    \norm{ \nabla_{\matr{\Theta}_{u}^{(1)}} f_{u}^{(1)} (\vect{x}; \matr{\Theta}_{u}^{(1)}) - \nabla_{[\matr{\Theta}_{u}^{(1)}]_{0}} f_{u}^{(1)} (\vect{x}; [\matr{\Theta}_{u}^{(1)}]_{0}) }_{2} \leq
    \mathcal{O}(\omega^{1/3}L^{3}\sqrt{\log(m)})
\end{displaymath}
\end{lemma}

% -------------------
\begin{lemma} \label{lemma_FC_network_bound_sampled_parameters}

Assume $m, \eta_{1}$ satisfy the condition in \btheorem \ref{theorem_regret_bound}. 
With the probability at least $1-\delta$, we have

\begin{displaymath}
\begin{split}
    \sum_{\tau\in \mathcal{T}_{u, t}} & \abs{ f_{u}^{(1)}(\vect{x}_{\tau}; [\widehat{\matr{\Theta}}_{u}^{(1)}]_{\tau}) - r_\tau} \leq
    \sum_{\tau\in \mathcal{T}_{u, t}} \abs{ f_{u}^{(1)}(\vect{x}_{\tau}; [\widehat{\matr{\Theta}}_{u}^{(1)}]_{t}) - r_\tau} + 
    \frac{3L\sqrt{2T_{u, t}}}{2} 
\end{split}  
\end{displaymath}

\end{lemma}

\textbf{Proof.}
With the notation from Lemma 4.3 in \citep{generalization_bound_cao2019generalization}, set $R=\frac{T_{u, t}^{3}\log(m)}{\delta}$, $\nu=R^{2}$, and $\epsilon=\frac{LR}{\sqrt{2\nu T_{u, t}}}$. Then, 
considering the loss function to be $\mathcal{L}(\matr{\Theta}_{u}^{(1)}) := \sum_{\tau\in \mathcal{T}_{u, t}} \abs{ f_{u}^{(1)}(\vect{x}_{\tau}; \matr{\Theta}_{u}^{(1)}) - r_\tau}$ would complete the proof.
\qed

\section{Proof of the Regret Bound}     \label{sec_appd_analysis_GNN}

In this section, we present the generalization results of GNN models $f_{gnn}^{(1)}(\cdot; \matr{\Theta}_{gnn}^{(1)}), f_{gnn}^{(2)}(\cdot; \matr{\Theta}_{gnn}^{(2)})$. 
Recall that up to round $t$, we have all the past arm-reward pairs $\mathcal{P}_{t} = \{(\vect{x}_{\tau}, r_{\tau})\}_{\tau\in [t-1]}$ for the previous $t-1$ time steps. 
Analogous to the generalization analysis of user models in Section \ref{sec_appd_analysis_user}, we adopt the the operation in \textbf{Eq.} \ref{eq_concat_transformation} on the gradients $\nabla_{\matr{\Theta}_{gnn}^{(1)}} f_{gnn}^{(1)}(\cdot; \matr{\Theta}_{gnn}^{(1)})$ to comply with the assumptions of unit-length and separateness, and the transformed gradient input is denoted as $\nabla f^{(1)} (\vect{x})$ given the arm $\vect{x}$.

% --------------------

\subsection{Bounding the Parameter Estimation Error}    \label{subsec_bounding_I_1}

Regarding \textbf{Eq.}\ref{eq_single_CB}, given an arbitrary candidate arm $\vect{x} \in \mathcal{X}_{t}$ with its reward $r$, and its user graphs $\mathcal{G}^{(1)}, \mathcal{G}^{(2)}$, we have the bound for the estimation error as
\begin{displaymath}
\begin{split}
    & \mathsf{CB}_{t}(\vect{x})  = \mathbb{E} \bigg[ \abs{f_{gnn}^{(2)}(\nabla f_{t}^{(1)} (\vect{x}),~\mathcal{G}^{(2)}; [\matr{\Theta}_{gnn}^{(2)}]_{t-1}) - (r_{t} -  f_{gnn}^{(1)}(\vect{x},~\mathcal{G}^{(1)}; [\matr{\Theta}_{gnn}^{(1)}]_{t-1}))}  \bigg| u_{t}, \mathcal{X}_{t} \bigg] \\
    & \leq \underbrace{\mathbb{E} \bigg[ \abs{f_{gnn}^{(2)}(\nabla f_{t}^{(1), *} (\vect{x}),~\mathcal{G}^{(2), *}; [\matr{\Theta}_{gnn}^{(2)}]_{t-1}) - (r_{t} -  f_{gnn}^{(1)}(\vect{x},~\mathcal{G}^{(1), *}; [\matr{\Theta}_{gnn}^{(1)}]_{t-1}))}  \bigg| u_{t}, \mathcal{X}_{t} \bigg]}_{I_{1}} \\
    & \quad + I_{2} + I_{3} + I_{4}
\end{split}
\end{displaymath}
where we have the term $I_{1}$ representing the estimation error induced by the GNN model parameters $\{[\matr{\Theta}_{gnn}^{(1)}]_{t-1}, [\matr{\Theta}_{gnn}^{(2)}]_{t-1}\}$.  
Based on our arm selected strategy given in \textbf{Algorithm} \ref{algo_main}, we have the selected arms and their rewards $\{\vect{x}_{\tau}, r_{\tau}\}_{\tau\in [t-1]}$ up to round $t$. And we first proceed to bound term $I_{1}$ w.r.t. the selected arm $\vect{x}_{t}$, i.e., $\textsc{CB}_{t}(\vect{x}_{t})$.

Analogous to the user-specific models, we also have bounded outputs for the GNN models shown in the following lemma.

\begin{lemma} \label{lemma_GNN_f_1_output_bound}
For the constants $\rho \in (0, \mathcal{O}(\frac{1}{L}))$ and $\xi_{2} \in (0, 1)$, the past records $\mathcal{P}_{t}$ up to time step $t$, we suppose $m, \eta_{1}, \eta_{2}, J_{1}, J_{2}$ satisfy the conditions in \btheorem \ref{theorem_regret_bound}.
Then, with probability at least $1 - \delta$ and given an arm-reward pair $(\vect{x}, r)$, we have
\begin{displaymath}
    \abs{f_{gnn}^{(1)} (\vect{x}; [\widehat{\matr{\Theta}}_{gnn}^{(1)}]_{t})} \leq \gamma_{2}
\end{displaymath}
where 
\begin{displaymath}
    \gamma_{2} = 
    2 + \mathcal{O} \left( \frac{ t^3 L}{ \rho\sqrt{m}} \log m \right)
    + \mathcal{O} \left(  \frac{ L^{2}t^4 }{ \rho^{4 / 3} m^{1 / 6}} \log^{11 / 6} (m) \right).
\end{displaymath}
\end{lemma}

\textbf{Proof.}
The proof of this lemma follows an analogous approach as in \blemma \ref{lemma_user_f_1_output_bound} where we have proved the conclusion for the FC networks. 
Given an arm $\vect{x}$, we denote the adjacency matrix of its estimated user graph $\mathcal{G}^{(1)}$ as $\matr{A}^{(1)}$, and we have the normalized adjacency matrices as $\matr{S}^{(1)} = \matr{A}^{(1)} / n$.
For the received user $u_{t}\in \mathcal{U}$, we could deem the corresponding row of the matrix multiplication $\matr{S} \cdot \matr{X}$, represented by $\vect{h}_{u_{t}} = [\matr{S} \cdot \matr{X}]_{i:}$, as the aggregated input for the network for the user-arm pair $(\vect{x}, u_{t})$. Note that in this way, the rest of the network could be regarded as a $L+1$-layer FC network (one layer GNN + $L$-layer FC network), where the weight matrix of the first layer is $\matr{\Theta}_{agg}^{(1)}$. 
Then, to make sure each aggregated input has the norm of 1, we apply an additional transformation mentioned in \textbf{Eq.} \ref{eq_concat_transformation}
as $\Tilde{\vect{h}}_{u_{t}} = \phi(\vect{h}_{u_{t}}, \vect{x}) = 
    (\frac{\vect{h}_{u_{t}}}{\sqrt{2}}, \frac{\vect{x}}{2}, c_{u_{t}})$ where $c_{u_{t}} = \sqrt{\frac{3}{4} - \frac{1}{2}\norm{\vect{h}_{u_{t}}}_{2}^{2}}$. 
This transformation ensures $\norm{\Tilde{\vect{h}}_{u_{t}}}_{2} = 1$ while preserving the original information w.r.t. the user-arm pair $(\vect{x}, u_{t})$, as it does not change the original aggregated hidden representation. Meantime, this transformation also ensures the separateness of the transformed contexts to be at least $\frac{\rho}{2}$, which would fit the original data separateness assumption (\textbf{Assumption} \ref{assumption_separateness}). Finally, following a similar approach as in the FC networks (\blemma \ref{lemma_user_f_1_output_bound}), on the transformed aggregated hidden representations would complete the proof. 

\qed

% ----------------------------------------------------------

Regarding the definition for the true reward mapping function in Section \ref{sec_problem_def_notation}, we have the following lemma for term $I_{1}$ given the arm-reward pair $(\vect{x}_{t}, r_{t})$.

\begin{lemma} \label{lemma_term_I_1_selected_arm}

For the constants $\rho \in (0, \mathcal{O}(\frac{1}{L}))$ and $\xi_{2} \in (0, 1)$, given user $u\in\mathcal{U}$ and its past records $\mathcal{P}_{u, t}$, we suppose $m, \eta_{1}, \eta_{2}, J_{1}, J_{2}$ satisfy the conditions in \btheorem \ref{theorem_regret_bound}, and randomly draw the parameters $[\matr{\Theta}_{gnn}^{(1)}]_{t} \sim \{[\widehat{\matr{\Theta}}_{gnn}^{(1)}]_{\tau}\}_{\tau\in [t]}$, $[\matr{\Theta}_{gnn}^{(2)}]_{t} \sim \{[\widehat{\matr{\Theta}}_{gnn}^{(2)}]_{\tau}\}_{\tau\in [t]}$.
Then, with probability at least $1 - \delta$ given a sampled arm-reward pair $(\vect{x}, r)$, we have
\begin{displaymath}
\begin{split}
    & \sum_{\tau\in [t]} \mathbb{E} 
    \bigg[ \bigg| f_{gnn}^{(2)}\bigg(\nabla f_{t}^{(1), *} (\vect{x}_{t}),~\mathcal{G}_{t}^{(2), *}; [\matr{\Theta}_{gnn}^{(2)}]_{t-1}\bigg) - \bigg(r_{t} -  f_{gnn}^{(1)}(\vect{x}_{t},~\mathcal{G}_{t}^{(1), *}; [\matr{\Theta}_{gnn}^{(1)}]_{t-1}) \bigg) \bigg|  | u_{t}, \mathcal{X}_{t} \bigg] \\
    & \quad \leq 
    \sqrt{t} \cdot \bigg( \sqrt{2\xi_{2}} + \frac{3L}{\sqrt{2}} + (1 + \gamma_{2}) \sqrt{2\log(\frac{tn\cdot a}{\delta})}\bigg)
\end{split}
\end{displaymath}
where 
\begin{displaymath}
    \gamma_{2} = 
    2 + \mathcal{O} \left( \frac{ t^3 L}{  \rho\sqrt{m}} \log m \right)
    + \mathcal{O} \left(  \frac{ L^{2}t^4 }{ \rho^{4 / 3} m^{1 / 6}} \log^{11 / 6} (m) \right).
\end{displaymath}

\end{lemma}

\textbf{Proof.}
Based on the conclusion of \blemma\ref{lemma_GNN_f_1_output_bound}, we have the upper bound as
\begin{displaymath}
\begin{split}
    & \bigg| f_{gnn}^{(2)}\bigg(\nabla f_{t}^{(1), *} (\vect{x}_{t}),~\mathcal{G}_{t}^{(2), *}; [\matr{\Theta}_{gnn}^{(2)}]_{t-1} \bigg) - \bigg( r_{t} -  f_{gnn}^{(1)}(\vect{x}_{t},~\mathcal{G}_{t}^{(1), *}; [\matr{\Theta}_{gnn}^{(1)}]_{t-1}) \bigg) \bigg| \leq 1+2\gamma_{2}
\end{split}
\end{displaymath}
by simply using the triangular inequality. Then we proceed to define the sequence $V_{\tau}, \tau \in [t]$ as
\begin{displaymath}
\begin{split}
    V_{\tau} = & \mathbb{E}_{ \mathcal{X}_{\tau}}
    \bigg[ \bigg| f_{gnn}^{(2)}(\nabla f_{\tau}^{(1), *} (\vect{x}_{\tau}),~\mathcal{G}_{\tau}^{(2), *}; [\matr{\Theta}_{gnn}^{(2)}]_{\tau-1}) - (r_{\tau} -  f_{gnn}^{(1)}(\vect{x}_{\tau},~\mathcal{G}_{\tau}^{(1), *}; [\matr{\Theta}_{gnn}^{(1)}]_{\tau-1})) \bigg| \bigg] \\
    & - \bigg| f_{gnn}^{(2)}(\nabla f_{\tau}^{(1), *} (\vect{x}_{\tau}),~\mathcal{G}_{\tau}^{(2), *}; [\matr{\Theta}_{gnn}^{(2)}]_{\tau-1}) - (r_{\tau} -  f_{gnn}^{(1)}(\vect{x}_{\tau},~\mathcal{G}_{\tau}^{(1), *}; [\matr{\Theta}_{gnn}^{(1)}]_{\tau-1})) \bigg|.
\end{split}
\end{displaymath}
And since the candidate arms and the corresponding rewards are associated with the same reward mapping function $h(\cdot)$, the sequence $V_{\tau}$ is a martingale difference sequence with the expectation
\begin{displaymath}
\begin{split}
    \mathbb{E}[V_{\tau} & \big| F_{\tau} ] = \mathbb{E}_{\mathcal{X}_{\tau}}
    \bigg[ \bigg| f_{gnn}^{(2)}(\nabla f_{\tau}^{(1), *} (\vect{x}_{\tau}),~\mathcal{G}_{\tau}^{(2), *}; [\matr{\Theta}_{gnn}^{(2)}]_{\tau-1}) - (r_{\tau} -  f_{gnn}^{(1)}(\vect{x}_{\tau},~\mathcal{G}_{\tau}^{(1), *}; [\matr{\Theta}_{gnn}^{(1)}]_{\tau-1})) \bigg| \bigg] \\
    & - \mathbb{E}_{\mathcal{X}_{\tau}} \bigg[ \bigg| f_{gnn}^{(2)}(\nabla f_{\tau}^{(1), *} (\vect{x}_{\tau}),~\mathcal{G}_{\tau}^{(2), *}; [\matr{\Theta}_{gnn}^{(2)}]_{\tau-1}) - (r_{\tau} -  f_{gnn}^{(1)}(\vect{x}_{\tau},~\mathcal{G}_{\tau}^{(1), *}; [\matr{\Theta}_{gnn}^{(1)}]_{\tau-1})) \bigg| \bigg] = 0.
\end{split}
\end{displaymath}
where $F_{\tau}$ denotes the filtration of all the past records $\mathcal{P}_{\tau}$ up to time step $\tau$. 
Then, we will have the mean value for this sequence as
\begin{displaymath}
\begin{split}
    & \frac{1}{t}\sum_{\tau\in [t]}V_{\tau} = \\
    & \frac{1}{t}\sum_{\tau\in [t]} \mathbb{E}_{ \mathcal{X}_{\tau}}
    \bigg[ \bigg| f_{gnn}^{(2)}(\nabla f_{\tau}^{(1), *} (\vect{x}_{\tau}),~\mathcal{G}_{\tau}^{(2), *}; [\matr{\Theta}_{gnn}^{(2)}]_{\tau-1}) - (r_{\tau} -  f_{gnn}^{(1)}(\vect{x}_{\tau},~\mathcal{G}_{\tau}^{(1), *}; [\matr{\Theta}_{gnn}^{(1)}]_{\tau-1})) \bigg| \bigg] \\
    & - 
    \frac{1}{t}\sum_{\tau\in [t]} \bigg| f_{gnn}^{(2)}(\nabla f_{\tau}^{(1), *} (\vect{x}_{\tau}),~\mathcal{G}_{\tau}^{(2), *}; [\matr{\Theta}_{gnn}^{(2)}]_{\tau-1}) - (r_{\tau} -  f_{gnn}^{(1)}(\vect{x}_{\tau},~\mathcal{G}_{\tau}^{(1), *}; [\matr{\Theta}_{gnn}^{(1)}]_{\tau-1})) \bigg|.
\end{split}
\end{displaymath}
As it has shown that the sequence is a martingale difference sequence, by directly applying the Azuma-Hoeffding inequality, we could bound the difference between the mean and its expectation as
\begin{displaymath}
\begin{split}
    \mathbb{P} \bigg[ 
    \frac{1}{t}\sum_{\tau\in [t]}V_{\tau} - \frac{1}{t}\sum_{\tau\in [t]} \mathbb{E} [V_{\tau}] \geq (1+2\gamma_{2}) \sqrt{\frac{2\log(1 / \delta)}{t}}
    \bigg] \leq \delta
\end{split}
\end{displaymath}
with the probability at least $1 - 2\delta$. Since it has shown that the $V_{\tau}$ is of zero expectation, we have the second term on the LHS of the inequality to be zero.
Then, the inequality above is equivalent to 
\begin{displaymath}
\begin{split}
    & \frac{1}{t}\sum_{\tau\in [t]}V_{\tau} \leq (1+2\gamma_{2}) \sqrt{\frac{2\log(1 / \delta)}{t}} \implies \\
    & \frac{1}{t}\sum_{\tau\in [t]} \mathbb{E}_{ \mathcal{X}_{\tau}}
    \bigg[ \bigg| f_{gnn}^{(2)}(\nabla f_{\tau}^{(1), *} (\vect{x}_{\tau}),~\mathcal{G}_{\tau}^{(2), *}; [\matr{\Theta}_{gnn}^{(2)}]_{\tau-1}) - (r_{\tau}  -  f_{gnn}^{(1)}(\vect{x}_{\tau},~\mathcal{G}_{\tau}^{(1), *}; [\matr{\Theta}_{gnn}^{(1)}]_{\tau-1})) \bigg| \bigg] \\
    & \leq \frac{1}{t}\sum_{\tau\in [t]} 
    \bigg| f_{gnn}^{(2)}(\nabla f_{\tau}^{(1), *} (\vect{x}_{\tau}),~\mathcal{G}_{\tau}^{(2), *}; [\matr{\Theta}_{gnn}^{(2)}]_{\tau-1}) - (r_{\tau} -  f_{gnn}^{(1)}(\vect{x}_{\tau},~\mathcal{G}_{\tau}^{(1), *}; [\matr{\Theta}_{gnn}^{(1)}]_{\tau-1})) \bigg| \\
    & \qquad + (1+2\gamma_{2}) \sqrt{\frac{2\log(1 / \delta)}{t}}
\end{split}
\end{displaymath}
with the probability at least $1 - 2\delta$. Then, for the RHS of the above inequality, by further applying \blemma \ref{lemma_convergence_GNN_f_1} and \blemma \ref{lemma_GNN_network_bound_sampled_parameters}, we have
\begin{displaymath}
\begin{split}
    & \frac{1}{t}\sum_{\tau\in [t]} 
    \bigg| f_{gnn}^{(2)}(\nabla f_{\tau}^{(1), *} (\vect{x}_{\tau}),~\mathcal{G}_{\tau}^{(2), *}; [\matr{\Theta}_{gnn}^{(2)}]_{\tau-1}) - (r_{\tau} -  f_{gnn}^{(1)}(\vect{x}_{\tau},~\mathcal{G}_{\tau}^{(1), *}; [\matr{\Theta}_{gnn}^{(1)}]_{\tau-1})) \bigg| \\
    & \quad \leq 
    \frac{1}{t}\sum_{\tau\in [t]} 
    \bigg| f_{gnn}^{(2)}(\nabla f_{\tau}^{(1), *} (\vect{x}_{\tau}),~\mathcal{G}_{\tau}^{(2), *}; \widetilde{\matr{\Theta}}_{agg}^{(1)}) - (r_{\tau} -  f_{gnn}^{(1)}(\vect{x}_{\tau},~\mathcal{G}_{\tau}^{(1), *}; [\matr{\Theta}_{gnn}^{(1)}]_{\tau-1})) \bigg| \\
    & \qquad + \frac{3L\sqrt{2t}}{2} 
\end{split}
\end{displaymath}
with regard to the parameter $\widetilde{\matr{\Theta}}_{agg}^{(1)}$ s.t. $\norm{\widetilde{\matr{\Theta}}_{agg}^{(1)} - [\matr{\Theta}_{gnn}^{(2)}]_{0}}_{2} \leq \mathcal{O} \left( \frac{ t^3}{ \rho \sqrt{m}} \log m \right)$. Therefore, by applying the conclusion from \blemma \ref{lemma_convergence_GNN_f_1}, we could bound the empirical loss w.r.t. $\widetilde{\matr{\Theta}}_{agg}^{(1)}$ as
\begin{displaymath}
\begin{split}
    & \frac{1}{t}\sum_{\tau\in [t]} 
    \bigg| f_{gnn}^{(2)}(\nabla f_{\tau}^{(1), *} (\vect{x}_{\tau}),~\mathcal{G}_{\tau}^{(2), *}; \widetilde{\matr{\Theta}}_{agg}^{(1)}) - (r_{\tau} -  f_{gnn}^{(1)}(\vect{x}_{\tau},~\mathcal{G}_{\tau}^{(1), *}; [\matr{\Theta}_{gnn}^{(1)}]_{\tau-1})) \bigg| \\
    & \leq \frac{1}{\sqrt{t}} 
    \sqrt{\sum_{\tau\in [t]} 
    \bigg| f_{gnn}^{(2)}(\nabla f_{\tau}^{(1), *} (\vect{x}_{\tau}),~\mathcal{G}_{\tau}^{(2), *}; \widetilde{\matr{\Theta}}_{agg}^{(1)}) - (r_{\tau} -  f_{gnn}^{(1)}(\vect{x}_{\tau},~\mathcal{G}_{\tau}^{(1), *}; [\matr{\Theta}_{gnn}^{(1)}]_{\tau-1})) \bigg| ^{2}} \\
    & \leq \sqrt{\frac{2\xi_{2}}{t}}.
\end{split}
\end{displaymath}
Finally, assembling all the components and applying the union bound would complete the proof.

\qed

% ------------------------------------------------------------------------------------------------------------------------
% ------------------------------------------------------------------------------------------------------------------------

Analogous to the \blemma \ref{lemma_GNN_f_1_output_bound}, we could also have the following corollary of the generalization results for the optimal arms and their rewards $\{\vect{x}_{\tau}^{*}, r_{\tau}^{*}\}_{\tau\in [t]}$ up to round $t$. Then, let $[\widehat{\matr{\Theta}}_{gnn}^{(1), *}]_{t}$ be the parameter that is trained on $\{\vect{x}_{\tau}^{*}, r_{\tau}^{*}\}_{\tau\in [t]}$, and denote $[\widehat{\matr{\Theta}}_{gnn}^{(2), *}]_{t}$ as the parameter of $f_{gnn}^{(2)}(\cdot)$ trained on corresponding gradients and residuals.

\begin{corollary} \label{corollary_term_I_1_optimal_arm}
For the constants $\rho \in (0, \mathcal{O}(\frac{1}{L}))$ and $\xi_{2} \in (0, 1)$, given user $u\in\mathcal{U}$ and its past records $\mathcal{P}_{u, t}$, we suppose $m, \eta_{1}, \eta_{2}, J_{1}, J_{2}$ satisfy the conditions in \btheorem \ref{theorem_regret_bound}, and randomly draw the parameter $[\matr{\Theta}_{gnn}^{(1), *}]_{t} \sim \{[\widehat{\matr{\Theta}}_{gnn}^{(1), *}]_{\tau}\}_{\tau\in [t]}$, $[\matr{\Theta}_{gnn}^{(2), *}]_{t} \sim \{[\widehat{\matr{\Theta}}_{gnn}^{(2), *}]_{\tau}\}_{\tau\in [t]}$.
Then, with probability at least $1 - \delta$ given a sampled arm-reward pair $(\vect{x}, r)$, we have
\begin{displaymath}
\begin{split}
    & \sum_{\tau\in [t]} \mathbb{E} 
    \bigg[ \bigg| f_{gnn}^{(2)}\bigg(\nabla f_{t}^{(1), *} (\vect{x}_{t}^{*}),~\mathcal{G}_{t, *}^{(2), *}; [\matr{\Theta}_{gnn}^{(2)}]_{t-1}\bigg) - \bigg(r_{t}^{*} -  f_{gnn}^{(1)}(\vect{x}_{t},~\mathcal{G}_{t, *}^{(1), *}; [\matr{\Theta}_{gnn}^{(1)}]_{t-1}) \bigg) \bigg|  | u_{t}, \mathcal{X}_{t} \bigg] \\
    & \quad \leq 
    \sqrt{t} \cdot \bigg( \sqrt{2\xi_{2}} + \frac{3L}{\sqrt{2}} + (1 + \gamma_{2}) \sqrt{2\log(\frac{tn\cdot a}{\delta})}\bigg) + \Gamma_{t}
\end{split}
\end{displaymath}
where 
\begin{displaymath}
    \gamma_{2} = 
    2 + \mathcal{O} \left( \frac{ t^3 L}{  \rho\sqrt{m}} \log m \right)
    + \mathcal{O} \left(  \frac{ L^{2}t^4 }{ \rho^{4 / 3} m^{1 / 6}} \log^{11 / 6} (m) \right).
\end{displaymath}

\end{corollary}

\textbf{Proof.}
The proof of this corollary is comparable to the proof of \blemma \ref{lemma_term_I_1_selected_arm}. At each time step $t$, regarding the definition of the optimal arm, we have $\vect{x}_{t}^{*} = \max_{\vect{x}_{i, t}\in \mathcal{X}_{t}} \mathbb{E}[r_{i, t} | u_{t}, \vect{x}_{i, t}]$. Then, analogously, we could define the difference sequence as
\begin{displaymath}
\begin{split}
    V_{\tau}^{*} = & \mathbb{E}_{ \mathcal{X}_{\tau}}
    \bigg[ \bigg| f_{gnn}^{(2)}(\nabla f_{\tau}^{(1), *} (\vect{x}_{\tau}^{*}),~\mathcal{G}_{\tau}^{(2), *}; [\matr{\Theta}_{gnn}^{(2), *}]_{\tau-1}) - (r_{\tau} -  f_{gnn}^{(1)}(\vect{x}_{\tau}^{*},~\mathcal{G}_{\tau}^{(1), *}; [\matr{\Theta}_{gnn}^{(1), *}]_{\tau-1})) \bigg| \bigg] \\
    & - \bigg| f_{gnn}^{(2)}(\nabla f_{\tau}^{(1), *} (\vect{x}_{\tau}^{*}),~\mathcal{G}_{\tau}^{(2), *}; [\matr{\Theta}_{gnn}^{(2), *}]_{\tau-1}) - (r_{\tau} -  f_{gnn}^{(1)}(\vect{x}_{\tau}^{*},~\mathcal{G}_{\tau}^{(1), *}; [\matr{\Theta}_{gnn}^{(1), *}]_{\tau-1})) \bigg|
\end{split}
\end{displaymath}
where by reusing the notation, we denote $\mathcal{G}_{\tau}^{(1), *}, \mathcal{G}_{\tau}^{(2), *}$ to be the true user graphs w.r.t. the optimal arm $\vect{x}_{\tau}^{*}$ here. Then, similar to the proof of \blemma \ref{lemma_term_I_1_selected_arm}, we have the sequence to be the martingale difference sequence as 
\begin{displaymath}
\begin{split}
    \mathbb{E}[V_{\tau}^{*} & \big| F_{\tau}^{*} ] = 
    \mathbb{E}_{ \mathcal{X}_{\tau}}
    \bigg[ \bigg| f_{gnn}^{(2)}(\nabla f_{\tau}^{(1), *} (\vect{x}_{\tau}^{*}),~\mathcal{G}_{\tau}^{(2), *}; [\matr{\Theta}_{gnn}^{(2), *}]_{\tau-1}) - (r_{\tau} -  f_{gnn}^{(1)}(\vect{x}_{\tau}^{*},~\mathcal{G}_{\tau}^{(1), *}; [\matr{\Theta}_{gnn}^{(1), *}]_{\tau-1})) \bigg| \bigg] \\
    & - 
    \mathbb{E}_{ \mathcal{X}_{\tau}}
    \bigg[ \bigg| f_{gnn}^{(2), *}(\nabla f_{\tau}^{(1), *} (\vect{x}_{\tau}^{*}),~\mathcal{G}_{\tau}^{(2), *}; [\matr{\Theta}_{gnn}^{(2), *}]_{\tau-1}) - (r_{\tau} -  f_{gnn}^{(1)}(\vect{x}_{\tau}^{*},~\mathcal{G}_{\tau}^{(1), *}; [\matr{\Theta}_{gnn}^{(1), *}]_{\tau-1})) \bigg| \bigg] = 0
\end{split}
\end{displaymath}
with $F_{\tau}^{*}$ being the filtration of past optimal arms up to round $\tau$. Then, we could also applying the Azuma-Hoeffding inequality to bound the difference between the mean $\frac{1}{t} \sum_{\tau\in [t]} V_{\tau}^{*}$ and its expectation $\frac{1}{t} \sum_{\tau\in [t]} \mathbb{E}[V_{\tau}^{*}]$. Finally, like in the proof of \blemma \ref{lemma_term_I_1_selected_arm}, applying the conclusion from \blemma \ref{lemma_convergence_GNN_f_1} and \blemma \ref{lemma_GNN_network_bound_sampled_parameters} would complete the proof.

\qed

% ---------------------------------------- 
Then, recall the definition of of the confidence bound function $\textsf{CB}_{t}(\vect{x}_{t}^{*})$ w.r.t. the optimal arm $\vect{x}_{t}^{*}$, we 
the corresponding term $I_{1}$ as
\begin{displaymath}
\begin{split}
    I_{1} = \mathbb{E} \bigg[ \abs{f_{gnn}^{(2)}(\nabla f_{t}^{(1), *} (\vect{x}_{t}^{*}),~\mathcal{G}_{t}^{(2), *}; [\matr{\Theta}_{gnn}^{(2)}]_{t-1}) - (r_{t} -  f_{gnn}^{(1)}(\vect{x}_{t}^{*},~\mathcal{G}_{t}^{(1), *}; [\matr{\Theta}_{gnn}^{(1)}]_{t-1}))}  \bigg| u_{t}, \mathcal{X}_{t} \bigg].
\end{split}
\end{displaymath}
And it can be further decomposed as
\begin{displaymath}
\begin{split}
    & \abs{f_{gnn}^{(2)}(\nabla f_{t}^{(1), *} (\vect{x}_{t}^{*}),~\mathcal{G}_{t}^{(2), *}; [\matr{\Theta}_{gnn}^{(2)}]_{t-1}) - (r_{t} -  f_{gnn}^{(1)}(\vect{x}_{t}^{*},~\mathcal{G}_{t}^{(1), *}; [\matr{\Theta}_{gnn}^{(1)}]_{t-1}))} \\
    & \leq \abs{f_{gnn}^{(2)}(\nabla f_{t}^{(1), *} (\vect{x}_{t}^{*}),~\mathcal{G}_{t}^{(2), *}; [\matr{\Theta}_{gnn}^{(2), *}]_{t-1}) - (r_{t} -  f_{gnn}^{(1)}(\vect{x}_{t}^{*},~\mathcal{G}_{t}^{(1), *}; [\matr{\Theta}_{gnn}^{(1), *}]_{t-1}))} + \\
    & \qquad + 
    \abs{f_{gnn}^{(1)}(\vect{x}_{t}^{*},~\mathcal{G}_{t}^{(1), *}; [\matr{\Theta}_{gnn}^{(1), *}]_{t-1}) - f_{gnn}^{(1)}(\vect{x}_{t}^{*},~\mathcal{G}_{t}^{(1), *}; [\matr{\Theta}_{gnn}^{(1)}]_{t-1})} \\
    & \qquad +
    \abs{f_{gnn}^{(2)}(\nabla f_{t}^{(1), *} (\vect{x}_{t}^{*}),~\mathcal{G}_{t}^{(2), *}; [\matr{\Theta}_{gnn}^{(2), *}]_{t-1}) - f_{gnn}^{(2)}(\nabla f_{t}^{(1), *} (\vect{x}_{t}^{*}),~\mathcal{G}_{t}^{(2), *}; [\matr{\Theta}_{gnn}^{(2)}]_{t-1})} 
\end{split}
\end{displaymath}
where the first term on the RHS could be bounded by \textbf{Corollary} \ref{corollary_term_I_1_optimal_arm}. Then, for the second term, we first denote $\vect{h}_{i}^{*} \in \mathbb{R}^{m}$ to be the aggregated hidden representation w.r.t. the user-arm pair $(u_{i}, \vect{x}_{t}^{*})$ where $u_{i}$ is the $i$-th user. Here, $\vect{h}_{t}^{*}$ is essentially the row in the aggregated representation matrix $\matr{H}_{agg}$ corresponding to the user arm pair $(u_{t}, \vect{x}_{t}^{*})$. Therefore, for the received user $u_{t}\in \mathcal{U}$, the reward estimation based on two samples regarding the two sets of parameters would have the same the input $\vect{h}_{t}^{*}$.
Then, for the second term, since the outputs w.r.t. two sets of parameters have the same input $\vect{h}_{t}^{*}$, we could apply the conclusion from \blemma \ref{lemma_FC_network_different_parameters}, which will lead to
\begin{displaymath}
\begin{split}
    \abs{f_{gnn}^{(1)}(\vect{x}_{t}^{*}, & ~\mathcal{G}_{t}^{(1), *}; [\matr{\Theta}_{gnn}^{(1), *}]_{t-1}) - f_{gnn}^{(1)}(\vect{x}_{t}^{*},~\mathcal{G}_{t}^{(1), *}; [\matr{\Theta}_{gnn}^{(1)}]_{t-1})} \\
    & \leq 
    \bigg(1 + \mathcal{O}(\frac{tL^{3} \log^{5/6} (m)}{\rho^{1/3}m^{1/6}})\bigg)\cdot \mathcal{O}(\frac{t^{3}L}{\rho\sqrt{m}}\log(m)) + 
    \mathcal{O}\bigg( \frac{t^{4}L^{2} \log^{11/6} (m)}{\rho^{4/3}m^{1/6}} \bigg).
\end{split}
\end{displaymath}
Analogously, we could also have the same bound for the third term on the RHS. Summing up the bounds for three terms on the RHS would finish deriving the upper bound for term $I_{1}$.

% ------------------------------------------------------------------------------------------------------------------------
% ------------------------------------------------------------------------------------------------------------------------

\subsection{Bounding the Exploitation Graph Estimation Error}   \label{subsec_bounding_I_2}

Then, we proceed to bound the error induced by the estimation of user exploitation graph, i.e., the error term $I_{2}$. Recall that the confidence bound function $\textsf{CB}_{t}(\vect{x})$ for the given arm $\vect{x}\in \mathcal{X}_{t}$ is
\begin{displaymath}
\begin{split}
    & \mathsf{CB}_{t}(\vect{x})  = \mathbb{E} \bigg[ \abs{f_{gnn}^{(2)}(\nabla f_{t}^{(1)} (\vect{x}),~\mathcal{G}^{(2)}; [\matr{\Theta}_{gnn}^{(2)}]_{t-1}) - (r -  f_{gnn}^{(1)}(\vect{x},~\mathcal{G}^{(1)}; [\matr{\Theta}_{gnn}^{(1)}]_{t-1}))}  \bigg| u_{t}, \mathcal{X}_{t} \bigg] \\
    & \leq 
    \underbrace{\mathbb{E} \bigg[ 
    \abs{f_{gnn}^{(1)}(\vect{x},~\mathcal{G}^{(1), *}; [\matr{\Theta}_{gnn}^{(1)}]_{t-1}) - f_{gnn}^{(1)}(\vect{x},~\mathcal{G}^{(1)}; [\matr{\Theta}_{gnn}^{(1)}]_{t-1})}  \bigg| u_{t}, \mathcal{X}_{t} \bigg]}_{I_{2}} + I_{1} + I_{3} + I_{4}.
\end{split}
\end{displaymath}
given an arbitrary arm $\vect{x} \in \mathcal{X}_{t}$. For arm $\vect{x}$, we use the following lemma to bound the error caused by the difference between the estimated exploitation graph $\mathcal{G}^{(1)}$ and the true exploitation graph $\mathcal{G}^{(1), *}$ associated with arm $\vect{x}$.

Denoting the adjacency matrix of the estimated graph $\mathcal{G}^{(1)}$ as $\matr{A}^{(1)}$, and the adjacency matrix for the true user exploitation graph $\mathcal{G}^{(1), *}$ as $\matr{A}^{(1), *}$, we have the normalized adjacency matrices as $\matr{S}^{(1)} = \matr{A}^{(1)} / n$ and $\matr{S}^{(1), *} = \matr{A}^{(1), *} / n$.
For the $i$-th user $u_{i}\in \mathcal{U}$, we could deem the $i$-th row of the matrix multiplication $\matr{S} \cdot \matr{X}$, represented by $\vect{h}_{0, i} = [\matr{S} \cdot \matr{X}]_{i:}$, as the aggregated input for the network for the user-arm pair $(\vect{x}, u_{i})$. Note that in this way, the rest of the network could be regarded as a $L+1$-layer FC network, where the weight matrix for the first layer is $\matr{\Theta}_{agg}^{(1)}$. 
Then, to make sure each aggregated input has the norm of 1, we apply an additional transformation mentioned in \textbf{Eq.} \ref{eq_concat_transformation}
as $\Tilde{\vect{h}}_{0, i} = \phi(\vect{h}_{0, i}, \vect{x}) = 
    (\frac{\vect{h}_{0, i}}{\sqrt{2}}, \frac{\vect{x}}{2}, c_{0, i})$ where $c_{0, i} = \sqrt{\frac{3}{4} - \frac{1}{2}\norm{\vect{h}_{0, i}}_{2}^{2}}$. 
And this transformation ensures $\norm{\Tilde{\vect{h}}_{0, i}}_{2} = 1$ and $c_{0, i} \geq \frac{1}{2}$.
Since this transformation does not alter the original aggregated representation $\vect{h}_{0, i}$, it will not impair the original information w.r.t. the user-arm pair $(\vect{x}, u_{i})$. Meantime, note that this transformation also ensures the separateness of the transformed contexts to be at least $\frac{\rho}{2}$.

% --------------------------------
\begin{lemma}
For the constants $\rho \in (0, \mathcal{O}(\frac{1}{L}))$ and $\xi_{1} \in (0, 1)$, given past records $\mathcal{P}_{t-1}$, we suppose $m, \eta_{1}, \eta_{2}, J_{1}, J_{2}$ satisfy the conditions in \btheorem \ref{theorem_regret_bound}, and randomly draw the parameter $[\matr{\Theta}_{gnn}^{(1)}]_{t} \sim \{[\widehat{\matr{\Theta}}_{gnn}^{(1)}]_{\tau}\}_{\tau\in [t]}$.
Then, with probability at least $1 - \delta$, given an arm $\vect{x}\in \mathbb{R}^{d}$, we have
\begin{displaymath}
\begin{split}
    \sum_{\tau\in [t]} \abs{f_{gnn}^{(1)}(\vect{x},~\mathcal{G}_{\tau}^{(1), *}; [\matr{\Theta}_{gnn}^{(1)}]_{\tau-1}) & - f_{gnn}^{(1)}(\vect{x},~\mathcal{G}_{\tau}^{(1)}; [\matr{\Theta}_{gnn}^{(1)}]_{\tau-1})} \\
    & \leq \mathcal{O}(L) \cdot \sqrt{8t} \cdot \bigg( \sqrt{2\xi_{1}} + \frac{3L}{\sqrt{2}} + (1 + \gamma_{1}) \sqrt{2\log(\frac{tn\cdot a}{\delta})}\bigg) + \Gamma_{t}.
\end{split}
\end{displaymath}

\label{lemma_I_2_bound}
\end{lemma}

\textbf{Proof.}
By the conclusion of \blemma \ref{lemma_user_f_1_est_error}, at time step $t$, the reward estimation error of the user exploitation model could be bounded as
\begin{displaymath}
    \sum_{\tau\in [t]}\mathbb{E} \big[ \abs{f_{u}^{(1)} (\vect{x}; [\matr{\Theta}_{u}^{(1)}]_{t}) - r} ~\big| \mathcal{X}_{t} \big] \leq 
    \sqrt{\frac{t}{n}} \cdot \bigg( \sqrt{2\xi_{1}} + \frac{3L}{\sqrt{2}} + (1 + \gamma_{1}) \sqrt{2\log(\frac{tn\cdot a}{\delta})}\bigg) + \Gamma_{t}.
\end{displaymath}
with the probability at least $1 - \delta$.
And given two users $u_{i}, u_{j} \in \mathcal{U}$ and an arbitrary arm $\vect{x} \in \mathcal{X}_{t}$, we denote their individual reward as $r_{i}, r_{j}$ separately. We omit the expectation notation below for simplicity.
Then, we could bound the absolute difference between the reward estimations as
\begin{displaymath}
\begin{split}
     \abs{ \abs{r_{i} - r_{j}} & -
    \abs{f_{u}^{(1)} (\vect{x}_{i}; [\matr{\Theta}_{u_{i}}^{(1)}]_{t}) - f_{u}^{(1)} (\vect{x}_{j}; [\matr{\Theta}_{u_{j}}^{(1)}]_{t})} } 
    \leq \abs{ (r_{i} - f_{u}^{(1)} (\vect{x}_{i}; [\matr{\Theta}_{u_{i}}^{(1)}]_{t})) - (r_{j} - f_{u}^{(1)} (\vect{x}_{j}; [\matr{\Theta}_{u_{j}}^{(1)}]_{t})) } \\
    & = \abs{ (r_{i} - f_{u}^{(1)} (\vect{x}_{i}; [\matr{\Theta}_{u_{i}}^{(1)}]_{t})) - (r_{j} - f_{u}^{(1)} (\vect{x}_{j}; [\matr{\Theta}_{u_{j}}^{(1)}]_{t})) } \\
    & \leq \abs{ (r_{i} - f_{u}^{(1)} (\vect{x}_{i}; [\matr{\Theta}_{u_{i}}^{(1)}]_{t})) } + 
    \abs{ (r_{j} - f_{u}^{(1)} (\vect{x}_{j}; [\matr{\Theta}_{u_{j}}^{(1)}]_{t})) }.
\end{split}
\end{displaymath}
Based on the definition of the mapping function $\Psi_{1}$, it would naturally be Lipschitz continuous with the coefficient of 1, which is 
\begin{displaymath}
\begin{split}
     \abs{ \exp(-\abs{r_{i} - r_{j}}) -
    \exp(-\abs{f_{u}^{(1)} (\vect{x}_{i}; [\matr{\Theta}_{u_{i}}^{(1)}]_{t}) & - f_{u}^{(1)} (\vect{x}_{j}; [\matr{\Theta}_{u_{j}}^{(1)}]_{t})}) } \\
    &\leq \abs{ \abs{r_{i} - r_{j}} -
    \abs{f_{u}^{(1)} (\vect{x}_{i}; [\matr{\Theta}_{u_{i}}^{(1)}]_{t}) - f_{u}^{(1)} (\vect{x}_{j}; [\matr{\Theta}_{u_{j}}^{(1)}]_{t})} }.
\end{split}
\end{displaymath}

with the probability at least $1 - \delta$. Finally, applying the union bound for all the $(n^{2} - n) / 2$ user pairs and re-scaling the $\delta$ would give us the estimation error bound for the reward difference for each pair of users. 
To achieve the upper bound, we apply the Corollary \ref{corollary_user_f_1_est_error_other_arms} by considering the trajectory $\tilde{\mathcal{P}}_{u, t}$ consists of the past arm-reward pairs $\{\vect{x}_{i_{\tau}, \tau}, r_{i_{\tau}, \tau}\}_{\tau\in [t]}$, where arm $\vect{x}_{i_{\tau}, \tau}$ leads to the largest estimation error of the estimation model $f_{u_{\tau}}^{(1)}(\cdot)$ in each round $\tau\in [t]$.  
Thus, we have the bound for the edge weight difference, where the difference of an arbitrary $i$-th row could be bounded by
\begin{displaymath}
\begin{split}
    \sum_{\tau\in [t]}\norm{ [\matr{A}_{\tau}^{(1)}]_{i:} - [\matr{A}_{\tau}^{(1), *}]_{i:} }_{2} 
    \leq 2n \sqrt{t} \cdot \bigg( \sqrt{2\xi_{1}} + \frac{3L}{\sqrt{2}} + (1 + \gamma_{1}) \sqrt{2\log(\frac{tn\cdot a}{\delta})}\bigg) + \Gamma_{t},
\end{split}
\end{displaymath}
which implies
\begin{displaymath}
\begin{split}
    \sum_{\tau\in [t]}\norm{ [\matr{S}_{\tau}^{(1)}]_{i:} - [\matr{S}_{\tau}^{(1), *}]_{i:} }_{2} 
    \leq 2 \sqrt{t} \cdot \bigg( \sqrt{2\xi_{1}} + \frac{3L}{\sqrt{2}} + (1 + \gamma_{1}) \sqrt{2\log(\frac{tn\cdot a}{\delta})}\bigg) + \Gamma_{t}.
\end{split}
\end{displaymath}
Therefore, applying the conclusions from \blemma \ref{lemma_user_f_1_est_error}, it leads to
% \begin{displaymath}
% \begin{split}
%     & \abs{r_{i} - r_{j}} = 
%     \abs{r_{i} - f_{u}^{(1)} (\vect{x}; [\matr{\Theta}_{u_{i}}^{(1)}]_{t}) 
%     + f_{u}^{(1)} (\vect{x}; [\matr{\Theta}_{u_{i}}^{(1)}]_{t}) - 
%     f_{u}^{(1)} (\vect{x}; [\matr{\Theta}_{u_{j}}^{(1)}]_{t})
%     + f_{u}^{(1)} (\vect{x}; [\matr{\Theta}_{u_{j}}^{(1)}]_{t}) - r_{j}} \\
%     & \quad \leq \abs{r_{i} - f_{u}^{(1)} (\vect{x}; [\matr{\Theta}_{u_{i}}^{(1)}]_{t}) }
%     + \abs{ f_{u}^{(1)} (\vect{x}; [\matr{\Theta}_{u_{i}}^{(1)}]_{t}) - 
%     f_{u}^{(1)} (\vect{x}; [\matr{\Theta}_{u_{j}}^{(1)}]_{t}) }
%     + \abs{ f_{u}^{(1)} (\vect{x}; [\matr{\Theta}_{u_{j}}^{(1)}]_{t}) - r_{j} } \\
%     & \quad \leq 
%     2\sqrt{\frac{1}{t}} \cdot \bigg( \sqrt{2\xi_{1}} + \frac{3L}{\sqrt{2}} + (1 + \gamma_{1}) \sqrt{2\log(\frac{tn\cdot a}{\delta})}\bigg) + 
%     \abs{ f_{u}^{(1)} (\vect{x}; [\matr{\Theta}_{u_{i}}^{(1)}]_{t}) - 
%     f_{u}^{(1)} (\vect{x}; [\matr{\Theta}_{u_{j}}^{(1)}]_{t}) }.
% \end{split}
% \end{displaymath}
% And this is equivalent to 
\begin{displaymath}
\begin{split}
    \sum_{\tau\in [t]} \abs{ \abs{r_{i, \tau} - r_{j, \tau}} -
    \abs{f_{u}^{(1)} (\vect{x}_{i, \tau}; &[\matr{\Theta}_{u_{i}}^{(1)}]_{t}) -  f_{u}^{(1)} (\vect{x}_{j, \tau}; [\matr{\Theta}_{u_{j}}^{(1)}]_{t})} } \\ & \leq 
    2 \sqrt{\frac{t}{n}} \cdot \bigg( \sqrt{2\xi_{1}} + \frac{3L}{\sqrt{2}} + (1 + \gamma_{1}) \sqrt{2\log(\frac{tn\cdot a}{\delta})}\bigg) + \Gamma_{t}
\end{split}
\end{displaymath}
Afterwards, recalling the transformation at the beginning of this subsection, and given an user-arm pair $(u_{i}, \vect{x})$ for the $i$-th user, we denote $\vect{h} = [\matr{S}^{(1)} \cdot \matr{X}]_{i:}$ and $\vect{h}^{*} = [\matr{S}^{(1), *} \cdot \matr{X}]_{i:}$. Based the aforementioned transformation in \textbf{Eq.} \ref{eq_concat_transformation}, their transformed form could naturally be $\Tilde{\vect{h}} = (\frac{\sqrt{2}}{2}\vect{h}, \frac{\vect{x}}{2}, c)$ and $\Tilde{\vect{h}}^{*} = (\frac{\sqrt{2}}{2}\vect{h}^{*}, \frac{\vect{x}}{2}, c^{*})$ with $\norm{\vect{x}}_{2} = 1$. Without the loss of generality, we let $c > c^{*}$.
Then, we could have

\begin{displaymath}
\begin{split}
    \norm{ \Tilde{\vect{h}} - \Tilde{\vect{h}}^{*} }_{2} & = 
    \sqrt{ \norm{\vect{h} - \vect{h}^{*} }_{2}^{2} + (c - c^{*})^{2} }
    \underset{(\text{i})}{\leq} \sqrt{ \norm{\vect{h} - \vect{h}^{*} }_{2}^{2} + (c^{2} - (c^{*})^{2})^{2} } \\
    & \underset{(\text{ii})}{=} \sqrt{ \norm{\vect{h} - \vect{h}^{*} }_{2}^{2} + \frac{1}{4}(\norm{\vect{h}^{*}}_{2}^{2} - \norm{\vect{h}}_{2}^{2})^{2} } \\
    & = \sqrt{ \norm{\vect{h} - \vect{h}^{*} }_{2}^{2} + \frac{1}{4}( \norm{\vect{h}^{*} - \vect{h}}_{2} \cdot \norm{\vect{h}^{*} + \vect{h}}_{2} )^{2} } \\
    & \underset{(\text{iii})}{\leq} \sqrt{2} \cdot \norm{\vect{h} - \vect{h}^{*}}_{2}
\end{split}
\end{displaymath}
Here, (i) is because $c, c^{*} \geq \frac{1}{2}$. (ii) is because of $c^{2} + \frac{\norm{\vect{h}}_{2}^{2}}{2} = (c^{*})^{2} + \frac{\norm{\vect{h}^{*}}_{2}^{2}}{2} = \frac{3}{4}$, and (iii) is due to $\norm{\vect{h}}_{2}, \norm{\vect{h}^{*}}_{2} \leq 1$.

% \begin{displaymath}
% \begin{split}
%     \norm{ \Tilde{\vect{h}} - \Tilde{\vect{h}}^{*} }_{2} & = \sqrt{ \norm{\Tilde{\vect{h}} - \Tilde{\vect{h}}^{*} }_{2}^{2} + (c - c^{*})^{2} }
%     \underset{(\text{i})}{\leq} \sqrt{ \norm{\Tilde{\vect{h}} - \Tilde{\vect{h}}^{*} }_{2}^{2} + c^{2} - (c^{*})^{2} } \\
%     & \underset{(\text{ii})}{=} \sqrt{ \norm{\Tilde{\vect{h}} - \Tilde{\vect{h}}^{*} }_{2}^{2} + \norm{\Tilde{\vect{h}}^{*}}_{2}^{2} - \norm{\Tilde{\vect{h}}}_{2}^{2} } \\
%     & \underset{(\text{iii})}{\leq} \sqrt{2\cdot \norm{\vect{h} - \vect{h}^{*}}_{2}}.
% \end{split}
% \end{displaymath}
% Here, (i) is because $(a - b)^{2} \leq a^{2} - b^{2}$ for any $a, b \geq 0$ and $a \geq b$. (ii) is because of $c^{2} + \norm{\Tilde{\vect{h}}}_{2}^{2} = 1$, and (iii) is due to $\norm{\vect{h}}_{2}, \norm{\vect{h}^{*}}_{2} \leq 1$.
% Then, we proceed to bound $\norm{\vect{h} - \vect{h}^{*}}_{2}$. Recall the definition from \textbf{Eq.} \ref{eq_new_embedding_matrix}, we have

Then, we proceed to bound $\norm{\vect{h} - \vect{h}^{*}}_{2}$. Recall the definition from \textbf{Eq.} \ref{eq_new_embedding_matrix}. Extending the above conclusion across different rounds $\tau\in [t]$, we will have
\begin{displaymath}
\begin{split}
    \sum_{\tau\in [t]}\norm{\vect{h}_{\tau} - \vect{h}_{\tau}^{*}}_{2} \leq \norm{\vect{x}_{\tau}} \cdot \norm{ [\matr{S}_{\tau}^{(1)}]_{i:} - [\matr{S}_{\tau}^{(1), *}]_{i:} }_{2} 
    \leq 2 \sqrt{t} \bigg( \sqrt{2\xi_{1}} + \frac{3L}{\sqrt{2}} + (1 + \gamma_{1}) \sqrt{2\log(\frac{tn\cdot a}{\delta})}\bigg) + \Gamma_{t}.
\end{split}
\end{displaymath}
% Therefore, we end up with the bound
% \begin{displaymath}
% \begin{split}
%     \norm{ \Tilde{\vect{h}} - \Tilde{\vect{h}}^{*} }_{2} & \leq 2\sqrt{2} \cdot \sqrt{\frac{1}{t}} \cdot \bigg( \sqrt{2\xi_{1}} + \frac{3L}{\sqrt{2}} + (1 + \gamma_{1}) \sqrt{2\log(\frac{tn\cdot a}{\delta})}\bigg).
% \end{split}
% \end{displaymath}
Finally, combining the conclusion from \blemma \ref{lemma_FC_network_Lipschitz}, we finally have
\begin{displaymath}
\begin{split}
    \sum_{\tau\in [t]} \abs{f_{gnn}^{(1)}(\vect{x},~\mathcal{G}_{\tau}^{(1), *}; [\matr{\Theta}_{gnn}^{(1)}]_{\tau-1}) & - f_{gnn}^{(1)}(\vect{x},~\mathcal{G}_{\tau}^{(1)}; [\matr{\Theta}_{gnn}^{(1)}]_{\tau-1})} \\
    & \leq \mathcal{O}(L) \cdot \sqrt{8t} \cdot \bigg( \sqrt{2\xi_{1}} + \frac{3L}{\sqrt{2}} + (1 + \gamma_{1}) \sqrt{2\log(\frac{tn\cdot a}{\delta})}\bigg) + \Gamma_{t}
\end{split}
\end{displaymath}
which concludes the proof.

\qed

% --------------------

\subsection{Bounding the Exploration Graph Estimation Error}
Again, recall the definition of the confidence bound function $\textsf{CB}_{t}(\vect{x})$ which is 
\begin{displaymath}
\begin{split}
    & \mathsf{CB}_{t}(\vect{x})  = \mathbb{E} \bigg[ \abs{f_{gnn}^{(2)}(\nabla f_{t}^{(1)} (\vect{x}),~\mathcal{G}^{(2)}; [\matr{\Theta}_{gnn}^{(2)}]_{t-1}) - (r -  f_{gnn}^{(1)}(\vect{x},~\mathcal{G}^{(1)}; [\matr{\Theta}_{gnn}^{(1)}]_{t-1}))}  \bigg| u_{t}, \mathcal{X}_{t} \bigg] \\
    & \leq \underbrace{\mathbb{E} \bigg[ \abs{f_{gnn}^{(2)}(\nabla f_{t}^{(1), *} (\vect{x}),~\mathcal{G}^{(2), *}; [\matr{\Theta}_{gnn}^{(2)}]_{t-1}) - f_{gnn}^{(2)}(\nabla f_{t}^{(1), *} (\vect{x}),~\mathcal{G}^{(2)}; [\matr{\Theta}_{gnn}^{(2)}]_{t-1})}  \bigg| u_{t}, \mathcal{X}_{t}\bigg]}_{I_{3}} \\
    & \qquad + I_{1} + I_{2} + I_{4}.
\end{split}
\end{displaymath}
Analogous to the procedure for the user exploitation graph, we have the following lemma to bound the error induced by user exploitation graph estimation.

% --------------------------------
\begin{lemma}
For the constants $\rho \in (0, \mathcal{O}(\frac{1}{L}))$ and $\xi_{1} \in (0, 1)$, given past records $\mathcal{P}_{t-1}$, we suppose $m, \eta_{1}, \eta_{2}, J_{1}, J_{2}$ satisfy the conditions in \btheorem \ref{theorem_regret_bound}, and randomly draw the parameter $[\matr{\Theta}_{gnn}^{(2)}]_{t} \sim \{[\widehat{\matr{\Theta}}_{gnn}^{(2)}]_{\tau}\}_{\tau\in [t]}$.
Then, with probability at least $1 - \delta$, given an arm $\vect{x}\in \mathbb{R}^{d}$, we have
\begin{displaymath}
\begin{split}
    \sum_{\tau\in [t]} \abs{f_{gnn}^{(2)}(\nabla f_{t}^{(1), *} & (\vect{x}),~\mathcal{G}^{(2), *}; [\matr{\Theta}_{gnn}^{(2)}]_{t-1})  - f_{gnn}^{(2)}(\nabla f_{t}^{(1), *} (\vect{x}),~\mathcal{G}^{(2)}; [\matr{\Theta}_{gnn}^{(2)}]_{t-1})} \\
    & \leq \mathcal{O}(\xi_{L}) \cdot \sqrt{8t} \cdot \bigg( \sqrt{2\xi_{1}} + \frac{3L}{\sqrt{2}} + (1 + \gamma_{1}) \sqrt{2\log(\frac{tn\cdot a}{\delta})}\bigg) + \Gamma_{t}.
\end{split}
\end{displaymath}

\label{lemma_I_3_bound}
\end{lemma}

\textbf{Proof.} The proof of this lemma could be derived based on a similar approach as in \blemma \ref{lemma_I_2_bound}. 
Recall that for the exploration GNN model $f_{gnn}^{(2)} (\cdot)$, we have the gradients of the GNN exploitation model $\nabla f_{u, t}^{(1)} (\vect{x}) = \frac{\nabla_{[\matr{\Theta}_{u}^{(1)}]_{t}} f_{u}^{(1)} (\vect{x}; [\matr{\Theta}_{u}^{(1)}]_{t})}{c'_{g}L}$ as the input given an arm $\vect{x}$ and user $u\in \mathcal{U}$, whose norm $\norm{\nabla f_{u, t}^{(1)} (\vect{x})}_{2} \leq 1$.

Given two users $u_{i}, u_{j} \in \mathcal{U}$ and an arbitrary arm $\vect{x} \in \mathcal{X}_{t}$, we denote their individual reward as $r_{i}, r_{j}$ separately. Then, we could bound the absolute difference between the potential gain estimations as
\begin{displaymath}
\begin{split}
    & \abs{ \abs{ (r_{i} - f_{u}^{(1)} (\vect{x}; [\matr{\Theta}_{u_{i}}^{(1)}]_{t})) - (r_{j} - f_{u}^{(1)} (\vect{x}; [\matr{\Theta}_{u_{j}}^{(1)}]_{t})) }  -
    \abs{ f_{u}^{(2)} (\nabla f_{u_{i}, t}^{(1)} (\vect{x}); [\matr{\Theta}_{u_{i}}^{(2)}]_{t}) - f_{u}^{(2)} (\nabla f_{u_{j}, t}^{(1)} (\vect{x}); [\matr{\Theta}_{u_{j}}^{(2)}]_{t}) } } \\
    & \quad \leq \abs{ f_{u}^{(2)} (\nabla f_{u_{i}, t}^{(1)} (\vect{x}); [\matr{\Theta}_{u_{i}}^{(2)}]_{t})) - (r_{i} - f_{u}^{(1)} (\vect{x}; [\matr{\Theta}_{u_{i}}^{(1)}]_{t}))  } \\
    & \qquad\qquad\qquad
    +  \abs{ f_{u}^{(2)} (\nabla f_{u_{j}, t}^{(1)} (\vect{x}); [\matr{\Theta}_{u_{j}}^{(2)}]_{t}) - (r_{j} - f_{u}^{(1)} (\vect{x}; [\matr{\Theta}_{u_{j}}^{(1)}]_{t})) }.
\end{split}
\end{displaymath}
Afterwards, applying the conclusion from \blemma \ref{lemma_user_f_2_est_error} would lead to the result that
\begin{displaymath}
\begin{split}
    & \sum_{\tau\in [t]} \abs{ \abs{ (r_{i} - f_{u}^{(1)} (\vect{x}; [\matr{\Theta}_{u_{i}}^{(1)}]_{t})) - (r_{j} - f_{u}^{(1)} (\vect{x}; [\matr{\Theta}_{u_{j}}^{(1)}]_{t})) }  -
    \abs{ f_{u}^{(2)} (\nabla f_{u_{i}, t}^{(1)} (\vect{x}); [\matr{\Theta}_{u_{i}}^{(2)}]_{t}) - f_{u}^{(2)} (\nabla f_{u_{j}, t}^{(1)} (\vect{x}); [\matr{\Theta}_{u_{j}}^{(2)}]_{t}) } } \\
    & \qquad \leq 2 \sqrt{\frac{t}{n}} \cdot \bigg( \sqrt{2\xi_{1}} + \frac{3L}{\sqrt{2}} + (1 + \gamma_{1}) \sqrt{2\log(\frac{tn\cdot a}{\delta})}\bigg) + \Gamma_{t}.
\end{split}
\end{displaymath}
Following a similar approach as in the proof of \blemma \ref{lemma_I_2_bound}, we proceed to consider the aggregated hidden representations for the input gradients. Since the entries between $\matr{A}^{(2)} - \matr{A}^{(2), *}$ (and also the distance between $\matr{S}^{(2)} - \matr{S}^{(2), *}$) are bounded, by adopting the aforementioned transformation in \textbf{Eq.} \ref{eq_concat_transformation} on the aggregated hidden representations for the input gradients and the initial arm contexts $\vect{x}$, we would end up with the bound for the difference between transformed representations for input gradients. Finally, combining the conclusion from \blemma \ref{lemma_FC_network_Lipschitz} would give the proof.

\qed

% --------------------
% \clearpage
\subsection{Bounding the Gradient Input Estimation Error}

For the last term $I_{4}$ in the confidence bound function $\textsf{CB}_{t}(\vect{x})$, we have
\begin{displaymath}
\begin{split}
    I_{4} = \mathbb{E} \bigg[ \abs{f_{gnn}^{(2)}(\nabla f_{t}^{(1), *} (\vect{x}),~\mathcal{G}^{(2)}; [\matr{\Theta}_{gnn}^{(2)}]_{t-1}) - f_{gnn}^{(2)}(\nabla f_{t}^{(1)} (\vect{x}),~\mathcal{G}^{(2)}; [\matr{\Theta}_{gnn}^{(2)}]_{t-1})}  \bigg| u_{t}, \mathcal{X}_{t}\bigg]
\end{split}
\end{displaymath}
which represents the estimation error induced by the difference of input gradients. And we first bound the gradient difference with the following lemma.

\begin{lemma}
For the constants $\rho \in (0, \mathcal{O}(\frac{1}{L}))$ and $\xi_{1} \in (0, 1)$, given past records $\mathcal{P}_{t-1}$, we suppose $m, \eta_{1}, \eta_{2}, J_{1}, J_{2}$ satisfy the conditions in \btheorem \ref{theorem_regret_bound}, and randomly draw the parameter $[\matr{\Theta}_{gnn}^{(1)}]_{t} \sim \{[\widehat{\matr{\Theta}}_{gnn}^{(1)}]_{\tau}\}_{\tau\in [t]}$.
Then, with probability at least $1 - \delta$, given an arm $\vect{x}\in \mathbb{R}^{d}$, we have
\begin{displaymath}
\begin{split}
     \sum_{\tau\in [t]} &\norm{ \nabla f_{\tau}^{(1)} (\vect{x})  - \nabla f_{\tau}^{(1), *} (\vect{x}) }_{2} 
     \leq \mathcal{O} ( \frac{tL^{7/2}  \sqrt{\log m}}{\sqrt{mn}})
\end{split}
\end{displaymath}
where $\nabla f_{t}^{(1)} (\vect{x}) = \frac{\nabla_{\matr{\Theta}_{gnn}^{(1)}} f_{gnn}^{(1)}(\vect{x},~\mathcal{G}^{(1)}; [\matr{\Theta}_{gnn}^{(1)}]_{t-1})}{c_{g}L}$, and $\nabla f_{t}^{(1), *} (\vect{x}) = \frac{\nabla_{\matr{\Theta}_{gnn}^{(1)}} f_{gnn}^{(1)}(\vect{x},~\mathcal{G}^{(1), *}; [\matr{\Theta}_{gnn}^{(1)}]_{t-1})}{c_{g}L}$.

\label{lemma_I_4_gradient_difference_bound}
\end{lemma}

\textbf{Proof.}
% Following the aggregation procedure and transformation procedure shown in section \ref{subsec_bounding_I_2}, we have the transformed representations for given an user-arm pair $(u_{i}, \vect{x})$ with the $i$-th user, which are $\vect{h} = [\matr{S}^{(1)} \cdot \matr{X}]_{i:}$ and $\vect{h}^{*} = [\matr{S}^{(1), *} \cdot \matr{X}]_{i:}$. And their transformed form could naturally be $\Tilde{\vect{h}} = (\vect{h}, c)$ and $\Tilde{\vect{h}}^{*} = (\vect{h}^{*}, c^{*})$. From the conclusion of \blemma \ref{lemma_I_2_bound}, we have
% \begin{displaymath}
% \begin{split}
%     \sum_{\tau\in [t]} \norm{ \Tilde{\vect{h}}_{\tau} - \Tilde{\vect{h}}_{\tau}^{*} }_{2} \leq \sqrt{8t} \cdot \bigg( \sqrt{2\xi_{1}} + \frac{3L}{\sqrt{2}} + (1 + \gamma_{1}) \sqrt{2\log(\frac{tn\cdot a}{\delta})}\bigg) + \Gamma_{t}.
% \end{split}
% \end{displaymath}
% Then, applying the conclusion from \blemma \ref{lemma_FC_network_Lipschitz} would complete the proof.
%
Since the ReLU activation $\sigma(\cdot)$ is not Lipschitz smooth, we propose to bound the gradient difference from the perspective of parameter shift. Recall that in \textbf{Eq.} \ref{eq_GNN_aggegation}, with $k=1$, we have
\begin{displaymath}
\begin{split}
    \matr{H}_{agg} = \sigma\big(\matr{S}_{i, t}^{(1)} \cdot (\matr{X}_{i, t} \matr{\Theta}_{agg}^{(1)})\big).
\end{split}
\end{displaymath}
It is known that the gradient shift can be directly measured by the model parameter shift \cite{conv_theory-allen2019convergence,generalization_bound_cao2019generalization}.
In this case, since the model parameters $\matr{\Theta}_{gnn}^{(2)}$ stay the same for both $ \nabla f_{\tau}^{(1)} (\vect{x}) $ and $ \nabla f_{\tau}^{(1), *} (\vect{x})$, we can naturally integrate the effects of different adjacency matrices with the model parameters. Here, given the same input arm context $\vect{x}$, let $\widetilde{\matr{\Theta}}_{agg}^{(1)}, \widetilde{\matr{\Theta}}_{agg}^{(1), *}$ separately represent the integrated parameters for the aggregation layer w.r.t. $ \nabla f_{\tau}^{(1)} (\vect{x}) $ and $ \nabla f_{\tau}^{(1), *} (\vect{x})$.
With all other layers stay the same, the parameter shift can be directly measure by the difference $\norm{\widetilde{\matr{\Theta}}_{agg}^{(1)} - \widetilde{\matr{\Theta}}_{agg}^{(1), *}}$.

Then, we proceed to measure the parameter shift $\norm{\widetilde{\matr{\Theta}}_{agg}^{(1)} - \widetilde{\matr{\Theta}}_{agg}^{(1), *}}$.
Based on the aggregation operation, and given an user-arm pair $(u_{i}, \vect{x})$ for the $i$-th user, we denote $\vect{h} = [\matr{S}^{(1)} \cdot \matr{X}]_{i:}$ and $\vect{h}^{*} = [\matr{S}^{(1), *} \cdot \matr{X}]_{i:}$. 
By definition, $\vect{h}$ refers to the aggregated embedding for the user-arm pair $(u_{i}, \vect{x})$ with the estimated user graph, while $\vect{h}$ refers to the aggregated embedding with the true user graph.
Alternatively, with a new matrix $\widetilde{S}_{i}\in \mathbb{R}^{d\times nd}$ for user $u_{i}$, we can consider each aggregated embedding to be generated by $\vect{h} = \vect{x}^{\intercal}\cdot \widetilde{S}_{i}$, where
\begin{displaymath}
\begin{split}
    \widetilde{S}_{i} = 
\left(
    \begin{array}{cccc}
    w^{(1)}(u_{1}, u_{i}) & \matr{0} & \cdots & \matr{0} \\
    \matr{0} & w^{(1)}(u_{1}, u_{i}) & \cdots & \matr{0} \\
    \vdots  &       & \ddots   & \vdots \\
    \matr{0} & \matr{0} & \cdots & w^{(1)}(u_{1}, u_{i}) \\
    \end{array} \cdots
    \begin{array}{cccc}
    w^{(1)}(u_{n}, u_{i}) & \matr{0} & \cdots & \matr{0} \\
    \matr{0} & w^{(1)}(u_{n}, u_{i}) & \cdots & \matr{0} \\
    \vdots  &       & \ddots   & \vdots \\
    \matr{0} & \matr{0} & \cdots & w^{(1)}(u_{n}, u_{i}) \\
    \end{array}
\right) \in \mathbb{R}^{d\times nd}.
\end{split}
\end{displaymath}
Analogously, we can also define the matrix $\widetilde{S}^{*}_{i}\in \mathbb{R}^{d\times nd}$ for the true user graph. Therefore, the Euclidean difference $\norm{\vect{h} - \vect{h}^{*}}_{2} \leq \norm{\vect{x}}_{2} \cdot \norm{\widetilde{S}_{i} - \widetilde{S}^{*}_{i}}_{2} \leq \norm{\widetilde{S}_{i} - \widetilde{S}^{*}_{i}}_{2}$. Let $\Delta \widetilde{S}_{i} = \widetilde{S}_{i} - \widetilde{S}^{*}_{i}$. Obviously, we can represent the $\widetilde{\matr{\Theta}}_{agg}^{(1)} = \Delta \widetilde{S}_{i} \cdot\matr{\Theta}_{agg}^{(1)}$. By definition, it is obvious that each element of the matrix subtraction $|[\Delta \widetilde{S}_{i}]_{p, q}| \leq \frac{1}{n}, 0\geq p \leq n, 0\geq q\leq nd$.
Here, by the fact that $\norm{\Delta \widetilde{S}_{i}}_{2}^{2} = \max_{\vect{z}, \norm{\vect{z}}=1} [\vect{z}^{\intercal} ((\Delta \widetilde{S}_{i})^{\intercal}\Delta \widetilde{S}_{i}) \vect{z}]$ and the Rayleigh-Ritz theorem, we can prove $\norm{\Delta \widetilde{S}_{i}}_{2} = \norm{(\Delta \widetilde{S}_{i})^{\intercal}}_{2}$. 

Then, let us denote $\Delta w_{j}^{(1)} = w^{(1)}(u_{j}, u_{i}) - w^{(1), *}(u_{j}, u_{i})$ for the same given user $u_{i}$, which corresponds to the elements in $\Delta \widetilde{S}_{i}$.
For $\norm{(\Delta \widetilde{S}_{i})^{\intercal}}_{2}$, given any $\norm{\vect{z}}_{2} = 1$, we see that $\norm{(\Delta \widetilde{S}_{i})^{\intercal}\cdot \vect{z}}_{2} = \sqrt{\sum_{j=1}^{n} \Delta (w_{j}^{(1)})^{2} \norm{x}_{2}^{2}} \leq \frac{1}{\sqrt{n}}$, which means that $\norm{\Delta \widetilde{S}_{i}} \leq \frac{1}{\sqrt{n}}$.
Recall that we aim to measure the parameter shift $\norm{\widetilde{\matr{\Theta}}_{agg}^{(1)} - \widetilde{\matr{\Theta}}_{agg}^{(1), *}}$. Since the other parameter matrices stay the same for both $ \nabla f_{\tau}^{(1)} (\vect{x}) $ and $ \nabla f_{\tau}^{(1), *} (\vect{x})$, the parameter shift solely comes from the $\Delta \widetilde{S}$, which means that $\norm{\widetilde{\matr{\Theta}}_{agg}^{(1)} - \widetilde{\matr{\Theta}}_{agg}^{(1), *}} \leq \mathcal{O}(\frac{1}{\sqrt{n}})$. 
With $n \geq \mathcal{O}(\text{Poly}(L, \log(m)))$, based on \textbf{Lemma} \ref{lemma_theorem5_allenzhu}, we have 
$$
\norm{ \nabla f_{\tau}^{(1)} (\vect{x})  - \nabla f_{\tau}^{(1), *} (\vect{x}) }_{2}
\leq \mathcal{O} ( \frac{\omega L^3  \sqrt{\log m}}{\sqrt{m}}) \|   \nabla_\Theta f(\vect{x}; \Theta_0)    \|_2 
\leq \mathcal{O} ( \frac{L^{7/2}  \sqrt{\log m}}{\sqrt{mn}}).
$$

\qed

% -------------------------------------------------------------
% -------------------------------------------------------------
% -------------------------------------------------------------

Then, we have the following lemma to bound the term $I_{4}$.

\begin{lemma}
For the constants $\rho \in (0, \mathcal{O}(\frac{1}{L}))$ and $\xi_{1} \in (0, 1)$, given past records $\mathcal{P}_{t-1}$, we suppose $m, \eta_{1}, \eta_{2}, J_{1}, J_{2}$ satisfy the conditions in \btheorem \ref{theorem_regret_bound}, and randomly draw the parameter $[\matr{\Theta}_{gnn}^{(1)}]_{t} \sim \{[\widehat{\matr{\Theta}}_{gnn}^{(1)}]_{\tau}\}_{\tau\in [t]}$.
Then, with probability at least $1 - \delta$, given an arm $\vect{x}\in \mathbb{R}^{d}$, we have
\begin{displaymath}
\begin{split}
     \sum_{\tau\in [t]} \abs{f_{gnn}^{(2)}(\nabla f_{\tau}^{(1), *} (\vect{x})  ,~\mathcal{G}^{(2)}; [\matr{\Theta}_{gnn}^{(2)}]_{\tau-1}) &- f_{gnn}^{(2)}(\nabla f_{\tau}^{(1)} (\vect{x}),~\mathcal{G}^{(2)}; [\matr{\Theta}_{gnn}^{(2)}]_{\tau-1})}
      \leq \\
     & \mathcal{O} ( \frac{\xi_{L}tL^{7/2}  \sqrt{\log m}}{\sqrt{mn}}) +
     \bigg(1 + \mathcal{O}(\frac{tL^{3} \log^{5/6} (m)}{\rho^{1/3}m^{1/6}})\bigg)\cdot \mathcal{O}(\frac{t^{3}L}{\rho\sqrt{m}}\log(m)) + 
    \mathcal{O}\bigg( \frac{t^{4}L^{2} \log^{11/6} (m)}{\rho^{4/3}m^{1/6}} \bigg).
\end{split}
\end{displaymath}

\label{lemma_I_4_overall_bound}
\end{lemma}

\textbf{Proof.}
The proof can be obtained by directly applying \textbf{Lemma} \ref{lemma_I_4_gradient_difference_bound} and \textbf{Lemma} \ref{lemma_FC_network_Lipschitz}.

\qed

% ==================================================================
% \clearpage
\subsection{Lemmas for Over-parameterized Networks}

Based on the results from Subsection \ref{subsec_user_over_param_nets}, with $\mathcal{P}_{t-1}$ as the training data, we have the following convergence result for the exploitation GNN network $f_{gnn}^{(1)}(\cdot; \matr{\Theta}_{gnn}^{(1)})$ after GD.

\begin{lemma} [Lemma \ref{lemma_convergence_user_f_1} extended] \label{lemma_convergence_GNN_f_1}
For any $ 0 < \xi_{2} \leq 1$, $ 0 < \rho \leq \mathcal{O}(\frac{1}{L})$. Given past records $\mathcal{P}_{t-1}$, suppose $m, \eta_{1}, \eta_{2}, J_{1}, J_{2}$ satisfy the conditions in \btheorem \ref{theorem_regret_bound}, then with probability at least $1 - \delta$, we could have
\begin{enumerate}
    \item $\mathcal{L}(\matr{\Theta}_{gnn}^{(1)}) \leq \xi_{2}$ after $J_{2}$ iterations of GD.
    \item For any $j \in [J_{2}]$, $\norm{ [\matr{\Theta}_{gnn}^{(1)}]^{j}  -  [\matr{\Theta}_{gnn}^{(1)}]^{0}  } \leq \mathcal{O} \left( \frac{ t^3}{ \rho \sqrt{m}} \log m \right)$.
\end{enumerate}
\end{lemma}
In particular, \blemma \ref{lemma_convergence_GNN_f_1} above provides the convergence guarantee for $f_{gnn}^{(1)}(\cdot; \matr{\Theta}_{gnn}^{(1)})$ after certain rounds of GD training on the past records $\mathcal{P}_{t-1}$.

% -------------------
\begin{lemma} [Lemma \ref{lemma_FC_network_inner_product} extended] \label{lemma_GNN_network_inner_product}
Assume a constant $\omega$ such that $ \mathcal{O}( m^{-3/2} L^{-3/2} [ \log (TnL^2/\delta) ]^{3/2}  )   \leq \omega \leq  \mathcal{O} (L^{-6}[\log m]^{-3/2} )$ and $n$ training samples. With randomly initialized $[\matr{\Theta}_{gnn}^{(1)}]_{0}$, for parameters $\matr{\Theta}, \matr{\Theta}'$ satisfying $\norm{\matr{\Theta} - [\matr{\Theta}_{gnn}^{(1)}]_{0}}, \norm{\matr{\Theta} - [\matr{\Theta}_{gnn}^{(1)}]_{0}} \leq \omega$, we have
\begin{displaymath}
    \abs{f^{(1)} (\vect{x}; \matr{\Theta}) -  f^{(1)} (\vect{x}; \matr{\Theta}') - \inp{\nabla_{\matr{\Theta}'} f^{(1)} (\vect{x}; \matr{\Theta}')}{\matr{\Theta} - \matr{\Theta}'}} \leq 
    \mathcal{O}(\omega^{1/3}L^{2}\sqrt{m\log(m)}) \norm{\matr{\Theta} - \matr{\Theta}'}
\end{displaymath}
with the probability at least $1-\delta$.
\end{lemma}

% -------------------
\begin{lemma} [Lemma \ref{lemma_FC_network_output_and_gradient} extended] \label{lemma_GNN_network_output_and_gradient}
Assume $m, \eta_{1}, \eta_{2}, J_{1}, J_{2}$ satisfy the conditions in \btheorem \ref{theorem_regret_bound} and $[\matr{\Theta}_{gnn}^{(1)}]_{0}$ being randomly initialized.
Then, with probability at least $1 - \delta$ and given an arm $\norm{\vect{x}}_{2}=1$, we have
\begin{enumerate}
    \item $\abs{ f_{gnn}^{(1)} (\vect{x}; [\matr{\Theta}_{gnn}^{(1)}]_{0}) } \leq 2$,
    \item $ \norm{\nabla_{[\matr{\Theta}_{gnn}^{(1)}]_{0}} f_{gnn}^{(1)} (\vect{x}; [\matr{\Theta}_{gnn}^{(1)}]_{0})}_{2} \leq \mathcal{O}(L) $.
\end{enumerate}
\end{lemma}

\textbf{Proof.}
The conclusion (1) is a direct application of Lemma 7.1 in \citep{conv_theory-allen2019convergence}. For conclusion (2), for each weight matrix $\matr{\Theta}_{l} \in \{\matr{\Theta}_{0}^{(1)}, \matr{\Theta}_{1}^{(1)}, \dots, \matr{\Theta}_{L}^{(1)}\}$ where $\matr{\Theta}_{0}^{(1)} = \matr{\Theta}_{agg}^{(1)}$, we have
\begin{displaymath}
    \norm{\nabla_{\matr{\Theta}} f_{gnn}^{(1)} (\vect{x}; [\matr{\Theta}_{gnn}^{(1)}]_{0})}_{2} = 
    \norm{ (\matr{\Theta}_{L} \matr{D}_{L-1} \cdots \matr{D}_{l+1}\matr{\Theta}_{l+1})\cdot (\matr{D}_{l+1} \matr{\Theta}_{l+1} \cdots \matr{D}_{1}\matr{\Theta}_{1} \matr{D}_{0}\matr{\Theta}_{0}) \cdot \vect{h}^{\intercal} }_{2} \leq \mathcal{O}(\sqrt{L})
\end{displaymath}
by applying Lemma 7.3 in \citep{conv_theory-allen2019convergence}, and $\vect{h}$ denotes the aggregated hidden representation for each user-pair, namely the corresponding row in $\matr{H}_{agg}$. Therefore, by combining the bounds for all the weight matrices, we could have
\begin{displaymath}
\begin{split}
    \norm{\nabla_{[\matr{\Theta}_{gnn}^{(1)}]_{0}} f_{gnn}^{(1)} (\vect{x}; [\matr{\Theta}_{gnn}^{(1)}]_{0})}_{2} = \sqrt{\sum_{l\in \{0, \dots, L\}} \norm{\nabla_{\matr{\Theta}} f_{gnn}^{(1)} (\vect{x}; [\matr{\Theta}_{gnn}^{(1)}]_{0})}_{2}^{2}} = \mathcal{O}(L).
\end{split}
\end{displaymath}
which finishes the proof.

\qed

% -------------------
\begin{lemma} [Lemma \ref{lemma_FC_network_gradient_difference} extended] \label{lemma_GNN_network_gradient_difference}
Assume the training parameters $m, \eta_{2}, J_{2}$ satisfy the conditions in \btheorem \ref{theorem_regret_bound} and $[\matr{\Theta}_{gnn}^{(1)}]_{0}$ being randomly initialized.
Then, with probability at least $1 - \delta$, and for all parameter $\matr{\Theta}_{gnn}^{(1)}$ such that $\norm{\matr{\Theta}_{gnn}^{(1)} - [\matr{\Theta}_{gnn}^{(1)}]_{0}}_{2} \leq \omega $, we have
\begin{displaymath}
    \norm{ \nabla_{\matr{\Theta}_{gnn}^{(1)}} f_{gnn}^{(1)} (\vect{x}; \matr{\Theta}_{gnn}^{(1)}) - \nabla_{[\matr{\Theta}_{gnn}^{(1)}]_{0}} f_{gnn}^{(1)} (\vect{x}; [\matr{\Theta}_{gnn}^{(1)}]_{0}) }_{2} \leq
    \mathcal{O}(\omega^{1/3}L^{3}\sqrt{\log(m)})
\end{displaymath}
\end{lemma}

% -------------------
\begin{lemma} [Lemma \ref{lemma_FC_network_bound_sampled_parameters} extended] \label{lemma_GNN_network_bound_sampled_parameters}

Assume $m, \eta_{2}$ satisfy the condition in \btheorem \ref{theorem_regret_bound}. 
With the probability at least $1-\delta$, we have

\begin{displaymath}
\begin{split}
    \sum_{\tau\in [t]} & \abs{ f(\vect{x}_{\tau}; [\widehat{\matr{\Theta}}_{gnn}^{(1)}]_{\tau}) - r_\tau} \leq
    \sum_{\tau\in [t]} \abs{ f(\vect{x}_{\tau}; [\widehat{\matr{\Theta}}_{gnn}^{(1)}]_{t}) - r_\tau} + 
    \frac{3L\sqrt{2t}}{2} 
\end{split}  
\end{displaymath}

\end{lemma}

\textbf{Proof.}
With the notation from Lemma 4.3 in \citep{generalization_bound_cao2019generalization}, set $R=\frac{t^{3}\log(m)}{\delta}$, $\nu=R^{2}$, and $\epsilon=\frac{LR}{\sqrt{2\nu t}}$. Then, 
considering the loss function to be $\mathcal{L}(\matr{\Theta}_{gnn}^{(1)}) := \sum_{\tau\in [t]} \abs{ f(\vect{x}_{\tau}; \matr{\Theta}_{gnn}^{(1)}) - r_\tau}$ would complete the proof.
\qed

\begin{lemma} \label{lemma_theorem5_allenzhu}
Suppose $m, \eta_1, \eta_2$ satisfy the conditions in Theorem \ref{theorem_regret_bound}. 
With probability at least $1- \mathcal{O}(TkL^2)\exp[-\Omega(m \omega^{2/3} L)] $ over randomness of $\Theta_0$, for all $t \in [T], i \in [k]$, $\Theta$ satisfying $ \| \Theta  - \Theta_0 \|_2 \leq \omega $ with $\omega \leq \mathcal{O}(\text{Poly}(L, \log(m)))$, it holds uniformly:
\[
\| \nabla_\Theta f(\vect{x}; \Theta) -    \nabla_\Theta f(\vect{x}; \Theta_0) \| \leq \mathcal{O} ( \frac{\omega L^3  \sqrt{\log m}}{\sqrt{m}}) \|   \nabla_\Theta f(\vect{x}; \Theta_0)    \|_2 
\]
\end{lemma}

\textbf{Proof.}
This is an application of Theorem 5 \cite{conv_theory-allen2019convergence}. It holds uniformly that
\[
\| \sqrt{m} \nabla_\Theta f(\vect{x}; \Theta) -    \sqrt{m} \nabla_\Theta f(\vect{x}; \Theta_0) \| \leq \mathcal{O} (\omega L^3  \sqrt{\log m}) \|   \nabla_\Theta f(\vect{x}; \Theta_0)    \|_2 
\]
where the scale factor $\sqrt{m}$ is because the last layer of neural network in \cite{conv_theory-allen2019convergence} is initialized based on $N(0 ,1)$ while our last layer is initialized from $N(0 ,1/m)$.

\qed

% --------------------------------------
\begin{lemma}   \label{lemma_FC_network_Lipschitz}
Consider a L-layer fully-connected network $f(\cdot; \matr{\Theta}_{t})$ initialized w.r.t. Subsection \ref{subsec_GNN_arch}. For any $ 0 < \xi_{2} \leq 1$, $ 0 < \rho \leq \mathcal{O}(\frac{1}{L})$. Given the training data set with $t$ samples satisfying the unit-length and the $\rho$-separateness assumption, suppose the training parameters $m, \eta_{2}, J_{2}$ satisfy the conditions in \btheorem \ref{theorem_regret_bound}. Then, with probability at least $1 - \delta$, we have
\begin{displaymath}
\begin{split}
    & \abs{ f(\vect{x}; \matr{\Theta}_{t}) - f(\vect{x}'; \matr{\Theta}_{t}) } \leq \mathcal{O}(\xi_{L})\cdot \norm{ \vect{x} - \vect{x}' }_{2} + 
    \bigg(1 + \mathcal{O}(\frac{tL^{3} \log^{5/6} (m)}{\rho^{1/3}m^{1/6}})\bigg)\cdot \mathcal{O}(\frac{t^{3}L}{\rho\sqrt{m}}\log(m)) + 
    \mathcal{O}\bigg( \frac{t^{4}L^{2} \log^{11/6} (m)}{\rho^{4/3}m^{1/6}} \bigg) \\
    % & \norm{ \nabla_{\matr{\Theta}_{t}} f(\vect{x}; \matr{\Theta}_{t}) - \nabla_{\matr{\Theta}_{t}} f(\vect{x}'; \matr{\Theta}_{t}) }_{2} \leq
    % \mathcal{O}(\frac{tL^{4}\log^{5/6}(m)}{\rho^{1/3}m^{1/6}}) + 
    % \mathcal{O}(L)\cdot \norm{ \vect{x} - \vect{x}' }_{2}
\end{split}  
\end{displaymath}
when given two new samples $\vect{x}, \vect{x}'$ and $c_{\xi} > 0$, $\xi_{L} = (c_{\xi})^{L}$.

\end{lemma}

\textbf{Proof.}
This lemma can be proved based on the fact that the ReLU activation is 1-Lipschitz continuous, and the weight matrix of each layer in the neural network has the norm of $\mathcal{O}(1)$ at random initialization based on our Gaussian initialization of the weight matrices. Finally, combining the conclusion from \textbf{Lemma} \ref{lemma_FC_network_different_parameters} to bound $\abs{ f(\vect{x}; \matr{\Theta}_{t}) - f(\vect{x}; \matr{\Theta}_{0}) }$ will complete the proof.

\qed

% --------------------------------------
\begin{lemma}   \label{lemma_FC_network_different_parameters}
Consider a L-layer fully-connected network $f(\cdot; \matr{\Theta}_{t})$ initialized w.r.t. Section \ref{subsec_GNN_arch}. For any $ 0 < \xi_{2} \leq 1$, $ 0 < \rho \leq \mathcal{O}(\frac{1}{L})$. Let there be two sets of training samples $\mathcal{P}_{t}, \mathcal{P}'_{t}$ with the unit-length and the $\rho$-separateness assumption, and let $\matr{\Theta}_{t}$ be the trained parameter on $\mathcal{P}_{t}$ while $\matr{\Theta}'_{t}$ is the trained parameter on $\mathcal{P}'_{t}$.
Suppose $m, \eta_{1}, \eta_{2}, J_{1}, J_{2}$ satisfy the conditions in \btheorem \ref{theorem_regret_bound}. 
Then, with probability at least $1 - \delta$, we have
\begin{displaymath}
\begin{split}
    \abs{ f(\vect{x}; \matr{\Theta}_{t}) - & f(\vect{x}; \matr{\Theta}'_{t}) } \leq \\
    & \bigg(1 + \mathcal{O}(\frac{tL^{3} \log^{5/6} (m)}{\rho^{1/3}m^{1/6}})\bigg)\cdot \mathcal{O}(\frac{t^{3}L}{\rho\sqrt{m}}\log(m)) + 
    \mathcal{O}\bigg( \frac{t^{4}L^{2} \log^{11/6} (m)}{\rho^{4/3}m^{1/6}} \bigg)
    % & \norm{ \nabla_{\matr{\Theta}_{t}} f(\vect{x}; \matr{\Theta}_{t}) - \nabla_{\matr{\Theta}'_{t}} f(\vect{x}; \matr{\Theta}'_{t}) }_{2} \leq
\end{split}  
\end{displaymath}
when given a new sample $\vect{x} \in \mathbb{R}^{d}$.

\end{lemma}

\textbf{Proof.}
First, based on the conclusion from Theorem 1 from \citep{conv_theory-allen2019convergence} and regarding the $t$ samples, the trained the parameters satisfy $\norm{\matr{\Theta}_{t} - \matr{\Theta}_{0}}_{2}, \norm{\matr{\Theta}'_{t} - \matr{\Theta}_{0}}_{2} \leq \mathcal{O}(\frac{t^
{3}}{\rho\sqrt{m}}\log(m)) = \omega$ where $\matr{\Theta}_{0}$ is the randomly initialized parameter. Then, we could have
\begin{displaymath}
\begin{split}
    & \norm{\nabla_{\matr{\Theta}_{t}} f(\vect{x}; \matr{\Theta}_{t})}_{2} \leq 
    \norm{\nabla_{\matr{\Theta}_{0}} f(\vect{x}; \matr{\Theta}_{0})}_{2} +
    \norm{\nabla_{\matr{\Theta}_{t}} f(\vect{x}; \matr{\Theta}_{t}) - \nabla_{\matr{\Theta}_{0}} f(\vect{x}; \matr{\Theta}_{0})}_{2} \\
    & \qquad \leq \bigg(1 + \mathcal{O}(\frac{tL^{3} \log^{5/6} (m)}{\rho^{1/3}m^{1/6}})\bigg)\cdot \mathcal{O}(L)
\end{split}  
\end{displaymath}
w.r.t. the conclusion from Theorem 1 and Theorem 5 of \citep{conv_theory-allen2019convergence}. Then, regarding the Lemma 4.1 from \citep{generalization_bound_cao2019generalization}, we would have
\begin{displaymath}
\begin{split}
    & \abs{ f(\vect{x}; \matr{\Theta}_{t}) - f(\vect{x}; \matr{\Theta}'_{t}) - 
    \inp{\nabla_{\matr{\Theta}'_{t}} f(\vect{x}; \matr{\Theta}'_{t})}{\matr{\Theta}_{t} - \matr{\Theta}'_{t}} } \leq 
    \mathcal{O}(\omega^{1/3} L^{2} \sqrt{m\log(m)}) \cdot \norm{\matr{\Theta}_{t} - \matr{\Theta}'_{t}}_{2}.
\end{split}  
\end{displaymath}
Therefore, the our target could be reformed as
\begin{displaymath}
\begin{split}
    & \abs{ f(\vect{x}; \matr{\Theta}_{t}) - f(\vect{x}; \matr{\Theta}'_{t}) } \leq 
    \norm{ \nabla_{\matr{\Theta}'_{t}} f(\vect{x}; \matr{\Theta}'_{t}) }_{2}  \norm{\matr{\Theta}_{t} - \matr{\Theta}'_{t}}_{2}
    + \mathcal{O}(\omega^{1/3} L^{2} \sqrt{m\log(m)}) \cdot \norm{\matr{\Theta}_{t} - \matr{\Theta}'_{t}}_{2} \\
    & \qquad \leq 
    \bigg(1 + \mathcal{O}(\frac{tL^{3} \log^{5/6} (m)}{\rho^{1/3}m^{1/6}})\bigg)\cdot \mathcal{O}(L) \cdot \omega + \mathcal{O}(\omega^{4/3} L^{2} \sqrt{m\log(m)})
\end{split}  
\end{displaymath}
Substituting the $\omega$ with its value would complete the proof.

\qed

% \input{tex_Appd_tech_lemmas}

% \section{Computational Resources}

% All the experiments are conducted on a Windows machine with an Intel Core i7 CPU, 64GB RAM, and two RTX 5000 GPUs.

\end{document}